\documentclass[11pt]{article}  

\usepackage{times}
\usepackage{fullpage}
\usepackage{algorithm}
\usepackage{algorithmicx,algpseudocode}
\usepackage{amsmath}
\usepackage{amssymb}
\usepackage{amsthm}
\usepackage{array}
\usepackage{bbm}
\usepackage{cancel}
\usepackage{color}
\usepackage{cmap}
\usepackage{enumerate}
\usepackage{enumitem}
\usepackage{mathtools}
\usepackage{mathrsfs}
\usepackage{hyperref}
\usepackage{stackrel}
\usepackage{stmaryrd}
\usepackage{url}
\usepackage{verbatim}
\usepackage{wasysym}
\usepackage{wrapfig}
\usepackage{yhmath}
\usepackage[all,cmtip]{xy}

\usepackage{hyperref}

\usepackage[backend=biber,style=alphabetic]{biblatex}
\newtheorem{thm}{Theorem}[section]
\newtheorem*{thm*}{Theorem}
\newtheorem{prb}[thm]{Problem}
\newtheorem*{prb*}{Problem}

\newtheorem{asm}[thm]{Assumptions}

\newtheorem*{ax*}{Axiom}

\newtheorem*{clm*}{Claim}

\newtheorem*{conj*}{Conjecture}

\newtheorem{df}[thm]{Definition}
\newtheorem*{df*}{Definition}

\newtheorem*{ex*}{Example}

\newtheorem{lem}[thm]{Lemma}
\newtheorem*{lem*}{Lemma}

\newtheorem*{pos*}{Postulate}
\newtheorem{pr}[thm]{Proposition}
\newtheorem*{pr*}{Proposition}

\newtheorem*{qu*}{Question}
\newtheorem{rem}[thm]{Remark}
\newtheorem*{rem*}{Remark}
\def\shownotes{1}  \ifnum\shownotes=1
\newcommand{\authnote}[2]{$\ll$\textsf{\footnotesize #1 notes: #2}$\gg$}
\else
\newcommand{\authnote}[2]{}
\fi

\newcommand{\st}[0]{\text{st}}

\newcommand{\cE}[0]{\mathcal{E}}
\newcommand{\sE}[0]{\mathscr{E}}
\newcommand{\E}[0]{\mathbb{E}}
\newcommand{\EE}[0]{\mathop{\mathbb E}}

\newcommand{\sL}[0]{\mathscr{L}}
\newcommand{\N}[0]{\mathbb{N}}

\newcommand{\sP}[0]{\mathscr{P}}
\newcommand{\Pj}[0]{\mathbb{P}}

\newcommand{\R}[0]{\mathbb{R}}

\newcommand{\sT}[0]{\mathscr{T}}

\newcommand{\one}[0]{\mathbbm{1}}

\newcommand{\al}[0]{\alpha}
\newcommand{\be}[0]{\beta}
\newcommand{\ga}[0]{\gamma}

\newcommand{\de}[0]{\delta}
\newcommand{\De}[0]{\Delta}
\newcommand{\ep}[0]{\varepsilon}

\newcommand{\ka}[0]{\kappa}
\newcommand{\la}[0]{\lambda}

\newcommand{\rh}[0]{\rho}

\newcommand{\Te}[0]{\Theta}

\newcommand{\Om}[0]{\Omega}
\newcommand{\si}[0]{\sigma}
\newcommand{\Si}[0]{\Sigma}

\newcommand{\dellarge}{\Delta} 
\newcommand{\delsmall}{\tau}

\newcommand{\sub}[0]{\subset}

\newcommand{\subeq}[0]{\subseteq}

\newcommand{\iy}[0]{\infty}

\newcommand{\rc}[1]{\frac{1}{#1}}
\newcommand{\prc}[1]{\pa{\rc{#1}}}

\newcommand{\fc}[2]{\frac{#1}{#2}}
\newcommand{\sfc}[2]{\sqrt{\frac{#1}{#2}}}
\newcommand{\pf}[2]{\pa{\frac{#1}{#2}}}

\newcommand{\dd}[2]{\frac{d #1}{d #2}}

\newcommand{\ddd}[1]{\frac{d}{d #1}}

\newcommand{\nb}[0]{\nabla}

\newcommand{\dx}{\,dx}

\newcommand{\lra}[0]{\leftrightarrow}

\newcommand{\cil}[0]{\circlearrowleft}

\newcommand{\ab}[1]{\left| {#1} \right|}
\newcommand{\an}[1]{\left\langle {#1}\right\rangle}
\newcommand{\ba}[1]{\left[ {#1} \right]}
\newcommand{\bc}[1]{\left\{ {#1} \right\}}

\newcommand{\ce}[1]{\left\lceil {#1}\right\rceil}

\newcommand{\pa}[1]{\left( {#1} \right)}

\newcommand{\ve}[1]{\left\Vert {#1}\right\Vert}

\newcommand{\set}[2]{\left\{{#1}:{#2}\right\}}

\newcommand{\ol}[1]{\overline{#1}}

\newcommand{\ub}[2]{\underbrace{#1}_{#2}}

\newcommand{\wt}[1]{\widetilde{#1}}
\newcommand{\wh}[1]{\widehat{#1}}

\newcommand{\amin}{\operatorname{argmin}}

\newcommand{\Exp}{\operatorname{Exp}}

\newcommand{\Gap}{\operatorname{Gap}}

\newcommand{\Id}{\operatorname{Id}}

\newcommand{\KL}[0]{\mbox{KL}}

\newcommand{\poly}{\operatorname{poly}}

\newcommand{\Supp}{\operatorname{Supp}}

\newcommand{\Var}[0]{\operatorname{Var}}

\providecommand{\cal}[1]{\mathcal{#1}}
\renewcommand{\cal}[1]{\mathcal{#1}}

\newcommand{\pull}[9]{
#1\ar@/_/[ddr]_{#2} \ar@{.>}[rd]^{#3} \ar@/^/[rrd]^{#4} & &\\
& #5\ar[r]^{#6}\ar[d]^{#8} &#7\ar[d]^{#9} \\}

\newcommand{\cmp}[9]{
\xymatrix{
#1 \ar[r]^{#4}{#5} \ar@/_2pc/[rr]^{#8}_{#9} & #2 \ar[r]^{#6}_{#7} & #3
}
}

\newcommand{\ha}[1]{\ar@{^(->}[#1]}
\newcommand{\ls}[1]{\ar@{-}[#1]}
\newcommand{\sj}[1]{\ar@{->>}[#1]}
\newcommand{\aq}[1]{\ar@{=}[#1]}
\newcommand{\acir}[1]{\ar@{}[#1]|-{\textstyle{\circlearrowright}}}
\newcommand{\acil}[1]{\ar@{}[#1]|-{\textstyle{\circlearrowleft}}}
\newcommand{\ard}[1]{\ar@{.>}[#1]}
\newcommand{\mt}[1]{\ar@{|->}[#1]}
\newcommand{\inm}[1]{\ar@{}[#1]|-{\in}}
\newcommand{\inr}{\ar@{}[d]|-{\rotatebox[origin=c]{-90}{$\in$}}}
\newcommand{\inl}{\ar@{}[u]|-{\rotatebox[origin=c]{90}{$\in$}}}

\newcommand{\sumr}[2]{\sum_{\scriptsize \begin{array}{c}{#1}\\{#2}\end{array}}}

\newcommand{\sumo}[2]{\sum_{#1=1}^{#2}}

\newcommand{\prodo}[2]{\prod_{#1=1}^{#2}}

\newcommand{\beq}[1]{\begin{equation}\llabel{#1}}
\newcommand{\eeq}[0]{\end{equation}}
\newcommand{\bal}[0]{\begin{align*}}
\newcommand{\eal}[0]{\end{align*}}
\newcommand{\ban}[0]{\begin{align}}
\newcommand{\ean}[0]{\end{align}}

\newcommand{\fixme}[1]{{\color{red}#1}}
\newcommand{\llabel}[1]{\label{#1}\text{\fixme{\tiny#1}}}

\newcommand{\arxiv}[1]{\url{http://www.arxiv.org/abs/#1}}

\newcommand{\vocab}[1]{\textbf{#1}} 

\allowdisplaybreaks[2]

\DeclareFontFamily{U}{wncy}{}
    \DeclareFontShape{U}{wncy}{m}{n}{<->wncyr10}{}
    \DeclareSymbolFont{mcy}{U}{wncy}{m}{n}
    \DeclareMathSymbol{\Sh}{\mathord}{mcy}{"58} 
\newcommand{\weight}[0]{w} 
\newcommand{\mineig}[0]{\ka}
\newcommand{\maxeig}[0]{K}
\newcommand{\Anote}[1]{}
\newcommand{\Rnote}[1]{}
\newcommand{\Hnote}[1]{}
\newcommand{\wexp}[0]{3}
\newcommand{\wexpt}[0]{6}

\newcommand{\citep}[1]{\cite{#1}}

\usepackage{etoolbox}
\newtoggle{thesis}
\togglefalse{thesis}
\newcommand{\ift}[1]{\iftoggle{thesis}{#1}{}}

\numberwithin{equation}{section}

\begin{document}

\title{
Simulated Tempering Langevin Monte Carlo II: \\
An Improved Proof using Soft Markov Chain Decomposition}

\author{Rong Ge\thanks{Duke University, Computer Science Department \texttt{rongge@cs.duke.edu}}, Holden Lee\thanks{Duke University, Mathematics Department \texttt{holee@math.duke.edu}}, Andrej Risteski\thanks{Carnegie Mellon University, Machine Learning Department \texttt{aristesk@andrew.cmu.edu}}}

\date{\today}
\maketitle

\begin{abstract}
A key task in Bayesian machine learning is sampling from distributions that are only specified up to a partition function (i.e., constant of proportionality). One prevalent example of this is sampling posteriors in parametric 
distributions, such as latent-variable generative models.  However sampling (even very approximately) can be \#P-hard.

Classical results (going back to \cite{bakry1985diffusions}) on sampling focus on log-concave distributions, and show a natural Markov process called \emph{Langevin diffusion} mixes in polynomial time. However, all log-concave distributions are uni-modal, while in practice it is very common for the distribution of interest to have multiple modes.
In this case, Langevin diffusion suffers from torpid mixing. 

We address this problem by combining Langevin diffusion with \emph{simulated tempering}. The result is a Markov chain that mixes more rapidly by transitioning between different temperatures of the distribution. 
We analyze this Markov chain for a mixture of (strongly) log-concave distributions of the same shape. In particular, our technique applies to the canonical multi-modal distribution: a mixture of gaussians (of equal variance). 
Our algorithm efficiently samples from these distributions given only access to the gradient of the log-pdf. 
To the best of our knowledge, this is the first result that proves fast mixing for multimodal distributions in this setting. 

For the analysis, we introduce novel techniques for proving spectral gaps based on decomposing the action of the generator of the diffusion. 
Previous approaches rely on decomposing the state space as a partition of \emph{sets}, while our approach can be thought of as decomposing the stationary measure as a mixture of \emph{distributions} (a ``soft partition'').

Additional materials for the paper can be found at~\url{http://holdenlee.github.io/Simulated\%20tempering\%20Langevin\%20Monte\%20Carlo.html}. Note that the proof and results have been improved and generalized from the precursor at~\arxiv{1710.02736}. See Section~\ref{s:comparison} for a comparison. 
\end{abstract}
\pagebreak

\tableofcontents

\section{Introduction} 

\iftoggle{thesis}
{Dealing with multimodal distributions is a core challenge in Markov Chain Monte Carlo. The naive Markov chain often does not mix rapidly, and we obtain samples from only one part of the support of the distribution.}
{Sampling is a fundamental task in Bayesian statistics, and dealing with multimodal distributions is a core challenge. One common technique to sample from a probability distribution is to define a Markov chain with that distribution as its stationary distribution. This general approach is called \emph{Markov chain Monte Carlo}. However, in many practical problems, the Markov chain does not mix rapidly, and we obtain samples from only one part of the support of the distribution.}

Practitioners have dealt with this problem through a variety of heuristics. A popular family of approaches involve changing the \emph{temperature} of the distribution. However, there has been little theoretical analysis of such methods. We give provable guarantees for a temperature-based method called \emph{simulated tempering} when it is combined with \emph{Langevin diffusion}.

\iftoggle{thesis}{}{
More precisely, the setup we consider is sampling from a distribution given up to a constant of proportionality. This is inspired from sampling a \emph{posterior} distribution over the latent variables of a latent-variable Bayesian model with known parameters. In such models, the observable variables $x$ follow a distribution $p(x)$ which has a simple and succinct form \emph{given} the values of some latent variables $h$, i.e., the joint $p(h,x)$ factorizes as $p(h) p(x|h)$ where both factors are explicit. Hence, the \emph{posterior} distribution $p(h|x) $ has the form $p(h|x) = \frac{p(h) p(x|h)}{p(x)}$. Although the numerator is easy to evaluate, the denominator $p(x)=\int p(h)p(x|h)\,dh$ can be NP-hard to approximate even for simple models like topic models \citep{sontag2011complexity}.  Thus the problem is intractable without structural assumptions.}

Previous theoretical results on sampling have focused on {\em log-concave} distributions, i.e., distributions of the form $p(x) \propto e^{-f(x)}$ for a convex function $f(x)$.  This is analogous to {\em convex optimization} where the objective function $f(x)$ is convex.
Recently, there has been renewed interest in analyzing a popular Markov Chain for sampling from such distributions, when given gradient access to $f$---a natural setup for the posterior sampling task described above. In particular, a Markov chain called \emph{Langevin Monte Carlo} (see Section~\ref{sec:overview-l}), popular with Bayesian practitioners, has been proven to work, with various rates depending on the precise properties of $f$ \citep{dalalyan2016theoretical, durmus2016high,dalalyan2017further,cheng2017convergence,durmus2018analysis}.  

Yet, just as many interesting optimization problems are nonconvex, many interesting sampling problems are not log-concave. A log-concave distribution is necessarily uni-modal: its density function has only one local maximum, which is necessarily a global maximum. This fails to capture many interesting scenarios.
Many simple posterior distributions are neither log-concave nor uni-modal, for instance, the posterior distribution of the means for a mixture of gaussians, given a sample of points from the mixture of gaussians. 
In a more practical direction, complicated posterior distributions associated with deep generative models \citep{rezende2014stochastic} and variational auto-encoders \citep{kingma2013auto} are believed to be multimodal as well.  

In this work we initiate an  exploration of provable methods for sampling ``beyond log-concavity,'' in parallel to optimization ``beyond convexity''. As worst-case results are prohibited by hardness results, we must make assumptions on the distributions of interest. 
As a first step, we consider a mixture of strongly log-concave distributions of the same shape. This class of distributions captures the prototypical multimodal distribution,  a mixture of Gaussians with the same covariance matrix. Our result is also robust in the sense that even if the actual distribution has density that is only close to a mixture that we can handle, our algorithm can still sample from the distribution in polynomial time. Note that the requirement that all Gaussians have the same covariance matrix is in some sense necessary: in Appendix~\ref{sec:lb} we show that even if the covariance of two components differ by a constant factor, no algorithm (with query access to $f$ and $\nb f$) can achieve the same robustness guarantee in polynomial time. 

\subsection{Problem statement}

We formalize the problem of interest as follows. 

\ift{\nomenclature[2mui]{$\mu_i$}{Center of $i$th component (Chapter~\ref{ch:mm})}}
\ift{\nomenclature[2wu]{$w_i$}{Weight of $i$th component (Chapter~\ref{ch:mm})}}
\ift{\nomenclature[2f0]{$f_0$}{Base function (Chapter~\ref{ch:mm})}}
\ift{\nomenclature[2fi]{$f_i$}{Translate of base function $f_i(x) = f_0(x-\mu_i)$ (Chapter~\ref{ch:mm})}}
\ift{\nomenclature[1d]{$d$}{Dimension}}
\begin{prb}\label{p:main}
Let $f:\R^d\to \R$ be a function. 
Given query access to $\nabla f(x)$ and $f(x)$ at any point $x \in \mathbb{R}^d$, sample from the probability distribution with density function $p(x) \propto e^{-f(x)}$.

\ift{\nomenclature[2kappa]{$\ka$}{Strong convexity of $f_0$ (Chapter~\ref{ch:mm})}}
\ift{\nomenclature[2K]{$K$}{Smoothness of $f_0$  (Chapter~\ref{ch:mm})}}
In particular, consider the case where $e^{-f(x)}$ is the density function of a mixture of strongly log-concave distributions that are translates of each other. That is, there is a base function $f_0:\R^d\to \R$, centers $\mu_1,\mu_2,\ldots, \mu_m \in \R^d$, and weights $w_1,w_2,\ldots, w_m$ ($\sum_{i=1}^m w_i = 1$) such that 
\begin{equation}
f(x) = -\log\left(\sum_{i=1}^m w_i e^{-f_0(x-\mu_i)}\right), \label{eq:logconcave}
\end{equation}
For notational convenience, we will define $f_i(x) = f_0(x-\mu_i)$. 
\end{prb}
The function $f_0$ specifies a basic ``shape'' around the modes, and the means $\mu_i$ indicate the locations of the modes. 

Without loss of generality we assume the mode of the distribution $e^{-f_0(x)}$ is at $0$ ($\nabla f_0(0) = 0$). We also assume $f_0$ is twice differentiable, and for any $x$ the Hessian is sandwiched between $\mineig I\preceq \nabla^2 f_0(x)) \preceq \maxeig I$. Such functions are called $\mineig$-strongly-convex, $\maxeig$-smooth functions. The corresponding distribution $e^{-f_0(x)}$ are strongly log-concave distributions. 
\footnote{On a first read, we recommend concentrating on the case $f_0(x) = \frac{1}{2\sigma^2}\|x\|^2$. This corresponds to the case where all the components are spherical Gaussians with mean $\mu_i$ and covariance matrix $\sigma^2 I$. }


\subsection{Our results}
\label{s:assumptions}
We show that there is an efficient algorithm that can sample from this distribution given just access to $f(x)$ and $\nabla f(x)$.

\ift{\nomenclature[2D]{$D$}{Bound on centers $\ve{\mu_i}\le D$ (Chapter~\ref{ch:mm})}}
\ift{\nomenclature[2wmin]{$w_{\min}$}{Minimum weight $w_{\min}=\min_i w_i$ (Chapter~\ref{ch:mm})}}
\begin{thm}[Main] Given $f(x)$ as defined in Equation (\ref{eq:logconcave}), where the base function $f_0$ satisfies for any $x$, $\mineig I\preceq \nabla^2 f_0(x) \preceq \maxeig I$,
and $\|\mu_i\| \leq D$ for all $i\in [m]$,
there is an algorithm (given as Algorithm~\ref{a:mainalgo} with appropriate setting of parameters) with running time $\poly\left(\rc{w_{\min}},D,d,\rc{\ep}, \rc{\mineig}, \maxeig\right)$, which given query access to $\nabla f$ and $f$, outputs a sample from a distribution within TV-distance $\ep$ of $p(x)\propto e^{-f(x)}$.
\end{thm}


Note that importantly the algorithm does \emph{not} have direct access to the mixture parameters $\mu_i, w_i, i \in [n]$ (otherwise the problem would be trivial). Sampling from this mixture is thus non-trivial: algorithms that are based on making local steps (such as the ball-walk \citep{lovasz1993random,vempala2005geometric} and Langevin Monte Carlo) cannot move between different components of the gaussian mixture when the gaussians are well-separated. 
In the algorithm we use simulated tempering (see Section~\ref{sec:overview-st}), which is a technique that adjusts the ``temperature'' of the distribution
in order to move between different components.


Of course, requiring the distribution to be {\em exactly} a mixture of log-concave distributions is a very strong assumption. Our results can be generalized to all functions that are ``close'' to a mixture of log-concave distributions.

\ift{\nomenclature[2tau]{$\tau$}{Bound on gradient and Hessian perturbations $\ve{\nb \wt f - \nb f}_\iy\le \tau$, $\ve{\nb^2 \wt f - \nb^2 f}_\iy\le \tau$ (Chapter~\ref{ch:mm})}}
More precisely, assume the function $f$ satisfies the following properties: 
\begin{align}
\exists \tilde{f}: \mathbb{R}^d \to \mathbb{R}
\text{ where } & \ve{\tilde{f} - f}_{\infty} \leq \dellarge \text{   ,   } \ve{\nabla \tilde{f} - \nabla f}_{\infty} \leq \delsmall \text{ and } \|\nabla^2 \tilde{f} - \nabla^2 f\|_2 \leq \delsmall, \forall x \in \mathbb{R}^d \label{eq:A0}\\
\text{and } \tilde{f}(x) &= -\log\left(\sum_{i=1}^m w_i e^{-f_0(x-\mu_i)}\right) \label{eq:tildef}\\
\mbox{where }\nabla f_0(0) &= 0, \text{ and }\forall x, \mineig I \preceq \nabla^2 f_0(x) \preceq \maxeig I.\label{eq:hessianbound} 
\end{align}

That is, $f$ is within a $e^\Delta$ multiplicative factor of an (unknown) mixture of log-concave distributions. Our theorem can be generalized to this case.

\ift{\nomenclature[2Delta]{$\De$}{Perturbation of $f$, $\ve{\wt f-f}_\iy\le \De$ (Chapter~\ref{ch:mm})}}
\begin{thm}[general case] For function $f(x)$ that satisfies Equations \eqref{eq:A0}, \eqref{eq:tildef}, and \eqref{eq:hessianbound}, there is an algorithm (given as Algorithm~\ref{a:mainalgo} with appropriate setting of parameters) that runs in time $\poly\pa{\rc{w_{\min}}, D, d, \frac{1}{\ep}, e^{\dellarge}, \delsmall,\rc{\mineig},\maxeig}$, which given query access to $\nabla f$ and $f$, outputs a sample $x$ from a distribution that has TV-distance at most $\ep$ from $p(x)\propto e^{-f(x)}$.
\label{t:informalperturb}
\end{thm}

Both main theorems may seem simple. In particular, one might conjecture that it is easy to use local search algorithms to find all the modes. However in Section~\ref{sec:examples}, we give a few examples to show that such simple heuristics do not work (e.g. random initialization is not enough to find all the modes). 

The assumption that all the mixture components share the same $f_0$ (hence when applied to Gaussians, all Gaussians have same covariance) is also necessary. In Section~\ref{sec:lb}, we give an example where for a mixture of two gaussians, even if the covariance only differs by a constant factor, any algorithm that achieves similar gaurantees as Theorem~\ref{t:informalperturb} \emph{must} take exponential time. 
The limiting factor is approximately \emph{finding} all the mixture components. We note that when the approximate locations of the mixture are known, there are heuristic ways to temper them differently; see~\cite{tawn2018weight}.


\section{Overview of algorithm}

Our algorithm combines \emph{Langevin diffusion}, a chain for sampling from distributions in the form $p(x)\propto e^{-f(x)}$ given only gradient access to $f$ and \emph{simulated tempering}, a heuristic used for tackling multimodality. 
We briefly define both of these and recall what is known for both of these techniques. For 
technical prerequisites on Markov chains, the reader can refer to Appendix \iftoggle{thesis}{\ref{ch:mc}}{\ref{sec:mc}}.  

The basic idea to keep in mind is the following: A Markov chain with local moves such as Langevin diffusion gets stuck in a local mode. Creating a ``meta-Markov chain'' which changes the temperature (the simulated tempering chain) can exponentially speed up mixing.

\subsection{Langevin dynamics}
\label{sec:overview-l}
Langevin Monte Carlo is an algorithm for sampling from $p(x)\propto e^{-f(x)}$ given access to the gradient of the log-pdf, $\nabla f$. 

The continuous version, overdamped Langevin diffusion (often simply called Langevin diffusion), is a stochastic process described by the stochastic differential equation (henceforth SDE)
\begin{equation}
dX_t = -\nb f (X_t) \,dt + \sqrt{2}\,dW_t \label{eq:langevinsde}
\end{equation}
where $W_t$ is the Wiener process (Brownian motion). 
For us, the crucial 
fact  is that Langevin dynamics converges to the stationary distribution given by $p(x) \propto e^{-f(x)}$. We will always assume that $\int_{\R^d} e^{-f(x)}\dx<\iy$ and $f\in C^2(\R^d)$.

Substituting $\be f$ for $f$ in~\eqref{eq:langevinsde} gives the Langevin diffusion process for inverse temperature $\be$, which has stationary distribution $\propto e^{-\be f(x)}$. Equivalently we can consider the temperature as changing the magnitude of the noise:
$$
dX_t = -\nabla f(X_t)dt + \sqrt{2\beta^{-1}}dW_t.
$$

\ift{\nomenclature[1xi]{$\xi$}{Gaussian noise $\xi\sim N(0,I)$}}
\ift{\nomenclature[1beta]{$\beta$}{Inverse temperature}}
\ift{\nomenclature[1eta]{$\eta$}{Step size}}
Of course algorithmically we cannot run a continuous-time process, so we run a \emph{discretized} version of the above process: namely, we run a Markov chain where the random variable at time $t$ is described as 
\begin{equation} 
X_{t+1} = X_t - \eta \nb f(X_t)  + \sqrt{2 \eta }\xi_k, \quad \xi_k \sim N(0,I) \label{eq:langevind} 
\end{equation}
where $\eta$ is the step size. (The reason for the $\sqrt \eta$ scaling is that running Brownian motion for $\eta$ of the time scales the variance by $\sqrt{\eta}$.) This is analogous to how gradient descent is a discretization of gradient flow.


\subsubsection{Prior work on Langevin dynamics} 

For Langevin dynamics, convergence to the stationary distribution is a classic result \citep{bhattacharya1978criteria}. Fast mixing for log-concave distributions is also a classic result: \cite{bakry1985diffusions, bakry2008simple} show that log-concave distributions satisfy a Poincar\'e and log-Sobolev inequality, which characterize the rate of convergence---If $f$ is $\alpha$-strongly convex, then the mixing time is on the order of $\frac 1{\alpha}$. 
Of course, algorithmically, one can only run a ``discretized'' version of the Langevin dynamics. Analyses of the discretization are more recent: \cite{dalalyan2016theoretical, durmus2016high,dalalyan2017further,dalalyan2017user,cheng2017convergence,durmus2018analysis} give running times bounds for sampling from a log-concave distribution over $\mathbb{R}^d$, and \cite{bubeck2018sampling} give a algorithm to sample from a log-concave distribution restricted to a convex set by incorporating a projection.
We note these analysis and ours are for the simplest kind of Langevin dynamics, the overdamped case; better rates are known for underdamped dynamics~(\cite{cheng2017underdamped}), if a Metropolis-Hastings rejection step is used~(\cite{dwivedi2018log}), and for Hamiltonian Monte Carlo which takes into account momentum~(\cite{mangoubi2017rapid}).

\cite{raginsky2017non,cheng2018sharp,vempala2019rapid} carefully analyze the effect of discretization for arbitrary non-log-concave distributions with certain regularity properties, 
but the mixing time is exponential in general; furthermore, it has long been known that transitioning between different modes can take exponentially long, a phenomenon known as meta-stability \citep{bovier2002metastability, bovier2004metastability, bovier2005metastability}. The Holley-Stroock Theorem (see e.g. \citep{bakry2013analysis}) shows that guarantees for mixing extend to distributions $e^{-f(x)}$ where $f(x)$ is a ``nice'' function that is close to a convex function in $L^\iy$ distance; however, this does not address more global deviations from convexity. \cite{mangoubi2017convex} consider a more general model with multiplicative noise.

\subsection{Simulated tempering}

\label{sec:overview-st}

For distributions that are far from being log-concave and have many deep modes, additional techniques are necessary. One proposed heuristic, out of many, is simulated tempering, which swaps between Markov chains that are different temperature variants of the original chain. The intuition is that the Markov chains at higher temperature can move between modes more easily, and hence, the higher-temperature chain acts as a ``bridge'' to move between modes.

Indeed, Langevin dynamics corresponding to a higher temperature distribution---with $\beta f$ rather than $f$, where $\beta<1$---mixes faster. 
(Here, we use terminology from statistical physics, letting $\tau$ denote teh temperature and $\be=\rc\tau$ denote the inverse temperature.)
A high temperature flattens out the distribution.
However, we can't simply run Langevin at a higher temperature because the stationary distribution is wrong; the simulated tempering chain combines Markov chains at different temperatures in a way that preserves the stationary distribution.

\ift{\nomenclature[1Om]{$\Om$}{State space}}
\ift{\nomenclature[2L]{$L$}{Number of temperatures (Chapter~\ref{ch:mm})}}
We can define simulated tempering with respect to any sequence of Markov chains $M_i$ on the same space $\Om$. Think of $M_i$ as the Markov chain corresponding to temperature $i$, with stationary distribution $e^{-\be_i f}$.

Then we define the simulated tempering Markov chain as follows.
\begin{itemize}
\item
The \emph{state space} is $[L]\times \Om$: $L$ copies of the state space (in our case $\mathbb R^d$), one copy for each temperature.
\item The evolution is defined as follows.
\begin{enumerate}
\item If the current point is $(i,x)$, then evolve according to the $i$th chain $M_i$.
\item Propose swaps with some rate $\lambda$. When a swap is proposed, attempt to move to a neighboring chain, $i'=i\pm 1$. With probability $\min\{p_{i'}(x)/p_i(x), 1\}$, the transition is successful. Otherwise, stay at the same point. This is a \emph{Metropolis-Hastings step}; its purpose is to preserve the stationary distribution.\footnote{
This can be defined as either a discrete or continuous Markov chain. 
For a discrete chain, we propose a swap with probability $\lambda$ and follow the current chain with probability $1-\lambda$. For a continuous chain,  the time between swaps is an exponential distribution with decay $\lambda$ (in other words, the times of the swaps forms a Poisson process).
Note that simulated tempering is traditionally defined for discrete Markov chains, but we will use the continuous version. See Definition~\ref{df:cst} for the formal definition.
}
\end{enumerate}
\end{itemize}

The crucial fact to note is that the stationary distribution is a ``mixture'' of the distributions corresponding to the different temperatures. Namely:  

\begin{pr}\cite{marinari1992simulated,neal1996sampling}
If the $M_{k}$, $1\le k\le L$ are reversible Markov chains with stationary distributions $p_k$, then the simulated tempering chain $M$
is a reversible Markov chain with stationary distribution
$$
p(i,x) = 
\rc L p_i(x).
$$
\end{pr}

The typical setting of simulated tempering is as follows. The Markov chains come from a smooth family of Markov chains with parameter $\be\ge 0$, and $M_i$ is the Markov chain with parameter $\be_i$, where $0\le \be_1\le \ldots \le \be_{L}=1$. 
We are interested in sampling from the distribution when $\be$ is large ($\tau$ is small). However, the chain suffers from torpid mixing in this case, because the distribution is more peaked. The simulated tempering chain uses smaller $\be$ (larger $\tau$) to help with mixing.
For us, the stationary distribution at inverse temperature $\be$ is $\propto e^{-\be f(x)}$.

\subsubsection{Prior work on simulated tempering}

Provable results of this heuristic are few and far between.  
\cite{woodard2009conditions, zheng2003swapping} lower-bound the spectral gap for generic simulated tempering chains, using a Markov chain decomposition technique due to \cite{madras2002markov}. However, for the Problem~\ref{p:main} that we are interested in, the spectral gap bound in \cite{woodard2009conditions} is exponentially small as a function of the number of modes. 
Drawing inspiration from~\cite{madras2002markov}, we establish a Markov chain decomposition technique that overcomes this. 

One issue that comes up in simulated tempering is estimating the partition functions; various methods have been proposed for this~\cite{park2007choosing,liang2005determination}.

\subsection{Main algorithm}

Our algorithm is intuitively the following. Take a sequence of inverse temperatures $\be_i$, starting at a small value and increasing geometrically towards 1. Run simulated tempering Langevin on these temperatures, suitably discretized.
Take the samples that are at the $L$th temperature.

Note that there is one complication: the standard simulated tempering chain assumes that we can compute the ratio between temperatures $\fc{p_{i'}(x)}{p_i(x)}$. However, we only know the probability density functions up to a normalizing factor (the partition function). To overcome this, we note that if we use the ratios $\fc{r_{i'}p_{i'}(x)}{r_i p_i(x)}$ instead, for $\sumo iL r_i=1$, then the chain converges to the stationary distribution with $p(x,i) = r_i p_i(x)$. Thus, it suffices to estimate each partition function up to a constant factor. We can do this inductively: running the simulated tempering chain on the first $\ell$ levels, we can estimate the partition function $Z_{\ell+1}$; then we can run the simulated tempering chain on the first $\ell+1$ levels. This is what Algorithm~\ref{a:mainalgo} does when it calls Algorithm~\ref{a:stlmc} as subroutine.

 A formal description of the algorithm follows.

\ift{\nomenclature[2Zhat]{$Z_i$}{Partition function of $i$th temperature, $\int_{\R^d} e^{-\beta_i f(x)}\dx$ (Chapter~\ref{ch:mm})}}
\ift{\nomenclature[2Zhati]{$\wh Z_i$}{Estimate of partition function $Z_i$ (Chapter~\ref{ch:mm})}}
\begin{algorithm}[h!]
\begin{algorithmic}
\State INPUT: Temperatures $\be_1,\ldots, \be_\ell$; partition function estimates $\wh Z_1,\ldots, \wh Z_\ell$; step size $\eta$, time $T$, rate $\la$, variance of initial distribution $\si_0$.
\State OUTPUT: A random sample $x\in \R^d$ (approximately from the distribution $p_\ell(x)\propto e^{-\be_\ell f(x)}$).
\State Let $(i,x)=(1,x_0)$ where $x_0\sim N(0, \si_0^2 I )$.
\State Let $n=0, T_0=0$.
\While{$T_n<T$}
	\State Determine the next transition time: Draw $\xi_{n+1}$ from the exponential distribution $p(x) = \la e^{-\la x}$, $x\ge 0$. 
	\State 	Let $\xi_{n+1}\mapsfrom \min\{T-T_n, \xi_{n+1}\}$, $T_{n+1}=T_n + \xi_{n+1}$.
	\State Let $\eta' = \xi_{n+1}/\ce{\fc{\xi_{n+1}}{\eta}}$ (the largest step size $<\eta$ that evenly divides into $\xi_{n+1}$).
	\State Repeat $\ce{\fc{\xi_{n+1}}{\eta}}$ times: Update $x$ according to $x \mapsfrom x - \eta' \be_i\nb f(x) +\sqrt{2\eta'}\xi$, $\xi\sim N(0,I)$. 
\State If $T_{n+1}<T$ (i.e., the end time has not been reached), let $i'=i\pm 1$ with probability $\rc 2$. If $i'$ is out of bounds, do nothing. If $i'$ is in bounds, make a type 2 transition, where the acceptance ratio is 
$\min \bc{
\fc{e^{-\be_{i'}f(x)}/\wh Z_{i'}}
{e^{-\be_{i}f(x)}/\wh Z_{i}}, 1
}
$.
\State $n\mapsfrom n+1$.
\EndWhile
\State If the final state is $(\ell,x)$ for some $x\in \R^d$, return $x$. Otherwise, re-run the chain.
\end{algorithmic}
 \caption{Simulated tempering Langevin Monte Carlo}
 \label{a:stlmc}
\end{algorithm}

\begin{algorithm}[h!]
\begin{algorithmic}
\State INPUT: A function $ f: \mathbb{R}^d$, satisfying assumption~\eqref{eq:A0}, to which we have gradient access.  
\State OUTPUT: A random sample $x \in \mathbb{R}^d$. 
\State Let $0\le \be_1<\cdots < \be_L=1$ be a sequence of inverse temperatures satisfying~\eqref{eq:beta1} and~\eqref{eq:beta-diff}. 
\State Let $\wh Z_1=1$.
\For{$\ell = 1 \to L$}  
 \State Run the simulated tempering chain in Algorithm~\ref{a:stlmc} with temperatures 
$\be_1,\ldots, \be_{\ell}$, partition function estimates $\wh Z_1,\ldots, \wh Z_{\ell}$, step size $\eta$, time $T$, and rate $\la$ given by Lemma~\ref{lem:a1-correct}. 
 \State If $\ell=L$, return the sample.
 \State If $\ell<L$, repeat to get $n=O(L^2\ln \prc{\de})$ samples, and let $\wh{Z_{\ell+1}} = \wh{Z_\ell} \pa{
 \rc n \sumo jn e^{(-\be_{\ell+1}+\be_\ell)f(x_j)}}$.
\EndFor
\end{algorithmic}
 \caption{Main algorithm}
\label{a:mainalgo}
\end{algorithm}

\section{Overview of the proof techniques} 

We summarize the main ingredients and crucial techniques in the proof. Full proofs appear in the following sections.  \\

\textbf{Step 1:} Define a continuous version of the simulated tempering Markov chain  (Definition~\ref{df:cst}, Lemma~\ref{lem:st-generator}), where transition times are real numbers determined by an exponential weighting time distribution.  \\

\textbf{Step 2:} Prove a new \emph{decomposition theorem} (Theorem~\ref{thm:gap-prod-st}) for bounding the spectral gap (or equivalently, the mixing time) of the simulated tempering process we define. This is the main technical ingredient, and also a result of independent interest.

While decomposition theorems have appeared in the Markov chain literature (e.g. \cite{madras2002markov}), typically one partitions the \emph{state space}, and bounds the spectral gap using (1) the probability flow of the chain inside the individual sets, and (2) between different sets. 

In our case, we decompose the \emph{Markov process} itself; this includes a decomposition of the stationary distribution into components.  (More precisely, we decompose the generator of the process.) We would like to do this because in our setting, the stationary distribution is exactly  a mixture distribution (Problem~\ref{p:main}).

Our Markov process decomposition theorem bounds the spectral gap (mixing time) of a simulated tempering chain in terms of the spectral gap (mixing time) of two processes:
\begin{enumerate}
\item
``component' processes on the mixture components
\item
a ``projected'' process whose state space is the set of components, and 
which captures the action of the chain between components as well as the 
distance between the mixture components (measured in terms of their overlap) 
\end{enumerate}
This means that if the Markov process on the individual components mixes rapidly, and the ``projected'' process mixes rapidly, then the simulated tempering process mixes rapidly as well.
(Note \cite[Theorem 1.2]{madras2002markov} does partition into mixture components, but they only consider the special case where they components are laid out in a chain.) 

The mixing time of a Markov process is quantified by a Poincar\'e inequality.
\begin{thm*}[Simplified version of Theorem~\ref{thm:gap-prod-st}]
Consider the simulated tempering process $M$ with rate $\lambda = \rc{C}$, where the Markov process at the $i$th level (temperature) is $M_i = (\Om, \sL_i)$ with stationary distribution $p_i$, for $1\le i\le L$.
Suppose we have a decomposition of the Markov process at each level, $p_iM_i = \sumo jm w_{i,j} p_{i,j} M_{i,j}$, where $\sumo jm w_{i,j}=1$. 
If each $M_{i,j}$ satisfies a Poincar\'e inequality with constant $C$, and the projected chain $\ol M$  satisfies a Poincar\'e inequality with constant $\ol C$, then $M$ satisfies a Poincar\'e inequality with constant $O(C(1+\ol C))$. 

Here, the projected process $\ol M$ is the chain on $[L]\times [m]$ with probability flow in the same and adjacent levels given by 
\begin{align}
\ol T((i,j), (i,j')) &= w_{i,j'}\de_{(i,j),(i,j')}\\
\ol T((i,j), (i\pm 1,j)) &= \de_{(i,j),(i\pm 1,j)}
\end{align}
where $\de_{(i,j),(i',j')}:=\int_{\Om} \min\{p_{i,j}(x),p_{i',j'}(x)\}\dx$ is the overlap.
\end{thm*}

The decomposition theorem is the reason why we use a slightly different simulated tempering process, which is allowed to transition at arbitrary times, with some rate $\lambda$. Such a process ``composes'' nicely with the decomposition of the Langevin chain, and allows a better control of the Dirichlet form of the tempering process, which governs the mixing time.\\

\textbf{Step 3:} Finally, we need to apply the decomposition theorem to our setup, namely a distribution which is a mixture of strongly log-concave distributions. The ``components'' of the decomposition in our setup are simply the mixture components $e^{-f_0(x-\mu_j)}$. We rely crucially on the fact that Langevin diffusion on a mixture distribution decomposes into Langevin diffusion on the individual components.

We actually first analyze the \emph{hypothetical} simulated tempering Langevin process on $\wt p_i \propto \sumo jm w_j e^{-\be_jf_0(x-\mu_j)}$ (Theorem~\ref{thm:st-gaussians})---i.e., where the stationary distribution for each temperature is a mixture. Then in Lemma~\ref{lem:st-comparison} we compare to the \emph{actual} simulated tempering Langevin that we can run, where $p_i\propto p^{\be}$. To do this, we use the fact that $p_i$ is off from $\wt p_i$ by at most $\rc{w_{\min}}$. (This is the only place where a factor of $w_{\min}$ comes in.)

To use our Markov process decomposition theorem, we need to show two things:
\begin{enumerate}
\item
The component processes mix rapidly: this follows from the classic fact that Langevin diffusion mixes rapidly for log-concave distributions.
\item
The projected process mixes rapidly: The ``projected'' process is defined as having more probability flow between mixture components in the same or adjacent temperatures which are close together in $\chi^2$-divergence. 

By choosing the temperatures close enough, we can ensure that the corresponding mixture components in adjacent temperatures are close (in the sense of having high overlap). 
By choosing the highest temperature large enough, we can ensure that all the mixture components at the highest temperature are close. 

From this it follows that we can easily get from any component to any other (by traveling up to the highest temperature and then back down). Thus the projected process mixes rapidly from the method of canonical paths, Theorem~\ref{thm:can-path}.
\end{enumerate}
Note that the equal variance (for gaussians) or shape (for general log-concave distributions) condition is necessary here. For gaussians with different variance, the Markov process can fail to mix between components at the highest temperature. This is because scaling the temperature changes the variance of all the components equally, and preserves their ratio (which is not equal to 1).\\

\textbf{Step 4:} We analyze the error from discretization (Lemma~\ref{l:maindiscretize}), and choose parameters so that it is small. We show that in Algorithm~\ref{a:mainalgo} we can inductively estimate the partition functions. When we have all the estimates, we can run the simulated tempering chain on all the temperatures to get the desired sample.

\subsection{Comparison to our previous algorithm and proof}
\label{s:comparison}
We make some comparisons to our previous work~\cite{ge2017beyond} that addresses the same problem. Note that the proof was given for mixtures of  gaussians, but extends in a straightforward way to mixtures of log-concave distributions.
The main difference in the algorithm of~\cite{ge2017beyond} is that transitions between temperatures only happen at fixed times---after a certain number of steps of running the Markov chain at the current temperature.  The main difference from the proof in~\cite{ge2017beyond} is that it uses Markov chain decomposition theorem of~\cite{madras2002markov}, which requires partitioning the state space into disjoint \emph{sets} on which the restricted chain mixes well. Coming up with the partition requires an intricate argument relying on a spectral partitioning theorem for graphs given by~\cite{gharan2014partitioning}. Roughly, it says that if the $(m+1)$th eigenvalue of a Markov chain is bounded away from 0 (as is the case for Langevin diffusion on a mixture of $m$ log-concave distributions), then we can find a partition into $\le m$ sets, with good internal conductance and poor external conductance. 
However, since the theorem holds for discrete-time, discrete-space Markov chains, to use the theorem we need some technical discretization arguments.
Since ultimately we care about the spectral gap, we have to bound the spectral gap by the conductance, and lose a square by Cheeger's inequality. In this paper, we obtain better bounds by circumventing this issue with a soft decomposition theorem. We also circumvent the technical discretization arguments by working with Poincar\'e inequalities, which apply directly to the continuous chain.

Ignoring logarithmic factors and focusing on the dependence on $d$ (dimension), $\ep$, $m$ (number of components), and $w_{\min}$ (minimum weight of component), in~\cite{ge2017beyond}, the number of temperatures required is $L = \wt \Te(d)$, the amount of time to simulate the Markov chain is $t = \wt \Te\pf{L^4m^{16}}{w_{\min}^4} =\wt \Te\pf{d^4m^{16}}{w_{\min}^4}$, and the step size is $\eta = \wt \Te\pf{\ep^2}{dt}$\footnote{An error in the previous paper displayed the dependence of $\eta$ on $\ep$ to be $\ep$ rather than $\ep^2$.}, so the total amount of steps to run the Markov chain, once the partition function estimates are known, is
$\fc{t}{\eta} = \wt \Te\pf{t^2 d}{\ep^2} = \wt \Te\pf{d^9m^{32}}{\ep^2 w_{\min}^8}$.

In this paper, examining the parameters in Lemma~\ref{lem:a1-correct}, the number of temperatures required is $L = \wt \Te(d)$, the amount of time to simulate the Markov chain is $T = \wt \Te\pf{L^2}{w_{\min}^{\wexp}}$, the step size is $\eta = \wt \Te\pf{\ep^2}{dT}$, so the total amount of steps is $\fc{T}{\eta} = \wt \Te\pf{T^2 d}{\ep^2} = \wt \Te\pf{d^5}{\ep^2 w_{\min}^{\wexpt}}$. 
Note that in either case, to obtain the actual complexity, we need to additionally multiply by a factor of $L^4 = \wt \Te(d^4)$: one factor of $L$ comes from needing to estimate the partition function at each temperature, a factor of $L^2$ comes from the fact that we need $L^2$ samples at each temperature to do this,  and the final factor of $L$ comes from the fact that we reject the sample if the sample is not at the final temperature. (We have not made an effort to optimize this $L^4$ factor.)
\section{Theorem statements}
\label{s:thm}
We restate the main theorems more precisely. First define the assumptions.
\begin{asm}\label{asm}
The function $f$ satisfies the following. There exists a function $\tilde{f}$ that satisfies the following properties.
\begin{enumerate}
\item
$\tilde{f}$, $\nb \tilde{f}$, and $\nb^2 \tilde f$ are close to $f$:
\begin{gather}
 \ve{\tilde{f} - f}_{\infty} \leq \dellarge \text{   ,   } \ve{\nabla \tilde{f} - \nabla f}_{\infty} \leq \delsmall \text{ and } \nabla^2 \tilde{f}(x) \preceq \nabla^2 f(x) + \delsmall I, \forall x \in \mathbb{R}^d \label{eq:A0-p}
\end{gather}
\item
$\tilde f$ is the log-pdf of a mixture:
\begin{align}
\tilde{f}(x) &= -\log\left(\sum_{i=1}^m w_i e^{-f_0(x-\mu_i)}\right) \label{eq:tildef-p}
\end{align}
where $\nb f_0(0)=0$ and
\begin{enumerate}
\item
$f_0$ is $\mineig$-strongly convex: $\nb^2 f_0(x)\succeq \mineig I$ for $\mineig>0$.
\item
$f_0$ is $\maxeig$-smooth: $\nb^2 f_0(x)\preceq \maxeig I$.
\end{enumerate}
\end{enumerate}
\end{asm}

Our main theorem is the following. 
\begin{thm}[Main theorem, Gaussian version]\label{thm:main}
Suppose \begin{align*}f(x) = -\ln \pa{\sumo jm w_j \exp\pa{-\fc{\ve{x-\mu_j}^2}{2\si^2}}}\end{align*} on $\R^d$ where $\sumo jm w_j=1$, $w_{\min}=\min_{1\le j\le m}w_j>0$, and $D=\max_{1\le j\le m}\ve{\mu_j}$. Then 
Algorithm~\ref{a:mainalgo} with parameters 
satisfying $t, T, \eta^{-1}, \be_1^{-1}, (\be_i-\be_{i-1})^{-1} = 
\poly\pa{\rc{w_{\min}},D,d, \rc{\sigma^2},\rc{\ep}}$
produces a sample from a distribution $p'$ with $\ve{p-p'}_1\le \ep$ in time $\poly\pa{\rc{w_{\min}}, D, d, \rc{\si^2}, \frac{1}{\ep}}$.
\end{thm}
The precise parameter choices are given in Lemma~\ref{lem:a1-correct}.
\ift{Examining the parameters, the number of temperatures required is $L = \wt \Te(d)$, the amount of time to simulate the Markov chain is $T = \wt \Te\pf{L^2}{w_{\min}^{\wexp}}$, the step size is $\eta = \wt \Te\pf{\ep^2}{dT}$, so the total amount of steps is $\fc{T}{\eta} = \wt \Te\pf{T^2 d}{\ep^2} = \wt \Te\pf{d^5}{\ep^2 w_{\min}^{\wexpt}}$. 
Note that to obtain the actual complexity, we need to additionally multiply by a factor of $L^4 = \wt \Te(d^4)$: one factor of $L$ comes from needing to estimate the partition function at each temperature, a factor of $L^2$ comes from the fact that we need $L^2$ samples at each temperature to do this,  and the final factor of $L$ comes from the fact that we reject the sample if the sample is not at the final temperature. (We have not made an effort to optimize this $L^4$ factor.)}

Our more general theorem allows the mixture component to come from an arbitrary log-concave distribution $p(x)\propto e^{-f_0(x)}$.
\begin{thm}[Main theorem]\label{thm:mainlogconcave}
Suppose $p(x) \propto e^{-f(x)}$ and $f(x) = -\ln \pa{\sumo im w_i e^{-f_0(x-\mu_i)}}$ on $\R^d$, where function $f_0$ 
satisfies Assumption~\ref{asm}(2) ($f_0$ is $\mineig$-strongly convex, $\maxeig$-smooth, and has minimum at 0), $\sumo im w_i=1$, $w_{\min}=\min_{1\le i\le m}w_i>0$, and $D=\max_{1\le i\le n}\ve{\mu_i}$. Then 
Algorithm~\ref{a:mainalgo} with parameters
satisfying $t, T, \eta^{-1}, \be_1^{-1}, (\be_i-\be_{i-1})^{-1} = 
\poly\pa{\rc{w_{\min}},D,d,\rc{\ep}, \rc{\mineig},\maxeig}$
produces a sample from a distribution $p'$ with $\ve{p-p'}_1\le \ep$ in time $\poly\pa{\rc{w_{\min}},D,d,\rc{\ep}, \rc{\mineig},\maxeig}$.
\end{thm}
The precise parameter choices are given in Lemma~\ref{lem:a1-correct-gen}.

\begin{thm}[Main theorem with perturbations]\label{thm:perturb}
Keep the setup of Theorem~\ref{thm:mainlogconcave}. 
If instead $f$ satisfies Assumption~\ref{asm} ($f$ is $\De$-close in $L^\iy$ norm to the log-pdf of a mixture of log-concave distributions), then the result of Theorem~\ref{thm:mainlogconcave} holds with an additional factor of $\poly(e^{\De},\tau)$ in the running time. 
\end{thm}

\section{Simulated tempering}
\label{sec:st}

\ift{\nomenclature[1Lscript]{$\sL$}{Generator of Markov process}}
First we define a continuous version of the simulated tempering Markov chain (Definition~\ref{df:cst}). Unlike the usual definition of a simulated tempering chain in the literature, the transition times can be arbitrary real numbers. Our definition falls out naturally from writing down the generator $\sL$ as a combination of the generators for the individual chains and for the transitions between temperatures (Lemma~\ref{lem:st-generator}). Because $\sL$ decomposes in this way, the Dirichlet form $\sE$ will be easier to control in Theorem~\ref{thm:gap-prod-st}. 
%

\ift{\nomenclature[2ri]{$r_i$}{Relative probabilities  (Chapter~\ref{ch:mm})}}
\ift{\nomenclature[2lambda]{$\la$}{Rate of simulated tempering (Chapter~\ref{ch:mm})}}
\ift{\nomenclature[2Mst]{$M_{\text{st}}$}{Simulated tempering chain  (Chapter~\ref{ch:mm})}}
\begin{df}\label{df:cst}
Let $M_i, i\in [L]$ be a sequence of continuous Markov processes with state space $\Om$ with stationary distributions $p_i(x)$ (with respect to a reference measure). Let $r_i$, $1\le i\le L$  satisfy
$$
r_i> 0,\quad \sumo i{L} r_i = 1.
$$
Define the \vocab{continuous simulated tempering Markov process} $M_{\st}$ with \emph{rate} $\la$ and \emph{relative probabilities} $r_i$ as follows. 

The states of $M_{\st}$ are $[L]\times \Om$.

For the evolution, let $(T_n)_{n\ge 0}$ be a Poisson point process on $\R_{\ge0}$ with rate $\la$, i.e., $T_0=0$ and 
\begin{align*}
T_{n+1}-T_n | T_1,\ldots, T_n &\sim \Exp(\la)
\end{align*}
with probability density $p(t)=\la e^{-\la t}$.
If the state at time $T_n$ is $(i,x)$, then the Markov process evolves according to $M_i$ on the time interval $[T_n,T_{n+1})$. The state $X_{T_{n+1}}$ at time $T_{n+1}$ is obtained from the state $X_{T_{n+1}}^- := \lim_{t\to T_{n+1}^-} X_t$ by a ``Type 2'' transition: If $X_{T_{n+1}}^- = (i,x)$, then transition to $(j=i\pm 1, x)$ each with probability $$
\rc2\min \bc{\fc{r_{j}p_{j}(x)}{r_ip_i(x)}, 1}
$$
and stay at $(i,x)$ otherwise. (If $j$ is out of bounds, then don't move.)
\end{df}

\begin{lem}\label{lem:st-generator}
Let $M_i, i\in [L]$ be a sequence of continuous Markov processes with state space $\Om$, generators $\sL_i$ (with domains $\cal D(\sL_i)$), and  unique stationary distributions $p_i$. Then the {continuous simulated tempering Markov process} $M_{\st}$ with {rate} $\la$ and {relative probabilities} $r_i$ has generator $\sL$ defined by the following equation, where $g=(g_1,\ldots, g_L)\in  \prodo iL \cal D(\sL_i)$: 
\begin{align*}
(\sL g)(i,y) &=(\sL_i g_i)(y) +
\fc \la 2 
\pa{
\sumr{1\le j\le L}{j=i\pm 1}
\min\bc{\fc{r_jp_j(x)}{r_i p_i(x)},1} (g_j(x)-g_i(x))
}.
\end{align*}
\end{lem}

\ift{\nomenclature[1Escript]{$\sE$}{Dirichlet form}}
The corresponding Dirichlet form is
\begin{align}\label{eq:dir-st}
&\sE(g, g) =-\an{g,\sL g}\\
\nonumber
&= 
\sumo iL r_i\sE_i(g_i,g_i)
+\fc \la 2 \sumr{1\le i,j\le L}{j=i\pm 1}
\int_{\Om} r_ip_i(x) \min\bc{\fc{r_jp_j(x)}{r_i p_i(x)},1}(g_i(x)^2-g_i(x)g_j(x))\dx\\
\nonumber
&= 
\sumo iL r_i\sE_i(g_i,g_i)
+\fc \la 4 \sumr{1\le i,j\le L}{j=i\pm 1}
\int_{\Om} (g_i-g_j)^2 \min\{r_ip_i,r_jp_j\}\,dx
\end{align}
where $\sE_i(g_i,g_i)= -\an{g_i,\sL_ig_i}_{P_i}$.

\begin{proof}
Continuous simulated tempering is a Markov process because the Poisson process is memoryless. We show that its generator equals $\sL$. Let $F$ be the operator which acts by
\begin{align*}
Fg(x,i) &= g_i(x) +\rc 2\sumr{1\le j\le L}{j=i\pm 1}
\min\bc{\fc{r_jp_j(x)}{r_{i} p_{i}(x)},1}(g_{j}(x)-g_i(x))
\end{align*}
Let $N_t=\max\set{n}{T_n\le t}$.
Let $P_{j,t}$ be such that $(P_{j,t}g)(x)=\EE_{M_j} [g(x_t)|x_0=x]$, the expected value after running $M_j$ for time $t$, and let $P_t$ the same operator for $M$.

We have, letting $P_s'=\sumo jL \de_j \times P_{j,s}$ (where $\de_j(i) = \one_{i=j}$ is a function on $[L]$),
\begin{align*}
P_t g &= \Pj(N_t=0) \sumo jL\de_j \times P_{j,t} g_j + \int_0^t 
 P_{s}'FP_{t-s}' g \Pj(t_1=ds,N_t=1)
  + \Pj(N_t\ge 2)h.
\end{align*}
where $\ve{h}_{L^2(P)}\le \ve{g}_{L^2(P)}$ (by contractivity of Markov processes). Here, $P_s'FP_{t-s}'$ comes from moving for time $s$ at one level, doing a level change, then moving for time $t-s$ on the new level.
By basic properties of the Poisson process, $\Pj(N_t=0) = 1-\la t + O(t^2)$, $\Pj(t_1=s,N_t=1) = \la+ O(t)$ for $0\le s\le t$, and $\Pj(N\ge 2) = O(t^2)$, so 
\begin{align*}
\ddd t(P_t g)|_{t=0} &= -\la\ub{\sumo jL\de_j \times P_{j,t} g_j}{g} + \sumo jL \de_j\times \sL_j g_j +\la  Fg
=\sL g.
\end{align*}
\end{proof}

\section{Markov process decomposition theorems}

For ease of reading, we first prove a simple density decomposition theorem, Theorem~\ref{thm:gap-prod} (which will not be needed, but gives the main idea in a simpler setting). Then we prove the density decomposition theorem for simulated tempering, Theorem~\ref{thm:gap-prod-st}, which is the density decomposition theorem that we use to prove the main Theorem~\ref{thm:main}.

Both of these theorems are consequences of a more general decomposition theorem, Theorem~\ref{t:gen-decomp} (up to constants). In Appendix~\ref{a:decomp} we prove the general theorem and show how to specialize it to the case of simulated tempering to recover Theorem~\ref{thm:gap-prod-st}. We also give a version of the theorem for a continuous index set, Theorem~\ref{t:decomp-cts}.

We compare Theorems~\ref{thm:gap-prod} and~\ref{t:gen-decomp} to decomposition theorems in the literature,~\cite[Theorem 1.1, 1.2]{madras2002markov} and~\cite[Theorem 5.2]{woodard2009conditions}. Note that our theorems are stated for continuous-time Markov processes, while the others are stated for discrete-time; however, either proof could be adapted for the other setting. 
\begin{itemize}
\item
In Theorem~\ref{thm:gap-prod} we use the Poincar\'e constants of the component Markov processes, and the distance of their stationary distributions to each other, to bound the Poincar\'e constant of the original chain. (Theorem~\ref{thm:gap-prod} gives a bound in terms of the $\chi^2$ divergences, but Remark~\ref{r:decomp-simple} gives the bound in terms of the ``overlap" quantity which is used in the literature.)

This is a generalization of \cite[Theorem 1.2]{madras2002markov}, which deals with the special case where the state space is partitioned into overlapping sets, and \cite[Theorem 1.2]{madras2002markov}, which covers the special case where the component distributions are laid out in a chain. Our theorem can deal with any arrangement.
\item
In Theorem~\ref{t:gen-decomp} we additionally use the ``probability flow" \emph{between} components to get a more general bound. It involves partitioning the pairs of indices $I\times I$ into $S_{\cil}$ and $S_{\lra}$, where to get a good bound, one puts $(i,j)$ where $p_i$ and $p_j$ are close into $S_{\cil}$, and $(i,j)$ where there is a lot of ``probability flow" between $p_i$ and $p_j$ into $S_{\lra}$. Theorem~\ref{thm:gap-prod} is the special case of Theorem~\ref{t:gen-decomp} when $S_{\lra}=\phi$.

Note that \cite[Theorem 5.2]{woodard2009conditions} is similar to the case where $S_{\cil}=\phi$. However, they depend only on the probability flow, while we depend on the ``overlap" in the probability flow.
\end{itemize}
%
\subsection{Simple density decomposition theorem}

\ift{\nomenclature[1chi2]{$\chi^2(P||Q)$}{$\chi^2$-divergence, $\chi^2(P||Q)=\int_{\Om} \pa{\dd QP}P(dx)$}}
\ift{\nomenclature[1chi2max]{$\chi^2_{\max}(P||Q)$}{$\max\bc{\chi^2(P||Q),\chi^2(Q||P)}$}}
\ift{\nomenclature[1VarP]{$\Var_P$}{Variance with respect to $P$}}
\ift{\nomenclature[2Mbar]{$\ol M$}{Projected Markov process (Chapter~\ref{ch:mm})}}
\ift{\nomenclature[2Pbar]{$\ol P$}{Stationary measure of projected Markov process (Chapter~\ref{ch:mm})}}
\ift{\nomenclature[2Tbar]{$\ol T$}{Transition probabilities of projected Markov process (Chapter~\ref{ch:mm})}}
\ift{\nomenclature[2Lscriptbar]{$\sL$}{Generator of projected Markov process (Chapter~\ref{ch:mm})}}
\ift{\nomenclature[2Cbar]{$\ol C$}{Poincar\'e constant for projected Markov process (Chapter~\ref{ch:mm})}}
\begin{thm}[Simple density decomposition theorem]
\label{thm:gap-prod}
Let $M=(\Om, \sL)$ be a (continuous-time) Markov process with stationary measure $P$ and Dirichlet form $\sE(g,g)=-\an{g,\sL g}_P$. Suppose the following hold.
\begin{enumerate}
\item
There is a decomposition
\begin{align*}
\an{f,\sL g}_P &= \sumo jm w_j \an{f, \sL_j g}_{P_j}\\
P&= \sumo jm w_jP_j.
\end{align*}
where $\sL_j$ is the generator for some Markov chain $M_j$ on $\Om$ with stationary distribution $P_{j}$.
\item
(Mixing for each $M_j$) The Dirichlet form $\sE_j(f,g) = -\an{f,\sL_j g}_{P_j}$ satisfies the Poincar\'e inequality
\begin{align*}
\Var_{P_j}(g) &\le C \sE_j(g,g).
\end{align*}
\item
(Mixing for projected chain) 
Define the 
projected process $\ol M$ as the Markov process on $[m]$ generated by $\ol{\sL}$, where $\ol{\sL}$ acts on $\ol g\in L^2([m])$ by\footnote{$\ol M$ is defined so that the rate of diffusion from $j$ to $k$ is given by $\ol T(j,k)$.}
\begin{align*}
\ol{\sL} \ol g(j) &= \sum_{1\le k\le m, k\ne j} [\ol g(k)-\ol g(j)]
\ol T(j,k)\\
\text{where }
\ol T(j,k) &= \fc{w_k}{\chi^2_{\max}(P_j||P_k)}
\end{align*}
where $\chi^2_{\max}(P||Q):=\max\{\chi^2(P||Q),\chi^2(Q||P)\}$. (Define $\ol T(j,k)=0$ if this is infinite.)
Let $\ol P$ be the stationary distribution of $\ol M$; $\ol M$ satisfies the Poincar\'e inequality
\begin{align*}
\Var_{\ol P} (\ol g) & \le \ol C \ol{\sE}(g,g).
\end{align*}
\end{enumerate}
Then $M$ satisfies the Poincar\'e inequality
\begin{align}\label{e:simple-pi}
\Var_P(g) &\le C \pa{1+\fc{\ol C}2} \sE(g,g).
\end{align}
\end{thm}

\ift{\nomenclature[2Qjk]{$Q_{j,k}$}{Minimum of probability measures $P_j$, $P_k$: $\min\bc{\dd{P_k}{P_j},1}\,P_j$ (Chapter~\ref{ch:mm})}}
\ift{\nomenclature[2deltajk]{$\de_{j,k}$}{Overlap between measures $P_j$, $P_k$: $\int_{\Om}\min\bc{\dd{P_k}{P_j},1}\,P_j(dx)$ (Chapter~\ref{ch:mm})}}
\begin{rem}\label{r:decomp-simple}
The theorem also holds with $\ol T(j,k) =
 w_k \de_{j,k}$, where $\de_{j,k}$ is defined by 
\begin{align*}
Q_{j,k}(dx) &= \min\bc{\dd{P_k}{P_j}, 1}\,P_j(dx)  && = \min\{p_j,p_k\}\dx \\
\de_{j,k} &= P_{j,k}(\Om) && = \int_{\Om} \min\{p_j,p_k\}\dx 
\end{align*}
where the equalities on the RHS hold when each $P_i$ has density function $p_i$.
For this definition of $\ol T$, the theorem holds with conclusion
\begin{align}\label{e:simple-alt}
\Var_P(g) &\le C \pa{1+2\ol C} \sE(g,g).
\end{align}•
\end{rem}

\begin{proof}
First, note that a stationary distribution $\ol P$ of $\ol M$ is given by $\ol p(j) := \ol P(\{j\})= w_j$, because $w_j\ol T(j,k) = w_k \ol T(k,j)$. (Note that the reason $\ol T$ has a maximum of $\chi^2$ divergences in the denominator is to make this ``detailed balance'' condition hold.)

Given $g\in L^2(\Om)$, define $\ol g\in L^2([m])$ by $\ol g(j) = \EE_{P_j} g$. Then decomposing the variance into the variance within the $P_j$ and between the $P_j$, and using Assumptions 2 and 3 gives
\begin{align}
\nonumber
\Var_P(g) &=\sumo jm w_j \int (g(x)-\E[g(x)])^2\, P_j(dx)\\
\nonumber
&=\sumo jm w_j \int (g(x)-\EE_{P_j}[g(x)])^2 \, P_j(dx)
+ \sumo jm w_j (\EE_{P_j} g - \EE_P g)^2\\
\nonumber
&\le \sumo jm w_j \int (g-\EE_{P_j} g)^2 \,P_j(dx) + \sumo jm \ol p(j) (\ol g(j) - \EE_{\ol P}\ol g)^2\\
\nonumber
&\le C \sumo jm w_j \sE_{P_j}(g,g) + \Var_{\ol P}(\ol g)\\
&\le C \sE(g,g) + \ol C \ol \sE (\ol g, \ol g).
\label{eq:to-cont}
\end{align}
Note $\sE(g,g) = \sumo jm w_j \sE_{P_j}(g,g)$ follows from Assumption 1.
Now
\begin{align}
\nonumber
\ol \sE (\ol g, \ol g) &= \rc 2\sumo jm \sumo km (\ol g(j)-\ol g(k))^2 w_j \ol T(j,k) \\
\label{e:simple-decomp-chi}
&\le \rc2\sumo jm \sumo km (\ol g(j)-\ol g(k))^2 w_j \fc{w_k}{\chi^2(P_k||P_j)}\\
&\le \rc2\sumo jm \sumo km \Var_{P_j}(g)w_jw_k &\text{by Lemma~\ref{lem:change-dist}}\\
&\le 
\rc2\sumo jm w_j C\sE_j(g,g) = 
\fc C2 \sE(g,g).
\end{align}
Thus
\begin{align}
\eqref{eq:to-cont}
&\le C \sE(g,g) + \fc{\ol C C}2\sE (g, g)
\end{align}
as needed.

For Remark~\ref{r:decomp-simple}, let $\wt P_{j,k}(dx) = \fc{P_{j,k}(dx)}{\de_{j,k}}=\fc{P_{j,k}(dx)}{P_{j,k}(\Om)}$; it is $\wt P_{j,k}$ normalized to be a probability distribution. 
Note that we can instead bound~\eqref{e:simple-decomp-chi} as follows.
\begin{align}
\eqref{e:simple-decomp-chi}&\le 
\sumo jm \sumo km \ba{\pa{\EE_{P_j} g - \EE_{\wt P_{j,k}}g}^2 + \pa{\EE_{P_k} g - \EE_{\wt P_{j,k}}g}^2}w_j w_k \de_{j,k}\\
&\quad\text{by }(a+b)^2\le 2(a^2+b^2)\nonumber\\
&\le \sumo jm \sumo km [\Var_{P_j}(g)\chi^2(P_j||\wt P_{j,k}) + \Var_{P_k}(g) \chi^2(P_k||\wt P_{j,k})] w_jw_k \de_{j,k} \\
&\quad \text{by Lemma~\ref{lem:change-dist}}\nonumber\\
& \le\sumo jm \sumo km (\Var_{P_j}(g)+\Var_{P_k}(g)) w_jw_k  \\
&\quad \text{by Lemma~\ref{l:overlap-chi}}\nonumber\\
&\le \sumo jm w_j 2C\sE_j(g,g) = 2C\sE(g,g)
\end{align}
which gives~\eqref{e:simple-alt}.
\end{proof}


\subsection{Theorem for simulated tempering}

The simple decomposition theorem requires us to decompose the stationary measures into measures that overlap. In the case of simulated tempering, however, the measures $P_{(i,j)}$ at different levels $i$ have disjoint supports. To adapt it to this case, we will let the probability flow in the projected chain depend on not just the distance between the probability measures, but also the ``flow" between them in the original chain. Thus, in the projected chain, we can let there be flow between $(i,j)$ and $(i',j')$ in the same and adjacent levels such that $P_{(i,j)}$ and $P_{(i',j')}$ are close.

Actually, it suffices to include connections at the highest level and for the same component between adjacent levels, so the adjacency graph of $\ol T$ contains a complete graph at the highest temperature, and ``chains" going down the levels. For alternatives, see the discussion in Appendix~\ref{a:decomp}.

\begin{thm}[Density decomposition theorem for simulated tempering]
\label{thm:gap-prod-st}
Consider simulated tempering $M$ with Markov processes $M_i=(\Om,\sL_i)$, $1\le i\le L$. Let the stationary distribution of $M_i$ be $P_i$, the relative probabilities be $r_i$, and the rate be $\la$. Let the Dirichlet forms be $\sE(g,h) = -\an{g,\sL h}_P$  and $\sE_i(g,h) = -\an{g,\sL_i h}_{P_i}$. Assume the probability measures have density functions with respect to some reference measure $dx$, represented by the lower-case letter: $P_i(dx)=p_i(x)\dx$.

Represent a function $g\in  [L]\times \Om$ as $(g_1,\ldots, g_L)$. Let $P$ be the stationary distribution on $[L]\times \Om$, $\sL$ be the generator, and $\sE(g,h) = -\an{g,\sL h}_P$ be the Dirichlet form.

Suppose the following hold.
\begin{enumerate}
\item
There is a decomposition
\begin{align}
\an{f, \sL_ig}_{P_i} &= \sum_{j=1}^{m_i} w_{i,j} \an{f, \sL_{(i,j)}g}_{P_{(i,j)}}\text{ for }f,g:\Om_i\to \R\\
P_i &= \sumo j{m} w_{i,j}P_{(i,j)}.\label{eq:decomp}
\end{align}
where $\sL_{i,j}$ is the generator for some Markov chain $M_{i,j}$ on $ \{i\}\times\Om$ with stationary measure $P_{(i,j)}$.
\item
(Mixing for each $M_{i,j}$) $M_{i,j}$ satisfies the Poincar\'e inequality
\begin{align}
\Var_{P_{(i,j)}}(g) &\le C \sE_{i,j}(g,g)
\end{align}
where $\sE_{i,j}(g,g)=-\an{g,\sL_{i,j} g}_{P_{(i,j)}}$.
\item
(Mixing for projected chain) 
Define
\begin{align}\label{e:st-proj}
\ol T((i,j),(i',j')) &=
\begin{cases}
\fc{w_{1,j'}}{\chi^2_{\max}(P_{(1,j)}||P_{(1,j')})},&i=i'=1\\
K 
{\de_{(i,j),(i',j)}},&i'=i\pm 1, \,j=j'\\
0,&\text{else}
\end{cases}\end{align}
where $\chi^2_{\max}(P||Q):= \max\{\chi^2(P||Q),\chi^2(Q||P)\}$, $K>0$ is any constant, and 
\begin{align}\de_{(i,j),(i',j')}&=\int_{\Om} \min\bc{\fc{r_{i'}w_{i',j}p_{(i',j)}(x)}{r_{i}w_{i,j}} ,p_{(i',j')}(x)}\dx.\end{align}
Define the projected chain $\ol M$ as the Markov chain on $[n]$ generated by $\ol{\sL}=\ol{\sT}-\Id$, so that $\ol{\sL}$ acts on $\ol g\in L^2([n])$ by
\begin{align}
\ol{\sL} \ol g(i,j) &= 
\sumo iL \sumo j{m} 
\sumo {i'}L \sumo {j'}{m}
[\ol g(i',j') - \ol g(i,j)] \ol{T}((i,j),(i',j')).
\end{align}
Let $\ol P$ be the stationary distribution of $\ol M$; $\ol M$ satisfies the Poincar\'e inequality
\begin{align}
\Var_{\ol P} (\ol g) & \le \ol C \ol{\sE}(g,g).
\end{align}
\end{enumerate}
Then $M$ satisfies the Poincar\'e inequality
\begin{align}
\Var_P(g) &\le \max\bc{C \pa{1+
\pa{\rc 2 + 6K}\ol C}, \fc{
6  K \ol C}{\la}} \sE(g,g).
\end{align}
\end{thm}
We will use the following lemma.
\begin{lem}\label{lem:change-dist-f}
Let $P_1,P_2$ be probability distributions on $\Om$, and $g_1,g_2:\Om\to \R$, $g_i\in L^2(P_i)$.  
Let $Q$ be the measure that is the minimum of $P_1$ and $P_2$: $Q=\min\bc{\dd{P_2}{P_1},1} P_1$. Let $\de = Q(\Om)$ be the normalization constant (suppose $\de>0$) and $\wt Q = \rc{\de}Q$ be the normalized probability measure. Then
\begin{align}
\pa{
\int_\Om g_1 \,P_1(dx) - \int_\Om  g_2 \, P_2(dx)
}^2 
&\le 
3\ba{\Var_{P_1}(g_1) \chi^2(\wt Q||P_1) + \Var_{P_2}(g_2) \chi^2(\wt Q||P_2) 
+ \int_\Om (g_1-g_2)^2\wt Q(dx)}
\end{align}
\end{lem}
\begin{proof}
By Cauchy-Schwarz and Lemma~\ref{lem:change-dist} we have
\begin{align}
&\pa{
\int_\Om  g_1\,P_1(dx) - \int_\Om  g_2\, P_2(dx)}^2 \\
&\le \ba{\int_\Om  g_1(P_1(dx)-\wt Q(dx)) + \int_\Om  (g_1-g_2)\wt Q(dx) + \int_\Om  g_2(\wt Q(dx)-P_2(dx))}^2\\
&\le 3\ba{
\pa{\int_\Om  g_1(P_1(dx)-\wt Q(dx))}^2  + \pa{\int_\Om  g_2(\wt Q(dx)-P_2(dx))}^2 + \pa{\int_\Om  (g_1-g_2)\wt Q(dx)}^2
}\\
&\le  3\ba{\Var_{p_1}(g_1)\chi^2(\wt Q||P_1) + \Var_{P_2}(g_2) \chi^2 (Q||P_2) 
+ \int_{\Om} (g_1-g_2)^2 \wt Q(dx)\cancel{\int_{\Om} \wt Q(dx)}}.
\end{align}
\end{proof}


\begin{proof}[Proof of Theorem~\ref{thm:gap-prod-st}]
First, note that the stationary distribution $\ol P$ of $\ol M$ is given by $\ol p((i,j)) = r_i w_{i,j}$, because $\ol T((i,j),(i',j')) r_iw_{i,j} = \ol T((i',j'), (i,j)) r_{i'}w_{i',j'}$. We can check that 
\begin{align}r_iw_{i,j} \de_{(i,j),(i',j')} =\min\bc{r_{i'}w_{i',j'}p_{(i',j')},r_{i}w_{i,j}p_{(i,j)}} = r_{i'}w_{i',j'} \de_{(i',j'),(i,j)}.\end{align}

Given $g\in L^2([L]\times \Om)$, define $\ol g\in L^2\pa{\bigcup_{i=1}^L( \{i\}\times [m])}$ by $\ol g(i,j) = \EE_{P_{(i,j)}} g_i$.
\begin{align}
\Var_P(g) &=\sumo iL \sumo j{m} r_{i}w_{i,j} \int_\Om (g-\EE_P g)^2P_j(dx)\\
&=\sumo iL \sumo j{m} r_{i}w_{i,j}
\ba{
\pa{\int_\Om (g_i-\EE_{P_{(i,j)}} g_i)^2 P_{(i,j)}(dx)}
+\pa{\EE_{P_{(i,j)}} g_i - \EE_P g}^2}\\
&\le C \sumo iL \sumo j{m} r_{i}w_{i,j} \sE_{{i,j}}(g_i,g_i) + \Var_{\ol P}(\ol g)\\
&\le C \sumo iL r_i \sE_i(g_i,g_i) + \ol C \ol \sE (\ol g, \ol g).
\label{eq:to-cont2}
\end{align}
Now $\ol{\sE}$ has two terms; the first is bounded in the same way as in Theorem~\ref{thm:gap-prod}.
\begin{align}
\ol \sE(\ol g, \ol g) &=\rc2\ub{ 
\sumo jn \sumo{j'}n (\ol g(1,j)-\ol g(1,j'))^2
r_1w_{1,j}\ol T((1,j),(1,j'))}{A} \\
&\quad + 
\rc 2\ub{\sumo iL
\sumo j{m}
\sumr{i'=i\pm 1}{1\le i'\le L}
 (\ol g(i,j)-\ol g(i',j))^2
r_iw_{i,j}\ol T((i,j),(i',j))}{B}
\end{align}
First we bound
\begin{align}
A
&\le 
\sumo j{m}\sumo{j'}n (\ol g(1,j)-\ol g(1,j'))^2
r_1w_{1,j}\fc{w_{1,j'}}{\chi_{\max}^2(P_{1,j}||P_{1,j'})}
\\
&\le 
\sumo j{m}\sumo{j'}n r_1w_{1,j}w_{1,j'}
\Var_{P_{1,j}}(g_i)
 &\text{by Lemma~\ref{lem:change-dist-f}}\\
&\le  
\sumo j{m}  r_1 w_{1,j} \Var_{P_{1,j}}(g_i)\\
&\le 
r_1 C\sE_{1}(g_i,g_i).
\end{align}
For the second term, let $\wt P_{(i,j),(i',j')}$ be the probability measure with density function \begin{align}
\wt p_{(i,j),(i',j')}=
\rc{\de_{(i,j),(i',j')}} \min\bc{\fc{r_{i'}w_{i',j}p_{(i',j)}(x)}{r_{i}w_{i,j}} ,p_{(i',j')}(x)}.
\end{align}
We use Lemma~\ref{lem:change-dist-f}.
\begin{align}
B
&\le 3
\sumo iL
\sumo j{m}
\sumr{i'=i\pm 1}{1\le i'\le L}
\Bigg[
\pa{\Var_{P_{i,j}}(g_i)\chi^2(\wt P_{(i,j),(i',j)}||P_{i,j})+\Var_{P_{i',j}}(g_{i'})\chi^2(\wt P_{(i,j),(i',j)}||P_{i',j})}  \\
&\quad +
\int_\Om (g_i-g_{i'})^2  \fc{\min\bc{\fc{r_{i'}w_{i',j}p_{(i',j)}(x)}{r_{i}w_{i,j}} ,p_{(i',j)}(x)}}{\de_{(i,j),(i',j)}}\dx
\Bigg]\cdot 
r_iw_{i,j}K \de_{(i,j),(i',j)}\\
&\le 3K
\sumo iL
\sumo j{m}
\sumr{i'=i\pm 1}{1\le i'\le L}
\Bigg[\Big(\Var_{P_{i,j}}(g_i)r_iw_{i,j}+\Var_{P_{i',j}}(g_{i'})r_iw_{i,j}\ub{\fc{\de_{(i,j),(i',j)}}{\de_{(i',j),(i,j)}}}{\fc{r_{i'}w_{i',j}}{r_i w_{i,j}}}\Big)  \\
&\quad 
+ r_iw_{i,j} 
 \int_\Om (g_i-g_{i'})^2 \min\bc{\fc{r_{i'}w_{i',j}p_{(i',j)}(x)}{r_{i}w_{i,j}} ,p_{(i',j)}(x)}
\Bigg]\\
&\quad \text{by Lemma~\ref{l:overlap-chi}}\\
&\le 3K
\sumo iL
\sumo j{m}
\sumr{i'=i\pm 1}{1\le i'\le L}
\Bigg[r_iw_{i,j} \Var_{P_{i,j}}(g_i)
+ r_{i'}w_{i',j}\Var_{P_{i',j}}(g_{i'}) \\
&\quad
 + 3K\sumo iL
\sumr{i'=i\pm 1}{1\le i'\le L}
\int_\Om (g_i-g_{i'})^2 \min\bc{\sumo j{m}  r_iw_{i,j}
p_{(i,j)}(x),\sumo j{m} 
r_{i'}w_{i',j}p_{(i',j)}(x)}\dx\\
&\le 12K
\sumo iL
\sumo j{m} r_iw_{i,j}\Var_{P_{i,j}}(g_i)
 + 3K\sumo iL
\sumr{i'=i\pm 1}{1\le i'\le L}
\int_\Om (g_i-g_{i'})^2 \min\{r_ip_{i}(x), r_{i'}p_{i'}(x)\}\dx
\\
&\le 12K
\sumo iL r_i C\sE_i(g_i,g_i) + 3K
\sumo iL \sumr{i'=i\pm 1}{1\le i'\le L}\int_\Om (g_i-g_{i'})^2 \min \{r_ip_i(x),r_{i'}p_{i'}(x)\}\dx
\end{align}
Then 
\begin{align}
\eqref{eq:to-cont2}
&\le C\sumo iL r_i \sE_i(g_i,g_i)\\
&\quad 
+\fc{\ol C}2
\pa{
(1+12K)C\sumo iL r_i \sE_{i}(g_i,g_i) + 
\fc{
12K}{\la} \fc{\la}
4 
\sumo iL \sumr{i'=i\pm 1}{1\le i'\le L} (g_i-g_{i'})^2 \min \{r_ip_i(x),r_{i'}p_{i'}(x)\}\dx
}\\
&\le 
\max\bc{C \pa{\rc 2+6\ol C}, \fc{6K\ol C}{\la}}
\sE(g,g).
\end{align}
\end{proof}

\section{Simulated tempering for gaussians with equal variance}
\label{sec:st-gaussian}
\subsection{Mixtures of gaussians all the way down}

\begin{thm}\label{thm:st-gaussians}
Let $M$ be the continuous simulated tempering chain for the distributions with density functions
\begin{align}
p_i(x)&\propto \sumo jm w_j e^{-\be_i\fc{\ve{x-\mu_j}^2}{2\si^2}}
\end{align}
with rate $\Om\pf{1}{D^2}$, relative probabilities $r_i$, and temperatures $0<\be_1<\cdots <\be_L=1$ where
\begin{align}
D&= \max\{\max_j\ve{\mu_j}, \si\}\\
\be_1 &= \Te\pf{\si^2}{D^2}\\
\fc{\be_{i+1}}{\be_i} &\le 1+\rc d\\
L &= \Te\pa{d\ln\pf{D}{\si}+1}\\
r&=\fc{\min_i r_i}{\max_i r_i}.
\end{align}
Then $M$ satisfies the Poincar\'e inequality
\begin{align}
\Var(g) &\le O\pf{L^2D^2}{r^2}\sE(g,g) = O\pf{\pa{d\ln \pf{D}{\si}+1}^2D^2}{r^2}\sE(g,g).
\end{align}
\end{thm}


\begin{proof}
Note that forcing $D\le \si$ ensures $\be_1=\Om(1)$. 
We check all conditions for Theorem~\ref{thm:gap-prod-st}. We let $K=1$.
\begin{enumerate}
\item
Consider the decomposition where 
\begin{align}
p_{i,j}(x)&\propto \exp\pa{-\be_i \fc{\ve{x-\mu_j}^2}{2\si^2}},
\end{align}
$w_{i,j}=w_j$, and
and $M_{i,j}$ is the Langevin chain on $p_{i,j}$, so that $\sE_{ij}(g_i,g_i) = \int_{\R^d} \ve{\nb g_i}^2p_{i,j}\dx$. We check~\eqref{eq:decomp}:
\begin{align}
\sE_i(g_i,g_i) =\int_{\R^d} \ve{\nb g_i}^2p_i\dx = \int_{\R^d} \ve{\nb g_i}^2 \sumo jm w_j p_j \dx= \sumo jm w_j \sE_{i,j} (g_i,g_i).
\end{align}
\item
By Theorem~\ref{thm:bakry-emery} and the fact that $\be_1=\Om\pf{\si^2}{D^2}$, $\sE_{i,j}$ satisfies the Poincar\'e inequality
\begin{align}
\Var_{p_{i,j}}(g_i) &\le \fc{\si^2}{\be_i} \sE_{i,j}(g_i,g_i) =
O(D^2) \sE_{i,j}(g_i,g_i).
\end{align}
\item
To prove a Poincar\'e inequality for the projected chain, we use the method of canonical paths, Theorem~\ref{thm:can-path}. 
Consider the graph $G$ on $\bigcup_{i=1}^L \{i\}\times [m_i]$ that is the complete graph on the slice $i=1$, and the only other edges  are vertical edges $(i,j), (i\pm 1,j)$. 
$\ol T$ is nonzero exactly on the edges of $G$.
For vertices $x=(i,j)$ and $y=(i',j')$, define the canonical path as follows.
\begin{enumerate}
\item
If $j=j'$, without loss of generality $i<i'$. Define the path to be $(i,j), (i+1,j),\ldots, (i',j)$.
\item
Else, define the path to be $(i,j), (i-1,j),\ldots , (1,j), (1,j'), \ldots, (i,j')$.
\end{enumerate}


We calculate the transition probabilities~\eqref{e:st-proj}, which are given in terms of the 
$\chi^2$ distances $\chi^2_{\max} (P_{1,j}||P_{1,j'})$ and overlaps $\de_{(i,j),(i',j)}$.
\begin{enumerate}
\item
Bounding $\chi^2(P_{1,j}||P_{1,j'})$: By Lemma~\ref{lem:chi-squared-N} with $\Si_1=\Si_2=\be_1^{-1}I_d$,
\begin{align}
\chi^2(P_{1,j}||P_{1,j'}) &= \chi^2(N(\mu_j, \be_1I_d)||N(\mu_{j'}, \be_1I_d))\\
&=e^{\be_1\ve{\mu_1-\mu_2}^2/\si^2} = \rc 4
\label{e:chi2-p1}
\end{align}
when $\be_1\le c\fc{\si^2}{D^2}$ for a small enough constant $c$. 
\item
Bounding $\de_{i,j,i',j}$: Suppose that $\fc{\be_{i+1}}{\be_i} =1+\ep$ where $\de\le \rc d$. Then applying Lemma~\ref{lem:chi-squared-N} to $\Si_1=\be_i^{-1} I_d$ and $\Si_2=\be_{i+1}^{-1}I_d$,
\begin{align}
\chi^2(P_{i+1,j}||P_{i,j})
&= \chi^2(N(\mu_j, \be_{i+1}I_d)||N(\mu_j, \be_i I_d))\\
& = \pf{\be_{i+1}^2}{\be_{i}}^{\fc d2} (2\be_{i+1} - \be_i)^{-\fc d2}-1\\
&= \pf{\be_{i+1}}{\be_i}^{\fc d2} \pa{2-\fc{\be_i}{\be_{i+1}}}^{-\fc d2} - 1\\
&=O
\pa{
(1+d\ep)\pa{2-\prc{1+\ep}}^{-\fc d 2}-1
}=O(d\ep)
\label{eq:chi-adj}
\end{align}
so $\chi^2(P_{i+1,j}||P_{i,j})\le \rc 4$ when $\de\le c\rc{\de}$ for a small enough constant $c$. Similarly, $\chi^2(P_{i-1,j}||P_{i,j}) = \rc 4$ for $\de \le c\rc{\de}$.

Note that for probability distributions $P_1,P_2$ with density functions $p_1,p_2$,
\begin{align}
\pa{\int_{\Om} (p_1-\min\{p_1,p_2\})\dx}^2 &\le \int_{\Om} \fc{(p_1-\min\{p_1,p_2\})^2}{p_1}\dx = \chi^2(P_2||P_1)\\
\int_{\Om}\min\{p_1,p_2\}\dx &\ge  1-\sqrt{\chi^2(P_2||P_1)}.
\end{align}
Moreover, we have
\begin{align}
\de_{(i,j),(i\pm 1,j)} &=\int \min\bc{\fc{r_{i'}w_{j}p_{i',j}(x)}{r_{i}w_{j}}, p_{i,j}}\dx
\ge r\int \min\{p_{i',j}(x),p_{i,j}(x)\}\dx
\end{align}•
Hence $\de_{(i,j),(i\pm 1,j)}\ge \rc 2r$.
\end{enumerate}
Note that $|\ga_{x,y}|\le 2L-1$. Consider two kinds of edges in $G$.
\begin{enumerate}
\item
$z=(1,j)$, $w=(1,k)$. We have
\begin{align}\label{eq:st-path2}
\fc{\sum_{\ga_{x,y}\ni((1,j),(1,k))} |\ga_{x,y}|\ol p(x)\ol p(y)}{\ol p((1,j))T((1,j), (1,k))}
&\le \fc{(2L-1) \ol P([L]\times \{j\})\ol P([L]\times \{k\})}{\ol p((1,j))\ol T((1,j),(1,k))}.
\end{align}
because the paths going through $zw$ are exactly those between $[L]\times \{j\}$ and $[L]\times \{k\}$. 
Now note
\begin{align}
\fc{\ol P([L]\times \{j\})}{\ol p((1,j))}&\le \fc{L}r\\
\ol P([L]\times \{k\}) &= w_k\\
\ol T((1,j),(1,k))&=\rc 2 
\fc{w_k}{\chi^2_{\max}(P_{1,j}||P_{1,j'})}
=\Om(w_k)
\end{align}
by~\eqref{e:chi2-p1}. 
Thus $\eqref{eq:st-path2}=O\pf{L^2}r$.
\item
$z=(i,j)$, $w=(i-1,j)$. We have
\begin{align}\label{eq:st-path1}
\fc{\sum_{\ga_{x,y}\ni((i,j),(i-1,j))} |\ga_{x,y}|\ol p(x)\ol p(y)}{\ol p((i,j))\ol T((i,j), (i-1,j))}
&\le \fc{(2L-1)\ol P(S)\ol P(S^c)}{p((i,j))\ol T((i,j),(i-1,j))}
\end{align}
where $S= \{i,\ldots, L\}\times \{j\}$. 
This follows because cutting the edge $zw$ splits the graph into 2 connected components, one of which is $S$; the paths which go through $zw$ are exactly those between $x,y$ where one of $x,y$ is a subset of $S$ and the other is not. Now note
\begin{align}
\fc{\ol P(S)}{\ol p((i,j))}=
\fc{\ol P(\{i,\ldots ,L\}\times \{j\})}{\ol p((i,j))} &\le \fc{L}{r}\\
\ol P(S^c)&\le 1\\
\ol T((i,j),(i-1,j))&=
 \de_{(i,j),(i-1,j)} = \Om(r)
\end{align}
by~\eqref{e:st-proj} and the inequality $\de_{(i,j),(i-1,j)}\ge \rc2 r$. Hence $\eqref{eq:st-path1}= O\pf{L^2}{r^2}$.
\end{enumerate}
By Theorem~\ref{thm:can-path}, the projected chain satisfies a Poincar\'e inequality with constant $O\pf{L^2}{r^2}$.
\end{enumerate} 
Thus by Theorem \ref{thm:gap-prod-st}, the simulated tempering chain satisfies a Poincar\'e inequality with constant
\begin{align}
O\pa{\max\bc{
D^2 \pa{1+\fc{L^2}{r^2}}, \fc{L^2}{r^2\la}
}}.
\end{align}
Taking $\la = \fc{1}{D^2}$ makes this $O\pf{D^2L^2}{r^2}$.
\end{proof}

\begin{rem}\label{rem:conv-p}
Note there is no dependence on either $w_{\min}$ or the number of components.

If $p(x)\propto \sumo jm w_je^{-\fc{\ve{x-\mu_j}^2}{2\si^2}}$ and we have access to $\nb \ln (p*N(0,\tau I))$ for any $\tau$, then we can sample from $p$ efficiently, no matter how many components there are. In fact, passing to the continuous limit, we can sample from any $p$ in the form $p = w*N(0,\si^2I_d)$ where $\ve{w}_1=1$ and $\Supp(w)\subeq B_D$. 

In this way, Theorem~\ref{thm:st-gaussians} says that evolution of $p$ under the heat kernel is the most ``natural'' way to do simulated tempering.
We don't have access to $p*N(0,\tau I)$, but we will show that $p^{\be}$ approximates it well (within a factor of $\rc{w_{\min}}$).

Entropy-SGD~\cite{chaudhari2016entropy} attempts to estimate $\nb \ln (p*N(0,\tau I))$ for use in a temperature-based algorithm; this remark provides some heuristic justification for why this is a natural choice.
\end{rem}


\subsection{Comparing to the actual chain}


The following lemma shows that changing the temperature is approximately the same as changing the variance of the gaussian. We state it more generally, for arbitrary mixtures of distributions in the form $e^{-f_i(x)}$. 

\begin{lem}[Approximately scaling the temperature]\label{lem:close-to-sum}
Let $p_i (x)= e^{-f_i(x)}$ be probability distributions on $\Om$ such that for all $\be>0$, $\int_\Om e^{-\be f_i(x)}\dx<\iy$. Let 
\begin{align}
p(x) & = \sumo in \weight_i p_i(x)\\
f(x) &= -\ln p(x)
\end{align}
where $\weight_1,\ldots, \weight_n>0$ and $\sumo in \weight_i=1$. Let $w_{\min}=\min_{1\le i\le n}\weight_i$.

Define the distribution at inverse temperature $\be$ to be $p_\be(x)$, where
\begin{align}
g_\be(x) &= e^{-\be f(x)}\\
Z_\be &= \int_{\Om} e^{-\be f(x)}\dx\\
p_\be(x) &= \fc{g_\be(x)}{Z_\be}.
\end{align}
Define the distribution $\wt p_\be(x)$ by
\begin{align}
\wt g_\be(x) &= \sumo in \weight_i e^{-\be f_i(x)}\\
\wt Z_\be &= \int_{\Om} \sumo in {\weight_i e^{-\be f_i(x)}}\dx\\
\wt p_\be(x) &= \fc{\wt g_\be(x)}{\wt Z_\be}.
\end{align}
Then for $0\le \be\le 1$ and all $x$,
\begin{align}
\label{eq:scale-temp1}
g_\be(x) &\in \ba{1,\rc{w_{\min}}} \wt g_\be\\
\label{eq:scale-temp2}
p_\be(x)& \in \ba{1, \rc{w_{\min}}}\wt p_\be \fc{\wt Z_\be}{Z_\be}\sub \ba{w_{\min}, \rc{w_{\min}}}\wt p_\be.
\end{align}
%
%
\end{lem}
\begin{proof}
By the Power-Mean inequality,
\begin{align}
g_\be(x) &= \pa{\sumo in w_i e^{-f_i(x)}}^\be\\
&\ge \sumo in w_i e^{-\be f_i(x)} = \wt g_\be(x).
\end{align}
On the other hand, given $x$, setting $j=\amin_i f_i(x)$, 
\begin{align}
g_\be(x) & = \pa{\sumo in w_i e^{-f_i(x)}}^\be\\
&\le (e^{-f_j(x)})^{\be}\\
&\le \rc{w_{\min}}\sumo in w_i e^{-\be f_i(x)} = \rc{w_{\min}} \wt g_\be(x).
\end{align}
This gives~\eqref{eq:scale-temp1}. This implies $\fc{\wt Z_\be}{Z_\be} \in [w_{\min},1]$, which gives~\eqref{eq:scale-temp2}.
\end{proof}


\begin{lem}
\label{lem:poincare-liy}
Let $P_1$, $P_2$ be probability measures on $\R^d$ with density functions $p_1\propto e^{-f_1}$, $p_2\propto e^{-f_2}$ satisfying $\ve{f_1-f_2}_\iy\le \frac{\De}{2}$.
Then 
\begin{align}
\fc{\sE_{P_1}(g,g)}{\ve{g}_{L^2(P_1)}^2} \ge e^{-2\De} \fc{\sE_{P_2}(g,g)}{\ve{g}_{L^2(P_2)}^2}.
\end{align}
\end{lem}
\begin{proof}
The ratio between $p_1$ and $p_2$ is at most $e^{\Delta}$, so 
\begin{align}
\fc{\int_{\R^d} \ve{\nb g(x)}^2p_1(x)\dx}{\int_{\R^d} \ve{g(x)}^2 p_1(x)\dx}
&\ge 
\fc{ e^{-\dellarge}\int_{\R^d} \ve{\nb g(x)}^2p_2(x)\dx}{e^{\dellarge}\int_{\R^d} \ve{g(x)}^2 p_2(x)\dx}.
\end{align}
\end{proof}

\begin{lem}\label{lem:st-comparison}
Let $M$ and $\wt M$ be two continuous simulated tempering Langevin chains with functions $f_i$, $\wt f_i,$, respectively, for $i\in [L]$, with rate $\la$, and with relative probabilities $r_i$. Let their Dirichlet forms be $\sE$ and $\wt \sE$ and their stationary measures be $P$ and $\wt P$.

Suppose that $\ve{f_i(x)-\wt f_i(x)}_\iy\le \fc{\De}2$. 
Then\footnote{If adjacent temperatures are close enough, then $A$ and $B$ in the proof are close, so $[\min\{A,B\},\max\{Ae^\De, Be^\De\}]\subeq [C, C\cdot O(e^\De)]$ for some $C$, improving the factor to $\Om(e^{-2\De})$. A more careful analysis would likely improve the final dependence on $w_{\min}$ \ift{in Theorem~\ref{thm:main}} from $\rc{w_{\min}^6}$ to $\rc{w_{\min}^4}$. \iftoggle{thesis}{}{See Section~\ref{s:comparison}.}}
\begin{align}
\fc{\sE(g,g)}{\Var_P(g)} \ge e^{-\wexp\De} \fc{\wt \sE(g,g)}{\Var_{\wt P}(g)}.
\end{align}
\end{lem}
\begin{proof}
By Lemma~\ref{lem:poincare-liy},
\begin{align}
\fc{\sumo iL \sE_i (g_i,g_i)}{\Var_{P_i} (g_i)} &\ge e^{-2\De} 
\fc{\sumo iL \wt \sE_i (g_i,g_i)}{\Var_{\wt P_i} (g_i)}\\
\implies
\fc{\sumo iL r_i\sE_i (g_i,g_i)}{\Var_{P} (g_i)}
&\ge e^{-2\De} 
\fc{\sumo iL r_i\wt \sE_i (g_i,g_i)}{\Var_{\wt P} (g_i)}.\label{eq:comp1}
\end{align}
By Lemma~\ref{lem:close-to-sum}, we have $\fc{p_i}{\wt p_i}\in [A , Ae^{\De}]$, $\fc{p_j}{\wt p_j}\in [B, Be^\De]$ for some $A,B\ge e^{-\De}$, so $\fc{\min\{r_ip_i,r_jp_j\}}{\min\{r_i\wt p_i, r_j\wt p_j\}}\in [\min\{A,B\}, \max\{Ae^{\De}, Be^{\De}\}]\subeq[e^{-\De},e^{\De}]$. 
Hence, 
\begin{align}
\fc{\int_{\R^d} (g_i-g_j)^2 \min\{r_ip_i,r_jp_j\}\dx}{\int_{\R^d} (g_i-g_j)^2 \min\{r_i\wt p_i, r_j\wt p_j\}\dx}
&\in [e^{-\De},e^{\De}]\\
\fc{\Var_{P_i}(g_i)}{\Var_{\wt{P_i}}(g_i)} &\in [A, Ae^{\De}]\\
\implies 
\label{e:pi-perturb1}
\fc{\fc\la 4 \sumo iL \sum_{j=i\pm 1} \int_{\R^d} (g_i-g_j)^2\min\{r_ip_i,r_jp_j\}\dx}{\fc \la4 \sumo iL \sum_{j=i\pm 1}\int_{\R^d} (g_i-g_j)^2 \min\{r_i\wt p_i, r_j\wt p_j\}\dx} &\in 
[e^{-\De},e^{\De}]\\
\label{e:pi-perturb2}
\fc{\Var_P(g)}{\Var_{\wt P}(g)} = \fc{\sumo iL r_i \Var_{P_i}(g_i)}{\sumo iL r_i \Var_{\wt P_i}(g_i)} &\in [A,Ae^{\De}]
\end{align}
Dividing~\eqref{e:pi-perturb1} by~\eqref{e:pi-perturb2} gives
\begin{align}
&\fc{\fc\la 4 \sumo iL \sum_{j=i\pm 1} \int_{\R^d} (g_i-g_j)^2\min\{r_ip_i,r_jp_j\}\dx}{\Var_{P}(g)}\\
&\ge e^{-\wexp\De}\fc{\fc \la4 \sumo iL \sum_{j=i\pm 1}\int_{\R^d} (g_i-g_j)^2 \min\{r_i\wt p_i, r_j\wt p_j\}\dx}{ \Var_{\wt P}(g)}\label{eq:comp2}
\end{align}
Adding~\eqref{eq:comp1} and~\eqref{eq:comp2} gives the result.
\end{proof}

\begin{thm}\label{thm:st-gaussians2}
Suppose $\sumo jm w_j=1$, $w_{\min}=\min_{1\le j\le m} w_i >0$, and \\
$D=\max\{\max_{1\le j\le m} \ve{\mu_j}, \si\}$. 
Let $M$ be the continuous simulated tempering chain for the distributions
\begin{align}
p_i(x)&\propto \pa{\sumo jm w_j e^{-\fc{\ve{x-\mu_j}^2}{2\si^2}}}^{\be_i}
\end{align}
with rate $O\pf{1}{D^2}$, relative probabilities $r_i$, and temperatures $0<\be_1<\cdots <\be_L=1$ 
satisfying the same conditions as in Theorem~\ref{thm:st-gaussians}.
Then $M$ satisfies the Poincar\'e inequality
\begin{align}
\Var(g) &\le O\pf{L^2D^2}{r^2w_{\min}^{\wexp}}\sE(g,g) = O\pf{d^2\pa{\ln\pf{D}{\si}}^2D^2}{r^2w_{\min{}}^{\wexp}}\sE(g,g).
\end{align}
\end{thm}
\begin{proof}
Let $\wt p_i$ be the probability distributions in Theorem~\ref{thm:st-gaussians} with the same parameters as $p_i$ and let $\wt p$ be the stationary distribution of that simulated tempering chain. By Theorem~\ref{thm:st-gaussians}, $\Var_{\wt P}(g) =O\pf{L^2D^2}{r^2} \sE_{\wt P}(g,g)$. Now use 
By Lemma~\ref{lem:close-to-sum}, $\fc{p_i}{\wt p_i} \in \ba{1,\rc{w_{\min{}}}} \fc{\wt Z_i}{Z_i}$. Now use Lemma~\ref{lem:st-comparison} with $e^\De = \rc{w_{\min}}$.
\end{proof}

%

\section{Discretization}\label{sec:disc}

Throughout this section, let $f$ be as in Theorem~\ref{thm:mainlogconcave} ($f=\sumo im w_i e^{-f_0(x-\mu_i)}$, where $f_0$ is $\ka$-strongly convex, $K$-smooth, and has minimum at 0).

\begin{lem} 
Fix times $0<T_1<\cdots <T_n\le T$. 

Let $p^T, q^T: [L]\times  \mathbb{R}^d \to \mathbb{R}$ be probability density functions defined as follows (and let $P^T$, $Q^T$ denote the corresponding measures).
\begin{enumerate}
\item
$p^T$ is the density function of the continuous simulated tempering Markov process as in 
Definition~\ref{df:cst} but with fixed transition times $T_1,\ldots, T_n$. The component chains are Langevin diffusions on $p_i(x)\propto \pa{\sumo jm w_j e^{-f_0(x-\mu_i)}}^{\be_i}$.
\item
 $q^T$ is the discretized version as in Algorithm~\ref{a:stlmc}, again with fixed transition times $T_1,\ldots, T_n$, and with step size $\eta\le \fc{\si^2}2$.
\end{enumerate}
Then

\begin{align*} \mbox{KL} (P^T || Q^T) \lesssim \eta^2 D^6 \maxeig^7 \left(D^2 \frac{\maxeig^2}{\mineig}+d\right) T n + \eta^2 D^3 \maxeig^3 \max_i \E_{x \sim P^0( \cdot, i)}\|x - x^*\|_2^2 + \eta D^2 \maxeig^2 d T   \end{align*}

where $x^*$ is the maximum of $\sumo jm w_j e^{-f_0(x-\mu_j)}$ and satisfies $\ve{x^*}=O(D)$ where $D=\max\ve{\mu_j}$.
\label{l:maindiscretize}
\end{lem} 
\Anote{Propagate bound to ``main theorem'' bounds}

Before proving the above statement, we make a note on the location of $x^*$ to make sense of $\max_i \E_{x \sim P^0(i, \cdot)}\|x - x^*\|_2^2$.  Namely, we show:  

\begin{lem}[Location of minimum] Let $x^* = \mbox{argmin}_{x \in \mathbb{R}^d} f(x)$. Then, $\|x^*\| \leq D \sqrt{\frac{\maxeig}{\mineig}+1}$.  
\label{l:locatemin}
\end{lem} 
\begin{proof} 

Recall that $f_i(x) = f_0(x - \mu_i)$.
We claim that $f(0) \leq \frac{1}{2} \maxeig D^2$. 
Indeed, by smoothness, we have $f_i(0) \leq \frac{1}{2} \maxeig \|\mu_i\|^2 $, which implies that 
$f(0) \leq \frac{1}{2} \maxeig D^2$. 

Hence, it follows that $\min_{x \in \mathbb{R}^d} f(x) \leq \frac{1}{2} \maxeig D^2$. However, for any $x$, it holds that 
\begin{align*} f(x) &\geq \frac{1}{2}\min_i \mineig \|\mu_i - x\|^2 \\
&\geq \frac{1}{2} \mineig \left(\|x\|^2 - \max_i\|\mu_i\|^2\right) \\
&\geq \frac{1}{2} \mineig \left(\|x\|^2 - D^2\right) \end{align*} 
Hence, if $\|x\| >  D\sqrt{\frac{\maxeig}{\mineig}+1}$, $f(x) >  \min_{x \in \mathbb{R}^d} f(x)$. This implies the statement of the lemma. 
\end{proof} 

We prove a few technical lemmas. First, we prove that the continuous chain is essentially contained in a ball of radius $D$. More precisely, we show:

\begin{lem}[Reach of continuous chain] Let $P^{\beta}_T(X)$ be the Markov kernel corresponding to evolving Langevin diffusion 
\begin{equation*}
dX_t = - \beta \nabla f(X_t)\,dt + \mathop{d B_t}\end{equation*} 
where $\tilde{f}$ and $D$ are as defined in \eqref{eq:A0-p} for time $T$. Then, 
\begin{equation*}\E[\|X_t - x^*\|^2] \leq \E[\|X_0 - x^*\|^2] + \pa{400 \beta \frac{D^2 \maxeig^2}{\mineig} + 2 d}T. \end{equation*} 
\label{l:reachcontinuous}
\end{lem} 
\begin{proof} 
Let $Y_t = \|X_t - x^*\|^2$. By It\^{o}s Lemma, we have 
\begin{equation} d Y_t = -2 \an{X_t - x^*, \beta \sum_{i=1}^m \frac{w_i e^{-f_i(X_t)} \nabla f_i(X_t)}{\sum_{j=1}^m w_j e^{-f_j(X_t)}}  }+ 2 d \mathop{dt} + \sqrt{8} \sum_{i=1}^d (X_t)_i \mathop{d(B_i)_t} \label{eq:contdrift1} \end{equation} 
We will show that  
$$- \an{ X_t - x^*, \nabla f_i(X_t)} \leq 100 \frac{D^2 \maxeig^2}{\mineig}$$
Indeed, since $f_i(x) = f_0(x - \mu_i)$, by {\eqref{eq:contdrift1}}, we have 
$$\langle X_t, \nabla f_i(X_t) \rangle \geq \frac{\mineig}{2} \|X_t\|^2 - \frac{D^2(2\mineig+\maxeig)^2}{2\mineig} - \maxeig D^2$$
Also, by the Hessian bound $\mineig I \preceq \nb^2 f_0(x)\preceq \maxeig I$, 
we have 
$$\langle x^*, \nabla f_i(X_t)\rangle \leq \|x^*\| \|\nabla f_i(X_t) \| \leq  D \sqrt{\frac{\maxeig}{\mineig}+1} \|X_t-\mu_i\| \leq  D\sqrt{\frac{\maxeig}{\mineig}+1}(\|X_t\| + D)$$
Hence, 
$$ - \langle X_t - x^*, \nabla f_i(X_t) \rangle\leq - \frac{\mineig}{2} \|X_t\|^2 - \frac{D^2(2\mineig+\maxeig)^2}{2\mineig} -  D\sqrt{\frac{\maxeig}{\mineig}+1}(\|X_t\| + D) $$  
Solving for the extremal values of the quadratic on the RHS, we get 
$$ - \an{X_t - x^*, \nabla f_i(X_t)} \leq 100 \frac{D^2 \maxeig^2}{\mineig}  $$

Together with \eqref{eq:contdrift1}, we get 
\begin{equation*} d Y_t \leq  100 \beta \frac{D^2 \maxeig^2}{\mineig}  + 2 d \mathop{dt} + \sqrt{8} \sum_{i=1}^d (X_t)_i \mathop{d(B_i)_t}  \end{equation*} 
Integrating, we get 
\begin{equation*} Y_t \leq  Y_0 + 400 \beta \frac{D^2 \maxeig^2}{\mineig} T + 2 d T + \sqrt{8} \int^T_0 \sum_{i=1}^d (X_t)_i \mathop{d(B_i)_t}  \end{equation*} 
Taking expectations and using the martingale property of the It\^{o}  integral, we get the claim of the lemma. 
\end{proof} 

Next, we prove a few technical bound the drift of the discretized chain after $T/\eta$ discrete steps. The proofs follow similar calculations as those in \cite{dalalyan2016theoretical}.   

We will first need to bound the Hessian of $\tilde{f}$. 
\begin{lem}[Hessian bound] 
For all $x\in \R^d$, 
$$-2 (D \maxeig)^2 I \preceq \nabla^2 f(x) \preceq \maxeig I.$$ 
\label{l:hessianbound}
\end{lem}
\begin{proof}
For notational convenience, let $p(x) = \sum_{i=1}^m w_i e^{-f_i(x)}$. Note that $f(x) = - \log p(x)$. We proceed to the upper bound first. The Hessian of $f$ satisfies 
\begin{align*} \nabla^2 {f} 
&= \frac{\sum_i w_i e^{-f_i} \nabla^2 f_i}{p} - \frac{\frac{1}{2} \sum_{i,j} w_i w_j e^{-f_i} e^{-f_j} (\nabla f_i - \nabla f_j)^{\otimes 2}}{p^2} \\
&\preceq \max_i \nabla^2 f_i \preceq \maxeig I \end{align*}
as we need. 
As for the lower bound, we have 
\begin{align*} \nabla^2 {f} 
&\succeq -\rc 2\pa{\max_{i,j}  \ve{\nabla f_i - \nabla f_j}^{2}}I \end{align*}
But notice that since $f_i(x) = f_0(x - \mu_i)$, we have 
\begin{align*} \|\nabla f_i(x) - \nabla f_j(x)\| &= \|\nabla f_0(x-\mu_i) - \nabla f_0(x-\mu_j)\| \\
&\leq \maxeig \|\mu_i - \mu_j\| \\
&\leq 2 D \maxeig   \end{align*} 
where the next-to-last inequality follows from the strong-convexity of $f_0$. This
proves the statement of the lemma. 
\end{proof}

We introduce the following piece of notation in the following portion: we denote by $P_T\left(i,x\right)$ 
the measure on $[L] \times \mathbb{R}^d$ corresponding to running the Langevin diffusion process for $T$ time steps on the second coordinate, starting at $(i,x)$, and keeping the first coordinate fixed. Let us define 
 by $\widehat{P_T}\left(i,x\right): [L]\times \mathbb{R}^d \to \mathbb{R}$ the analogous  
measure, except running the discretized Langevin diffusion chain for $\frac{T}{\eta}$ time steps on the second coordinate, for $\frac{T}{\eta}$ an integer.    

\begin{lem}[Bounding interval drift] In the setting of this section, let $i\in [L], x \in \mathbb{R}^d$, and let $\eta \leq \frac{1}{\maxeig}$.
$$\mbox{KL}(P_T(i, x) || \widehat{P_T}(i,x)) \leq \frac{4 D^6 \eta^2 \maxeig^7}{3} \left(\|x - x^*\|_2^2 + 8 Td\right) + d T D^2 \eta \maxeig^2$$

\Anote{Doesn't seem we need the general $\alpha$ formulation -- setting it to 1 works?}
\label{l:intervaldrift}
\end{lem}    
\begin{proof}
Let $x_j, i \in [0, T/\eta - 1]$ be a random variable distributed as $\widehat{P_{\eta j}}(i,x)$. By Lemma 2 in \cite{dalalyan2016theoretical} and Lemma~\ref{l:hessianbound}
, we have 
\begin{equation*} \mbox{KL}(P_T(i, x) || \widehat{P_T}(i, x)) \leq \frac{\eta^3 D^2 \maxeig^2 }{3} \sum_{k=0}^{T/\eta - 1} \E[\|\nabla f(x^k)\|^2_2] + d T \eta D^2 \maxeig^2  \end{equation*}
Similarly, the proof of Corollary 4 in \cite{dalalyan2016theoretical} implies that%
\begin{equation*} \eta \sum_{k=0}^{T/\eta -1} \E[\|\nabla f(x^k)\|^2_2] \leq 4 D^4 \maxeig^4 \|x - x^*\|^2_2 + 8D \maxeig T d\end{equation*}
Plugging this in, we get the statement of the lemma. 
\end{proof} 


%

To prove the main claim, we will use Lemma~\ref{l:kl-mixture}, a decomposition theorem for the KL divergence of two mixtures of distributions, in terms of the KL divergence of the weights and the components in the mixture.
\begin{proof}[Proof of Lemma \ref{l:maindiscretize}]

Let's denote by $R\left(i,x\right)$ 
the measure on $[L]\times\mathbb{R}^d$, 
after one Type 2 transition in the simulated tempering process, starting at $(i,x)$.   

We will proceed by induction. Towards that, we can obviously write   
\begin{align*} 
p^{T_{i+1}} &= \frac{1}{2} \left( \int_{x \in \mathbb{R}^d} \sum_{j=0}^{L-1} p^{T_i}(j,x) P_{T_{i+1} - T_i}(j,x) \right) + \frac{1}{2} \left( \int_{x \in \mathbb{R}^d} \sum_{j=0}^{L-1}  p^{T_i}(j,x) R(j,x) \right) 
\end{align*} 
and similarly
\begin{align*} 
q^{T_{i+1}} &= \frac{1}{2} \left( \int_{x \in \mathbb{R}^d} \sum_{j=0}^{L-1} q^{T_i}(j,x) \widehat{P_{T_{i+1} - T_i}} (j,x) \right) + \frac{1}{2} \left( \int_{x \in \mathbb{R}^d} \sum_{j=0}^{L-1}  q^{T_i}(j,x) R(j,x) \right) 
\end{align*}
(Note: the $R$ transition matrix doesn't change in the discretized vs. continuous version.) 

By convexity of KL divergence, we have
\begin{align*}
\mbox{KL}(P^{T_{i+1}} || Q^{T_{i+1}}) &\leq \frac{1}{2} \mbox{KL}\left( \int_{x \in \mathbb{R}^d} \sum_{j=0}^{L-1} p^{T_i}(x,j) P_{T_{i+1} - T_i}(x, j) || \int_{x \in \mathbb{R}^d} \sum_{j=0}^{L-1} q^{T_{i}}(x,j) \widehat{P_{T_{i+1} - T_i}}(x, j) \right) \\ 
&+ \frac{1}{2} \mbox{KL}\left( \int_{x \in \mathbb{R}^d} \sum_{j=0}^{L-1}  p^{T_i}(x,j) R(x, j) || \int_{x \in \mathbb{R}^d} \sum_{j=0}^{L-1}  q^{T_i}(x,j) R(x,j)  \right)  
\end{align*} 

By Lemma~\ref{l:decomposingKL}, we have that 
$$\mbox{KL}\left( \int_{x \in \mathbb{R}^d} \sum_{j=0}^{L-1}  p^{T_i}(x,j) R(x,j) || \int_{x \in \mathbb{R}^d} \sum_{j=0}^{L-1}  q^{T_i}(x,j) R(x, j)  \right) \leq \mbox{KL} (P^{T_i} || Q^{T_i}). $$ 
Similarly, by Lemma~\ref{l:intervaldrift} together with Lemma~\ref{l:decomposingKL} we have
\begin{align*} & \mbox{KL}\left( \int_{x \in \mathbb{R}^d} \sum_{j=0}^{L-1} p^{T_i}(x,j) P_{T_{i+1} - T_i}(x, j) || \int_{x \in \mathbb{R}^d} \sum_{j=0}^{L-1} q^{T_i}(x,j) \widehat{P_{T_{i+1} - T_i}}(x, j) \right)  \leq \\ 
& \mbox{KL} (P^{T_i} || Q^{T_i}) + \frac{ 4 D^6 \maxeig^6 \eta^2}{3} \left( \max_j \E_{x \sim P^{t_i}( \cdot, j)}\|x - x^*\|_2^2 + 8 (T_{i+1} - T_i)d\right) + d (T_{i+1} - T_i) \eta \maxeig^2 \end{align*}

By Lemmas~\ref{l:reachcontinuous} and \ref{l:locatemin}, we have that for any $j \in [0, L-1]$, 
\begin{align*} \E_{x \sim P^{T_i} ( \cdot, j)}\|x - x^*\|_2^2 &\leq  \E_{x \sim P^{T_{i-1}}( \cdot, j)}\|x\|_2 + \left(400 \frac{D^2 \maxeig^2}{\mineig} + 2d\right) (T_i - T_{i-1}) \end{align*} 
Hence, inductively, we have $\E_{x \sim P^{T_i}( \cdot, j)}\|x - x^*\|_2^2 \leq \E_{x \sim P^0( \cdot, j)}\|x - x^*\|_2^2 +  \left(400 \frac{D^2 \maxeig^2}{\mineig} +2d\right) T_i$.

Putting everything together, we have 
\begin{align*}& \mbox{KL} (P^{T_{i+1}} || Q^{T_{i+1}}) \leq \mbox{KL}(P^{T_i} || Q^{T_i}) + \frac{4 \eta^2 D^6 \maxeig^7}{3} \cdot\\
& \left( \max_j \E_{x \sim P^0( \cdot, j)}\|x - x^*\|_2^2 + \left(400 \frac{D^2 D^2 \maxeig^2}{\mineig} + 2d\right) T + 8 Td \right) + d T \eta D^2 \maxeig^2 \end{align*}

By induction, we hence have 
\begin{align*} \mbox{KL} (P^T || Q^T) \lesssim \eta^2 D^6 \maxeig^7 \left(D^2 \frac{\maxeig^2}{\mineig}+d\right) T n + \eta^2 D^3 \maxeig^3 \max_i \E_{x \sim P^0( \cdot, i)}\|x - x^*\|_2^2 + \eta D^2 \maxeig^2 d T   \end{align*}
as needed.
\end{proof} 

%
%



\section{Proof of main theorem}
\label{sec:main-proof}

Before putting everything together, we show how to estimate the partition functions. 
We will apply the following to $g_1(x) = e^{-\be_{\ell}f(x)}$ and $g_2(x) = e^{-\be_{\ell+1} f(x)}$. 
\begin{lem}[Estimating the partition function to within a constant factor] 
Suppose that $P_1$ and $P_2$ are probability measures on $\Om$ with density functions (with respect to a reference measure) $p_1(x) =\fc{g_1(x)}{Z_1}$ and $p_2(x)=\fc{g_2(x)}{Z_2}$. 
Suppose $\wt P_1$ is a measure such that $d_{TV}(\wt P_1, P_1)<\fc{\ep}{2C^2}$, and $\fc{g_2(x)}{g_1(x)}\in [0, C]$ for all $x\in \Om$. Given $n$ samples $x_1,\ldots, x_n$ from $\wt P_1$, define the random variable
\begin{align}
\ol r = \rc{n} \sumo in \fc{g_2(x_i)}{g_1(x_i)}.
\end{align}
Let
\begin{align}
r = \EE_{x\sim P_1}\fc{g_2(x)}{g_1(x)} = \fc{Z_2}{Z_1}
\end{align}
and suppose $r\ge \rc{C}$. 
Then  with probability $\ge 1-e^{-\fc{n\ep^2}{2C^4}}$, 
\begin{align}
\ab{\fc{\ol r}{r}-1}& \le \ep.
\end{align}
\label{l:partitionfunc}
\end{lem}
\begin{proof}
We have that 
\begin{align}
\label{eq:est-Z1}
\ab{\EE_{x\sim \wt P_1} \fc{g_2(x)}{g_1(x)} - \EE_{x\sim P_1}\fc{g_2(x)}{g_1(x)}}
&\le Cd_{TV}(\wt P_1, P_1)\le \fc{\ep}{2C}.
\end{align}
The Chernoff bound gives
\begin{align}
\Pj
\pa{
\ab{r - \EE_{x\sim \wt P_1} \fc{g_2(x)}{g_1(x)}} \ge \fc{\ep}{2C}
}
&\le 
e^{-\fc{n\pf{\ep}{2C}^2}{2\pf{C}2^2}} = 
e^{-\fc{n\ep^2}{2C^4}}.\label{eq:est-Z2}
\end{align}
Combining \eqref{eq:est-Z1} and \eqref{eq:est-Z2} using the triangle inequality,
\begin{align}
\Pj\pa{|\ol r- r|\ge \rc{\ep}C} \le e^{-\fc{n\ep^2}{2C^4}}.
\end{align}
Dividing by $r$ and using $r\ge \rc{C}$ gives the result.
\end{proof}

\begin{lem}\label{lem:a1-correct}
Suppose that Algorithm~\ref{a:stlmc} is run on
$f(x) =-\ln \pa{\sumo jm w_j\exp\pa{-\fc{\ve{x-\mu_j}^2}{2\si^2}}}$ with 
 temperatures $0<\be_1<\cdots< \be_\ell\le 1$, $\ell\le L$, rate $\la$, and with partition function estimates $\wh{Z_1},\ldots, \wh{Z_\ell}$ satisfying
\begin{align}\label{eq:Z-ratio-correct}
\ab{\left.\fc{\wh{Z_i}}{Z_i}\right/\fc{\wh{Z_1}}{Z_1}}\in \ba{\pa{1-\rc L}^{i-1}, \pa{1+\rc L}^{i-1}}
\end{align} 
for all $1\le i\le \ell$.
Suppose $\sumo jm w_j=1$, $w_{\min}=\min_{1\le j\le m} w_i >0$, and $D=\max\{\max_{1\le j\le m} \ve{\mu_j}, \si\}$, and the parameters satisfy
\begin{align}
\la &= \Te\pf{1}{D^2}\\
\label{eq:beta1}
\be_1 &= \Te\pf{\si^2}{D^2}\\
\label{eq:beta-diff}
\fc{\be_{i+1}}{\be_{i}} &\le 
1+\rc {d+\ln\prc{w_{\min}}}\\
L &= \Te\pa{\pa{d+\ln\prc{w_{\min}}}\ln \pf{D}{\si}+1}\\
T&=\Om\pf{
L^2D^2\ln \pf{\ell}{\ep w_{\min{}}}}{w_{\min{}}^{\wexp}}\\
\eta &=
O\pa{
\fc{\si^3\ep}{D^2} \min \bc{
\fc{\si^4}{\pa{\fc{D}{\si} + \sqrt d}T}, 
\rc{D^{\rc 2}},
\fc{\si\ep}{dT}
}
}
\end{align}
Let $q^0$ be the distribution $\pa{N\pa{0,\fc{\si^2}{\be_1}}, 1}$ on $ [\ell]\times\R^d$. 
Then the distribution $q^T$ after running time $T$ satisfies $
\ve{p-q^T}_1\le \ep
$.

Setting $\ep = O\prc{\ell L}$ above and taking $n=\Om\pa{L^2\ln \prc{\de}}$ samples, with probability $1-\de$ the estimate 
\begin{align}\wh Z_{\ell+1}&=
\ol r\wh Z_\ell,&\ol r:&= {\rc{n}\sumo jn e^{(-\be_{\ell+1} + \be_\ell)f_i(x_j)}}
\end{align} also satisfies~\eqref{eq:Z-ratio-correct}.
\end{lem}
\begin{proof}
By the triangle inequality,
\begin{align}
\ve{p-q^T}_1 &\le \ve{p-p^T}_1 + \ve{p^T-q^T}_1.
\end{align}
For the first term, by Cauchy-Schwarz, (using capital letters for the probability measures)
\begin{align}\label{eq:chi-sq-final}
\ve{p-p^T}_1&\le\sqrt{ \chi^2(P^T||P)}\le  e^{-\fc{T}{2C}} \sqrt{\chi^2(P^0||P)}
\end{align}
where $C=O\pf{d^2\pa{\ln \pf{D}{\si}}^2 D^2}{w_{\min}^2}$ is an upper bound on the Poincar\'e constant as in Theorem~\ref{thm:st-gaussians2}. (The assumption on $\wh Z_i$ means that $r\le e$.)
Let $p_i$ be the distribution of $p$ on the $i$th temperature, and $\wt p_i$ be as in Lemma \ref{lem:close-to-sum}.

To calculate $\chi^2(P^0||P)$, first note by Lemma~\ref{lem:chi-squared-N}, the $\chi^2$ distance between $N\pa{0,\fc{\si^2}{\be_1}I_d}$ and $N(\mu, \fc{\si^2}{\be_1}I_d)$ is $\le e^{\ve{\mu}^2\be_1/\si^2}$.
Then
\begin{align}
&\quad \chi^2(P^0||P) \\
&=
O(\ell) \chi^2\pa{N\pa{0, \fc{\si^2}{\be_1}I_d} || P_1}\\
& = O\pf{\ell}{w_{\min}} \pa{1+\chi^2\pa{N\pa{0, \fc{\si^2}{\be_1}I_d}||\wt P_1}}\\
&\quad\text{by Lemma~\ref{lem:close-to-sum} and Lemma~\ref{lem:chi-liy}}\nonumber\\
&= O\pf{\ell}{w_{\min}}\pa{1+\sumo jm w_j\chi^2\pa{N\pa{0,  \fc{\si^2}{\be_1}I_d}||N\pa{\mu_j, \fc{\si^2}{\be_1}I_d}}}\label{eq:chi-warm-mix}\\
&\quad\text{by Lemma~\ref{lem:chi-sq-mixture}}\nonumber\\
&= O\pf{e^{\fc{D^2\be_1}{\si^2}}\ell}{w_{\min}} =O\pf{\ell}{w_{\min}}. 
\end{align}
Together with~\eqref{eq:chi-sq-final} this gives $\ve{p-p^T}_1\le \fc\ep 3$. 

For the second term $\ve{p^T-q^T}_1$, we first condition on there not being too many transitions before time $T$. Let $N_T=\max\set{n}{T_n\le T}$ be the number of transitions. 
Let $C$ be as in Lemma~\ref{lem:poisson-tail}. Note that $\pf{Cn}{T\la}^{-n}\le \ep\iff {e^{n\pf{T\la }{C}-\ln n}}\le \ep$, and that this inequality holds when $n\ge \fc{eT\la }{C}+\ln \prc{\ep}$. 
We have by Lemma~\ref{lem:poisson-tail} that $\Pj(N_T \ge  \fc{eT\la }{C}+\ln \prc{\ep}) \le \fc \ep3$. With our choice of $T$, $\ln \prc{\ep}=O(T)$.

If we condition on the event $A$ of the $T_i$'s being a particular sequence $T_1,\ldots, T_n$ with $n<\fc{eT\la }{C}+\ln \prc{\ep}$, 
Pinsker's inequality and Lemma~\ref{l:maindiscretize} (with $\maxeig=\mineig = \rc{\si^2}$) gives us 
\begin{align}
\ve{p^T(\cdot |A)-q^T(\cdot |A)}_1
&\le 
 \sqrt{2\KL(P^t(\cdot |A)||Q^t(\cdot |A))}\\
 &=O\pa{\max\bc{
 	\fc{\eta^2 D^6 T^2\la}{\si^{14}\pa{\fc{D^2}{\si^2}+d}}, \eta^2 D^3 \rc{\si^6} D^2, \eta D^2 \rc{\si^4}dT
 }} 
\end{align}
In order for this to be $\le \fc{\ep}{3}$, we need (for some absolute constant $C_1$)
\begin{align}
\eta&\le 
\fc{C_1 \si^3 \ep}{D^2}\min
\bc{
\fc{\si^4 
}{\pa{\fc{D}{\si}+\sqrt d}T}, \rc{D^{\rc 2}}, \fc{\si\ep}{dT}
}.
\end{align}

Putting everything together,
\begin{align}
\ve{p^T-q^T}_1 &\le \Pj(N_T\ge cT\la) + \ve{p^t(\cdot | N_T\ge c T\la) - q^t(\cdot |N_T\ge cT\la)}_1\le \fc\ep3+\fc\ep3 = \fc{2\ep}3.
\end{align}
This gives $\ve{p-q^T}_1 \le\ep$.

For the second part, setting $\ep=O\prc{\ell L}$ gives that $\ve{p_\ell - q^T_\ell}_1=O\prc{L}$.
We will apply Lemma~\ref{l:partitionfunc}. By Lemma~\ref{lem:delta} the assumptions of Lemma~\ref{l:partitionfunc} are satisfied with $C=O(1)$, as we have
\begin{align}
\fc{\be_{i+1}-\be_i}{\be_i} &=O\prc{\al \fc{D^2}{\si^2} + d + \ln\prc{w_{\min}}}.
\end{align}
By Lemma~\ref{l:partitionfunc}, 
 after collecting $n=\Om\pa{L^2\ln \prc{\de}}$ samples, with probability $\ge 1-\de$, $\ab{\fc{\wh{Z_{\ell+1}}/\wh{Z_\ell}}{Z_{\ell+1}/Z_\ell}-1}\le \rc L$. 
Set $\wh{Z_{\ell+1}} = \ol r\wh{Z_\ell}$. Then 
$\fc{\wh{Z_{\ell+1}}}{\wh{Z_\ell}} \in [1-\rc L , 1+\rc L] \fc{Z_{\ell+1}}{Z_\ell}$ and 
$\fc{\wh{Z_{\ell+1}}}{\wh{Z_1}} \in 
\ba{\pa{1-\rc L}^\ell, \pa{1+\rc L}^\ell}\fc{Z_{\ell+1}}{Z_1}$.
\end{proof}

\begin{proof}[Proof of Theorem~\ref{thm:main}]
Choose $\de=\fc{\ep}{2L}$ where $L$ is the number of temperatures. 
Use Lemma~\ref{lem:a1-correct} inductively, with probability $1-\fc{\ep}2$ each estimate satisfies $\fc{\wh{Z_\ell}}{\wh{Z_1}}\in [\rc e,e]$. Estimating the final distribution within $\fc{\ep}2$ accuracy gives the desired sample.
\end{proof}

\section{Conclusion}

We initiated a study of sampling ``beyond log-convexity." In so doing, we developed a new general technique to analyze simulated tempering, a classical algorithm used in practice to combat multimodality but that has seen little theoretical analysis. The technique is a new decomposition lemma for Markov chains based on decomposing the \emph{Markov chain} rather than just the \emph{state space}. We have analyzed simulated tempering with Langevin diffusion, but note that it can be applied to any with any other Markov chain with a notion of temperature.

Our result is the first result in its class (sampling multimodal, non-log-concave distributions with gradient oracle access). Admittedly, distributions encountered in practice are rarely mixtures of distributions with the same shape. However, we hope that our techniques may be built on to provide guarantees for more practical probability distributions. An exciting research direction is to provide (average-case) guarantees for probability  distributions encountered in practice, such as posteriors for clustering, topic models, and Ising models. For example, the posterior distribution for a mixture of gaussians can have exponentially many terms, but may perhaps be tractable in practice. Another interesting direction is to study other temperature heuristics used in practice, such as particle filters \cite{schweizer2012non,del2012concentration,paulin2015error,giraud2017nonasymptotic}, annealed importance sampling \cite{neal2001annealed}, and parallel tempering \cite{woodard2009conditions}.
We note that our decomposition theorem can also be used to analyze simulated tempering in the infinite switch limit \cite{martinsson2019simulated}.
\section*{Acknowledgements} 

This work was done in part while the authors were visiting the Simons Institute for the Theory of Computing. We thank Matus Telgarsky and Maxim Raginsky for illuminating conversations in the early stages of this work. Rong Ge acknowledges funding from NSF CCF-1704656.

\printbibliography
\appendix

\iftoggle{thesis}{\section{Markov chains and processes}}{
\section{Background on Markov chains and processes}}\label{sec:mc}
%
%
%

\label{sec:mdo}

Throughout, we will use upper-case $P$ for probability measures, and lower-case $p$ for the corresponding density function (although we will occasionally abuse notation and let $p$ stand in for the measure, as well).

A discrete-time 
(time-invariant)
Markov chain on $\Om$ is a probability law on a sequence of random variables $(X_t)_{t\in \N_0}$ taking values in $\Om$, such that the next state $X_{t+1}$ only depends on the previous state $X_t$, in a fixed way. More formally, letting $\cal F_t = \si((X_t)_{0\le s\le t})$, there is a \vocab{transition kernel} $T$ on $\Om$ (i.e. $T(x,\cdot)$ is a probability measure and $T(\cdot , A)$ is a measurable function for any measurable $A$) such that 
\begin{align}
\Pj(X_{t+1}\in \cdot |\cal F_t) &= T(X_t,\cdot).
\end{align}
A stationary measure is $P$ such that if $X_0\sim P$, then $X_t\sim P$ for all $t$. The idea of Markov chain Monte Carlo is to design a Markov chain whose stationary distribution is $P$ with good mixing; that is, if $\pi_t$ is the probability distribution at time $t$, then $\pi_t\to P$ rapidly as $t\to \iy$.

The Markov chains we consider will be discretized versions of continuous-time Markov processes, so we will mainly work with Markov processes (postponing discretization analysis until the end). 

\ift{\nomenclature[1Pt]{$P_t$}{Family of kernels defining Markov process}}
\ift{\nomenclature[1Ptscript]{$\sP_t$}{$\sP_tg(x)=\E_{y\sim P_t(x,\cdot)} g(y) = \int_{\Om} g(y)P_t(x,dy)$}}
\ift{\nomenclature[1DL]{$\cal D(\sL)$}{Domain of $\sL$}}
Instead of being defined by a single transition kernel $T$, a continuous time Markov process is instead defined by a family of kernels $(P_t)_{t\ge 0}$, and a more natural object to consider is the generator. 
\begin{df}\label{d:mp}
A (continuous-time, 
time-invariant) \vocab{Markov process} is given by $M=(\Om, (P_t)_{t\ge 0})$, where each $P_t$ is a transition kernel. It defines the random process $(X_t)_{t\ge 0}$ by
$$
\Pj(X_{s+t}\in A|\cal F_s) =
\Pj(X_{s+t}\in A|X_s) =
 P_t(X_s,A) =\int_A P_t(x,dy)
$$
where $\cal F_s=\si((X_r)_{0\le r\le s})$.  
Define the action of $\sP_t$ on functions by
\begin{align}
(\sP_t g)(x) &= \E_{y\sim P_t(x,\cdot)} g(y) = \int_{\Om} g(y)P_t(x,dy).
\end{align}

A \vocab{stationary measure} is $P$ such that if $X_0\sim P$, then $X_t\sim P$ for all $t$. 
A Markov process with stationary measure $P$ is \vocab{reversible} if $\sP_t$ is self-adjoint with respect to $L^2(P)$, i.e., as measures $P(dx)P_t(x,dy) = P(dy)P_t(y,dx)$.

Define the \vocab{generator} $\sL$ by
\begin{align}
\sL g &= \lim_{t\searrow 0} \fc{\sP_t g - g}{t},
\end{align}
and let $\cal D(\sL)$ denote the space of $g\in L^2(P)$ for which $\sL g\in L^2(P)$ is well-defined.
If $P$ is the unique stationary measure, define the \vocab{Dirichlet form} and the variance by
\begin{align}
\sE_M(g,h) &= -\an{g, \sL h}_P\\
\Var_P(g) &= \ve{g-\int_\Om g \,P(dx)}_{L^2(P)}^2
\end{align}
\end{df}
Note that in order for $(P_t)_{t\ge 0}$ to be a valid Markov process, it must be the case that $\sP_t\sP_u g = \sP_{t+u}g$, i.e., the $(\sP_t)_{t\ge 0}$ forms a \vocab{Markov semigroup}. All the Markov processes we consider will be reversible.

We will use the shorthand $\sE(g):=\sE(g,g)$. 

\begin{df}
A continuous Markov process (given by Definition~\ref{d:mp}) satisfies a \vocab{Poincar\'e inequality with constant $C$} if for all $g$ such that $g\in \cal D(\sL)$, 
\begin{align}
\sE_M(g,g) \ge \rc C\Var_P(g).
\end{align}
\end{df}
We will implicitly assume $g\in \cal D(\sL)$ every time we write $\sE_M(g,g)$. 
The minimal $\rh$ such that $\sE_M(g,g)\ge \rh \Var_P(g)$ for all $g$ is the \vocab{spectral gap} $\Gap(M)$ of the Markov process.

A Poincar\'e inequality implies rapid mixing: If $C$ is maximal such that $M$ satisfies a Poincar\'e inequality with constant $C$, it can be shown that\footnote{Note the subtle point that the Poincar\'e inequality as we defined it only makes sense for $g\in \cal D(\sL)$, whereas~\eqref{e:pi-intro} makes sense when $g\in L^2(P)$. For Langevin diffusion, it suffices to show the Poincar\'e inequality for $g\in \cal D(\sL)$ to obtain~\eqref{e:pi-intro} for all $g\in L^2(P)$. See~\cite{bakry2013analysis}. This will, however, not be an issue for us because we will start with a measure $\pi_0$ with smooth density.} 
\begin{align}\label{e:pi-intro}
\ve{\sP_t g - \E_P g}_{L^2(P)}^2\le e^{-t\Gap(M)} \ve{g - \E_P g}_{L^2(P)}^2= e^{-\fc tC}\ve{g - \E_P g}_{L^2(P)}^2.
\end{align}
We can turn this into a statement about probability distributions, as follows.
If the probability distribution at time $t$ is $\pi_t$, then setting $g=\dd{\pi_0}{P}$ (the Radon-Nikodym derivative) and assuming $\ve{\dd{\pi_0}{P}}_{L^2(P)}<\iy$, we have 
\begin{align}
\an{\sP_t f, \dd{\pi_0}{P}}_{L^2(P)} &= \int_{\Om} \sP_tf(x) \pi_0(dx) = \int_\Om\int_\Om f(y)P_t(x,dy)\pi_0(dx) \\
&= \int_{\Om} f(y)\pi_t(dy)
=\an{f,\dd{\pi_t}{P}}_{L^2(P)}.
\end{align}
If the Markov process is reversible, then $\an{\sP_tf, \dd{\pi_0}{P}}_{L^2(P)}=\an{f,\sP_t\dd{\pi_0}{P}}_{L^2(P)}$. Hence for all $f$, $\an{f,\sP_t\dd{\pi_0}{P}}_{L^2(P)}=\an{f,\dd{\pi_t}{P}}_{L^2(P)}$, 
so $\sP_t\dd{\pi_0}{P}=\dd{\pi_t}{P}$. The $\chi^2$ divergence is defined by $\chi^2(Q||P)=\ve{\dd QP-1}_{L^2(P)}^2$, so putting $g=\dd{\pi_0}{P}$ in~\eqref{e:pi-intro} gives
\begin{align}
\chi^2(\pi_t||P) &\le e^{-\fc tC} \chi^2(\pi_0||P).
\end{align}

The following gives one way to prove a Poincar\'e inequality.

\begin{thm}[Comparison theorem using canonical paths, \cite{diaconis1991geometric}]\label{thm:can-path}
Suppose $\Om$ is finite. 
Let $T:\Om\times \Om\to \R$ be a function with $T(x,y)\ge 0$ for $y\ne x$ and $\sum_{y\in \Om} T(x,y)=1$. (Think of $T$ as a matrix in $\R^{\Om\times \Om}$ that operates on functions $g:\Om\to \R$, i.e., $g\in \R^{\Om}$.) Let $L=T-I$, so that $L(x,y)=T(x,y)$ for $y\ne x$ and $L(x,x) = -\sum_{y\ne x}T(x,y)$. 

Consider the Markov process $M$ generated by $L$ ($L$ acts as $Lg(j) = \sum_{k\ne j} [g(k)-g(j)]T(j,k)$); let its Dirichlet form be $\sE(g,g) = -\an{g,Lg}$ and stationary distribution be $p$.

Suppose each pair $x,y\in \Om$, $x\ne y$ is associated with a path $\ga_{x,y}$. Define the congestion to be
$$
\rh(\ga) = \max_{z,w\in \Om, z\ne w} \ba{
\fc{\sum_{\ga_{x,y}\ni (z,w) }|\ga_{x,y}|p(x)p(y)}{p(z)T(z,w)}
}
$$
where $|\ga|$ denotes the length of path $\ga$.
Then
$$
\Var_p(g) \le \rh(\ga) \sE(g,g).
$$
\end{thm}
\begin{proof}
Note that the statement in \cite{diaconis1991geometric} is for discrete-time Markov chains; we show that our continuous-time result is a simple consequence.

Let $\ep>0$ be small enough such that $T_\ep = I + \ep L = I+\ep(T-I) = (1-\ep)I + \ep T$ has all entries $\ge 0$.\footnote{Alternatively, note that nothing in their proof actually depends on $T$ having all entries $\ge 0$, so taking $\ep=1$ is fine.}
Note that the stationary distribution for $M$ is the same as the stationary distribution of the discrete-time Markov chain generated by $T_\ep$, namely, the (appropriately scaled) eigenvector of $T$ corresponding to the eigenvalue 1. The Dirichlet form for $T_\ep$ is $-\an{g,(T_\ep-I)g} = -\ep\an{g,Lg}=\ep\sE(g,g)$.

Applying~\cite[Proposition $1'$]{diaconis1991geometric} to $T_\ep$ (note $T_\ep(z,w)=\ep T(z,w)$ for $z\ne w$) gives 
\begin{align}
\Var_p(g) &\le  \max_{z,w\in \Om, z\ne w} \ba{
\fc{\sum_{\ga_{x,y}\ni (z,w) }|\ga_{x,y}|p(x)p(y)}{p(z)\ep T(z,w)}
} ({\ep}\sE(g,g)) = \rh(\ga)\sE(g,g).
\end{align}
\end{proof}

\iftoggle{thesis}{\section{Langevin diffusion}}
{\subsection{Langevin diffusion}}

\iftoggle{thesis}{Langevin Monte Carlo is an algorithm for sampling from a measure $P$ with density function $p(x)\propto e^{-f(x)}$ given access to the gradient of the log-pdf, $\nabla f$. We will always assume that $\int_{\R^d} e^{-f(x)}\dx<\iy$ and $f\in C^2(\R^d)$.

The continuous version, overdamped Langevin diffusion (often simply called Langevin diffusion), is a stochastic process described by the stochastic differential equation
\begin{equation}
dX_t = -\nb f (X_t) \,dt + \sqrt{2}\,dW_t 
\end{equation}
where $W_t$ is the Wiener process (Brownian motion). 
For us, the crucial 
fact  is that Langevin dynamics converges to the stationary distribution given by $p(x) \propto e^{-f(x)}$.

The Dirichlet form is}{For Langevin diffusion with stationary measure $P$,}
\begin{align}
\sE_M(g,g) &= \ve{\nb g}_{L^2(P)}^2. 
\end{align}
Since this depends in a natural way on $P$, we will also write this as $\sE_P(g,g)$. A Poincar\'e inequality for Langevin diffusion thus takes the form
\begin{align}
\sE_P(g,g) = \int_{\Om} \ve{\nb g}^2 P(dx) &\ge \rc C \Var_P(g).
\end{align}
Showing mixing for Langevin diffusion reduces to showing such an inequality. If strongly log-concave measures, this is a classical result. 
\begin{thm}[\cite{bakry2013analysis}]\label{thm:bakry-emery}
Let $f$ be $\rh$-strongly convex and differentiable. 
Then for $g\in \cal D(\sE_P)$, 
the measure $P$ with density function $p(x)\propto e^{-f(x)}$ satisfies the Poincar\'e inequality
$$
\sE_P(g,g) \ge \rh \Var_P(g).
$$
\end{thm}
In particular, this holds for $f(x)={\fc{\ve{x-\mu}^2}{2}}$ with $\rh = 1$, giving a Poincar\'e inequality for the Gaussian distribution.

\section{General log-concave densities}
\label{sec:gen}
In this section we generalize the main theorem from gaussian to log-concave densities.

\subsection{Simulated tempering for log-concave densities} 
First we rework Section~\ref{sec:st-gaussian} for log-concave densities.

\begin{thm}[cf. Theorem \ref{thm:st-gaussians}]\label{thm:st-general}
Suppose $f_0$ satisfies Assumption~\ref{asm}(2) ($f_0$ is $\mineig$-strongly convex, $\maxeig$-smooth, and has minimum at 0).

Let $M$ be the continuous simulated tempering chain for the distributions
\begin{align}
p_i&\propto \sumo jm w_j e^{-\be_if_0(x-\mu_j)}
\end{align}
with rate $\Om\pf{r}{D^2}$, relative probabilities $r_i$, and temperatures $0<\be_1<\cdots <\be_L=1$ where
\begin{align}
D&=\max\bc{\max_j\ve{\mu_j} ,\fc{\mineig^{\rc 2}}{d^{\rc 2}\maxeig}}\\
\be_1 &= \Te\pf{\mineig}{d \maxeig^2 D^2
}\\
\fc{\be_{i+1}}{\be_i} &\le 1+\fc{\mineig}{\maxeig d
\pa{\ln\pf{\maxeig}{\mineig}+1}
}\\
L &= \Te\pa{\fc{\maxeig d \pa{\ln \pf{\maxeig}{\mineig}+1}^2}{\mineig} \ln \pf{d\maxeig D
}{\mineig}}\\
r&=\fc{\min_i r_i}{\max_i r_i}.
\end{align}
Then $M$ satisfies the Poincar\'e inequality
\begin{align}
\Var(g) &\le O\pf{L^2D^2}{r^2}\sE(g,g) = O\pf{\maxeig^2 D^2
 \ln\pa{\ln \pf{\maxeig}{\mineig}+1}^4\ln \pf{d\maxeig D}{\mineig}^2
  }{\mineig^2 r^2}\sE(g,g).
\end{align}
\end{thm}
\begin{proof}
Note that forcing $D\le \fc{\mineig^{\rc 2}}{d^{\rc 2}\maxeig}$ ensures $\be_1=\Om(1)$. 

The proof follows that of Theorem~\ref{thm:st-gaussians}, except that we need to use Lemmas~\ref{lem:chisq-temp} and \ref{lem:chisq-translate} to bound the $\chi^2$-divergences.
Steps 1 and 2 are the same: we consider the decomposition where $p_{i,j}\propto e^{-\be_i f_0(x-\mu_j)}$ and note $\sE_{i,j}$ satisfies the Poincar\'e inequality
\begin{align}
\Var_{p_{i,j}}(g_i) &\le \fc{1}{\mineig \be_i}\sE_{i,j} = O(D^2) \sE_{i,j} (g_i,g_i).
\end{align}

By Lemma \ref{lem:chisq-temp},
\begin{align}
\chi^2(p_{i-1,j}||p_{i,j}) &\le 
 e^{\rc 2\ab{1-\fc{\be_{i-1}}{\be_i}} \fc{\maxeig d}{\mineig  -\maxeig\ab{1-\fc{\be_{i-1}}{\be_i}}} \pa{\sqrt{\ln \pf{\maxeig}{\mineig}}+5}^2}\\
 &\cdot 
\pa{
\pa{1-\fc{\maxeig}{\mineig}\ab{1-\fc{\be_{i-1}}{\be_i}}}\pa{1+\ab{1-\fc{\be_{i-1}}{\be_i}}}
}^{-\fc d2}-1=O(1).
\end{align}
By Lemma \ref{lem:chisq-translate},
\begin{align}
\chi^2(p_{1,j'}||p_{1,j})
&\le 
e^{\rc 2 \be_1\mineig (2D)^2 +\sqrt{\be_1} \maxeig (2D)\sfc{d}{\mineig}\pa{
\sqrt{\ln \pf{\maxeig}{\mineig}}+5}
}\\
&\quad \cdot \pa{
e^{\maxeig (2D)\sfc{d}{\mineig}}
+
\sqrt{\be_1}\maxeig (2D) \sfc{4\pi}{\mineig} e^{ \fc{2\sqrt{\be_1}\maxeig (2D) \sqrt d}{\sqrt{\mineig}} + \fc{\be_1\maxeig^2 (2D)^2}{2\mineig}}}-1=O(1).
\end{align}
The rest of the proof is the same.
\end{proof}

\begin{thm}[cf. Theorem \ref{thm:st-gaussians2}]\label{thm:st-general2}
Suppose $f_0$ satisfies Assumption~\ref{asm}(2) ($f_0$ is $\mineig$-strongly convex, $\maxeig$-smooth, and has minimum at 0).

Suppose $\sumo jm w_j=1$, $w_{\min}=\min_{1\le j\le m} w_i >0$, and $D=\max_{1\le j\le m} \ve{\mu_j}$. 
Let $M$ be the continuous simulated tempering chain for the distributions
\begin{align}
p_i&\propto \pa{\sumo jm w_j e^{-f_0(x-\mu_j)}}^{\be_i}
\end{align}
with rate $O\pf{r}{D^2}$, relative probabilities $r_i$, and temperatures $0<\be_1<\cdots <\be_L=1$ 
satisfying the same conditions as in Theorem~\ref{thm:st-general}.
Then $M$ satisfies the Poincar\'e inequality
\begin{align}
\Var(g) &\le O\pf{L^2D^2}{r^2w_{\min}^{\wexp}}\sE(g,g) = 
 O\pf{\maxeig^2 d^2
 \pa{\ln \pf{\maxeig}{\mineig}+1}^4
  \ln \pf{d\maxeig D}{\mineig}^2
  }{\mineig^2 r^2w_{\min{}}^{\wexp}}\sE(g,g).
\end{align}
\end{thm}
\begin{proof}
Let $\wt p_i$ be the probability distributions in Theorem~\ref{thm:st-gaussians} with the same parameters as $p_i$ and let $\wt p$ be the stationary distribution of that simulated tempering chain. By Theorem~\ref{thm:st-general}, $\Var_{\wt p}(g) = O\pf{L^2D^2}{r^2} \sE\wt p(g,g)$. Now use 
By Lemma~\ref{lem:close-to-sum}, $\fc{p_i}{\wt p_i} \in \ba{1,\rc{w_{\min{}}}} \fc{\wt Z_i}{Z_i}$. Now use Lemma~\ref{lem:st-comparison} with $e^\De = \rc{w_{\min}}$.
\end{proof}
\subsection{Proof of main theorem for log-concave densities}
Next we rework Section~\ref{sec:main-proof} for log-concave densities, and prove the main theorem for log-concave densities, Theorem~\ref{thm:mainlogconcave}.

\begin{lem}[cf. Lemma \ref{lem:a1-correct}]\label{lem:a1-correct-gen}
Suppose $f_0$ satisfies Assumption~\ref{asm}(2) ($f_0$ is $\mineig$-strongly convex, $\maxeig$-smooth, and has minimum at 0).

Suppose that Algorithm~\ref{a:stlmc} is run on
$f(x) =-\ln \pa{\sumo jm w_jf_0(x-\mu_j)}$ with 
 temperatures $0<\be_1<\cdots< \be_\ell\le 1$, $\ell\le L$ with partition function estimates $\wh{Z_1},\ldots, \wh{Z_\ell}$ satisfying
\begin{align}\label{eq:Z-ratio-correct-gen}
\ab{\left.\fc{\wh{Z_i}}{Z_i}\right/\fc{\wh{Z_1}}{Z_1}}\in \ba{\pa{1-\rc L}^{i-1}, \pa{1+\rc L}^{i-1}}
\end{align} 
for all $1\le i\le \ell$.
Suppose $\sumo jm w_j=1$, $w_{\min}=\min_{1\le j\le m} w_i >0$, and $D=\max\bc{\max_j\ve{\mu_j} ,\fc{\mineig^{\rc 2}}{d^{\rc 2}\maxeig}}$, $K\ge 1$, and the parameters satisfy
\begin{align}
\la &= \Te\prc{D^2}\\
\label{eq:beta1-gen}
\be_1 &= O\pf{\mineig }{d \maxeig^2 D^2}\\
\label{eq:beta-diff-gen}
\fc{\be_{i+1}}{\be_{i}} &\le 
1+\fc{\mineig}{\maxeig d\pa{\ln\pf{\maxeig}{\mineig}+1}}\\
L&= \Te\pa{\fc{\maxeig d \pa{\ln \pf{\maxeig}{\mineig}+1}^2}{\mineig} \ln \pf{d\maxeig D
}{\mineig}}\\
T&=
\pa{
\fc{L^2D^2}{w_{\min}^{\wexp}}
  d\ln\pf{\ell}{\ep w_{\min}}\ln \pf{\maxeig}{\mineig}}\\
\eta &=O\pa{\min\bc{
\fc{\ep}{D^2\maxeig^{\fc 72} 
\pa{D \fc{\maxeig}{\mineig^{\rc 2}}+d^{\rc 2}}T},
\fc{\ep}{D^{\fc 52}\maxeig^{\fc 32} \pa{\pf{\maxeig}{\mineig}^{\rc 2}+1}}, 
\fc{\ep}{D^2\maxeig^2 dT}
}}.
\end{align}
Let $q^0$ be the distribution $\pa{N\pa{0,\fc{1}{\mineig\be_1}}, 1}$ on $ [\ell]\times\R^d$. 
The distribution $q^T$ after running time $T$ satisfies $
\ve{p-q^T}_1\le \ep
$.

Setting $\ep = O\prc{\ell L}$ above and taking $n=\Om\pa{L^2\ln \prc{\de}}$ samples, with probability $1-\de$ the estimate 
\begin{align}\wh Z_{\ell+1}&=
\ol r\wh Z_\ell,&\ol r:&= \pa{\rc{n}\sumo jn e^{(-\be_{\ell+1} + \be_\ell)f_i(x_j)}}
\end{align} also satisfies~\eqref{eq:Z-ratio-correct-gen}.
\end{lem}

\begin{proof}
Begin as in the proof of Lemma~\ref{lem:a1-correct}. Let $p_{\be,i}\propto e^{-\be_1 f_0(x-\mu_i)}$ be a probability density function. 

Write $\ve{p-q^T}_1 \le \ve{p-p^T}_1+\ve{p^T-q^T}_1$. Bound the first term by $\ve{p-p^T}_1\le \sqrt{\chi^2(P^T||P)} \le e^{-\fc{T}{2C}}\sqrt{\chi^2(P^0||P)}$ where $C$ is the upper bound on the Poincar\'e constant in Theorem~\ref{thm:st-general2}. 
As in~\eqref{eq:chi-warm-mix}, we get
\begin{align}
\chi^2(p||p^0) &=O\pf{\ell}{w_{\min}}
\pa{1+\sumo jm w_j \chi^2\pa{N\pa{ 0, \rc{\mineig \be_1}I_d}|| p_{\be_1,j} }}.
\end{align}
By Lemma~\ref{lem:chi-warm} with strong convexity constants $\mineig\be_1$ and $\maxeig \be_1$, this is
\begin{align}
O\pa{\fc{\ell}{w_{\min}} \pf{\maxeig}{\mineig}^{\fc d2} e^{\maxeig \be_1 D^2}} = O\pa{\fc{\ell}{w_{\min}}\pf{\maxeig}{\mineig}^{\fc d2}
}
\end{align}
when $\be_1 = O\pf{\maxeig}{D^2}$.
Thus for $T=\Om\pa{C\ln \pf{\ell}{\ep w_{\min}}d\ln \pf{\maxeig}{\mineig}}$, $\ve{p-p^T}_1\le\fc{\ep}3$.

Again conditioning on the event $A$ that $N_T=\max\set{n}{T_n\le T}=O(T\la)$, we get by Lemma~\ref{l:maindiscretize} that \begin{align}
\ve{p^T(\cdot|A) - q^T(\cdot|A)}_1&=
O\pa{
\eta^2 D^6 \maxeig^7 \pa{D^2 \fc{\maxeig^2}{\mineig}+d}Tn + \eta^2 D^5 \pa{ \fc{\maxeig}{\mineig}+1} + \eta D^2\maxeig^2dT
}.
\end{align}
Choosing $\eta$ as in the problem statement, we get $\ve{p-q^T}_1\le \ep$ as before. Finally, apply Lemma~\ref{l:partitionfunc}, checking the assumptions are satisfied using Lemma~\ref{lem:delta-gen}. The assumptions of Lemma~\ref{lem:delta-gen} hold, as
\begin{align}
\fc{\be_{i+1}-\be_i}{\be_i}
&= 
O\prc{\al \maxeig D^2 + \fc{d}{\mineig}\pa{1+\ln \pf{\maxeig}{\mineig}} + \rc{\mineig}\ln\prc{w_{\min}}}.
\end{align}
\end{proof}

\begin{proof}[Proof of Theorem~\ref{thm:mainlogconcave}]
This follows from Lemma~\ref{lem:a1-correct-gen} in exactly the same way that the main theorem for gaussians (Theorem~\ref{thm:main}) follows from Lemma~\ref{lem:a1-correct}.
\end{proof}

\section{Perturbation tolerance}
\label{sec:perturb}

The proof of Theorem~\ref{thm:perturb} will follow immediately from Lemma \ref{lem:a1-correct-perturb}, which is a straightforward analogue of Lemma \ref{lem:a1-correct}. 

\subsection{Simulated tempering for distribution with perturbation}

First, we consider the mixing time of the continuous tempering chain, analogously to Theorem \ref{thm:st-general2}: 

\begin{thm}[cf. Theorem \ref{thm:st-general2}]\label{thm:st-general2_perturb}
Suppose $f_0$ satisfies Assumption~\ref{asm} 

Let $M$ be the continuous simulated tempering chain 
with rate $O\pf{r}{D^2}$, relative probabilities $r_i$, and temperatures $0<\be_1<\cdots <\be_L=1$ 
satisfying the same conditions as in Lemma~\ref{lem:a1-correct-perturb}. 
Then $M$ satisfies the Poincar\'e inequality
\begin{align}
\Var(g) &\le O\pf{L^2D^2}{r^2w_{\min}^2}\sE(g,g) = 
 O\pf{\maxeig^2 d^2
 \pa{\ln \pf{\maxeig}{\mineig}+1}^4
  \ln \pf{d\maxeig D}{\mineig}^2
  }{\mineig^2 r^2e^{{\wexp}\Delta} w_{\min{}}^{\wexp}}\sE(g,g).
\end{align}
\end{thm}
\begin{proof}
The proof is almost the same as 
Let $\wt p_i$ be the probability distributions in Theorem~\ref{thm:st-gaussians} with the same parameters as $p_i$ and let $\wt p$ be the stationary distribution of that simulated tempering chain. By Theorem~\ref{thm:st-general}, $\Var_{\wt p}(g) = O\pf{L^2D^2}{r^2} \sE\wt p(g,g)$. Now use 
By Lemma~\ref{lem:close-to-sum}, $\fc{p_i}{\wt p_i} \in \ba{1,\rc{w_{\min{}}}} \fc{\wt Z_i}{Z_i}$. Now use Lemma~\ref{lem:st-comparison} with $e^\De$ substituted to be $e^{\De}\rc{w_{\min}}$.
\end{proof}

\subsection{Proof of main theorem with perturbations}

\begin{lem}[cf. Lemma \ref{lem:a1-correct-gen}]\label{lem:a1-correct-perturb}
Suppose that Algorithm~\ref{a:stlmc} is run on
$f(x) =-\ln \pa{\sumo jm w_jf_0(x-\mu_j)}$ with 
 temperatures $0<\be_1<\cdots< \be_\ell\le 1$, $\ell\le L$ with partition function estimates $\wh{Z_1},\ldots, \wh{Z_\ell}$ satisfying
\begin{align}\label{eq:Z-ratio-correct-gen-perturb}
\ab{\left.\fc{\wh{Z_i}}{Z_i}\right/\fc{\wh{Z_1}}{Z_1}}\in \ba{\pa{1-\rc L}^{i-1}, \pa{1+\rc L}^{i-1}}
\end{align} 
for all $1\le i\le \ell$.
Suppose $\sumo jm w_j=1$, $w_{\min}=\min_{1\le j\le m} w_i >0$, and $D=\max\bc{\max_j\ve{\mu_j} ,\fc{\mineig^{\rc 2}}{d^{\rc 2}\maxeig}}$, $K\ge 1$, and the parameters satisfy
\begin{align}
\la &= \Te\prc{D^2}\\
\label{eq:beta1-gen-perturb}
\be_1 &= O\left(\min\bc{\Delta, \frac{\mineig }{d \maxeig^2 D^2}}\right)\\
\label{eq:beta-diff-gen}
\fc{\be_{i+1}}{\be_{i}} &\le 
\min\bc{\Delta, 1+\fc{\mineig}{\maxeig d\pa{\ln\pf{\maxeig}{\mineig}+1}}}\\
L&= \Te\pa{\fc{\maxeig d \pa{\ln \pf{\maxeig}{\mineig}+1}^2}{\mineig} \ln \pf{d\maxeig D
}{\mineig}}\\
T&=
\pa{e^{{\wexp}\Delta}
\fc{L^2D^2}{w_{\min}^{\wexp}}
  d\ln\pf{\ell}{\ep w_{\min}}\ln \pf{\maxeig}{\mineig}}\\
\eta &=O\pa{\min\bc{
\fc{\ep}{D^2(\maxeig+\delsmall)^{\fc 72}
\pa{D \fc{\maxeig+\delsmall}{\mineig^{\rc 2}}+d^{\rc 2}}T},
\fc{\ep}{D^{\fc 52}(\maxeig+\delsmall)^{\fc 32} \pa{\pf{\maxeig+\delsmall}{\mineig}^{\rc 2}+1}}, 
\fc{\ep}{D^2(\maxeig+\delsmall)^2 dT}
}}.
\end{align}
Let $q^0$ be the distribution $\pa{N\pa{0,\fc{1}{\mineig\be_1}}, 1}$ on $ [\ell]\times\R^d$. 
The distribution $q^T$ after running time $T$ satisfies $
\ve{p-q^T}_1\le \ep
$.

Setting $\ep = O\prc{\ell L}$ above and taking $n=\Om\pa{L^2\ln \prc{\de}}$ samples, with probability $1-\de$ the estimate 
\begin{align}\wh Z_{\ell+1}&=
\ol r\wh Z_\ell,&\ol r:&= \pa{\rc{n}\sumo jn e^{(-\be_{\ell+1} + \be_\ell)f_i(x_j)}}
\end{align} also satisfies~\eqref{eq:Z-ratio-correct-gen-perturb}.
\end{lem}

The way we prove this theorem is to prove the tolerance of each of the proof ingredients to perturbations to $f$.

\subsubsection{Discretization} 

We now verify all the discretization lemmas continue to hold with perturbations. 

The proof of Lemma~\ref{l:reachcontinuous}, combined with the fact that $\ve{\nabla \tilde{f} - \nabla f}_{\infty} \leq \Delta$ gives

\begin{lem}[Perturbed reach of continuous chain] Let $P^{\beta}_T(X)$ be the Markov kernel corresponding to evolving Langevin diffusion 
\begin{equation*}dX_t = - \beta \nabla \tilde{f}(X_t)\,dt + \mathop{d B_t}\end{equation*} 
with $f$ and $D$ are as defined in \ref{eq:A0} for time $T$. Then, 
\begin{equation*}\E[\|X_t - x^*\|^2] \lesssim \E[\|X_0 - x^*\|^2] + \left(400 \beta \frac{D^2 \maxeig^2 \delsmall^2}{\mineig}  + d\right)T \end{equation*} 
\end{lem} 
\begin{proof} 
The proof proceeds exactly the same as Lemma~\ref{l:reachcontinuous}.
\end{proof} 

Furthermore, since $\|\nabla^2 \tilde{f}(x) - \nabla^2 f(x)\|_2 \leq \delsmall, \forall x \in \mathbb{R}^d$, from Lemma~\ref{l:hessianbound}, we get

\begin{lem}[Perturbed Hessian bound] 
$$\|\nabla^2 \tilde{f}(x)\|_2 \leq 4 (D \maxeig)^2 + \delsmall , \forall x \in \mathbb{R}^d$$ 
\end{lem}

As a consequence, the analogue of Lemma~\ref{l:intervaldrift} gives: 
\begin{lem}[Bounding interval drift] In the setting of Lemma~\ref{l:intervaldrift}, let $x \in \mathbb{R}^d, i \in [L]$, and let $\eta \leq \frac{(\frac{1}{\sigma} + \delsmall)^2}{\alpha}$. Then,
$$\mbox{KL}(P_T(x, i) || \widehat{P_T}(x,i)) \leq \frac{4 D^6 \eta^7 (\maxeig + \delsmall)^7}{3} \left(\|x - x^*\|_2^2 + 8 Td\right) + d T D^2 \eta (\maxeig + \delsmall)^2$$
\end{lem}    

Putting these together, we get the analogue of Lemma~\ref{l:maindiscretize}: 
\begin{lem} 
Fix times $0<T_1<\cdots <T_n\le T$. 

Let $p^T, q^T: [L]\times  \mathbb{R}^d \to \mathbb{R}$ be defined as follows.
\begin{enumerate}
\item
$p^T$ is the continuous simulated tempering Markov process as in 
Definition~\ref{df:cst} but with fixed transition times $T_1,\ldots, T_n$. The component chains are Langevin diffusions on $p_i\propto \pa{\sumo jm w_j e^{-f_0(x-\mu_i)}}^{\be_i}$.
\item
 $q^T$ is the discretized version as in Algorithm~\eqref{a:stlmc}, again with fixed transition times $T_1,\ldots, T_n$, and with step size $\eta\le \fc{\si^2}2$.
\end{enumerate}
Then

\begin{align*} \mbox{KL} (P^T || Q^T) &\lesssim \eta^2 D^6 (\maxeig + \delsmall)^6 \left(D^2 \frac{(\maxeig + \delsmall)^2}{\mineig}+d\right) T n\\
&\quad  + \eta^2 D^3 (\maxeig + \delsmall)^3 \max_i \E_{x \sim P^0( \cdot, i)}\|x - x^*\|_2^2 + \eta D^2 (\maxeig + \delsmall)^2 d T   \end{align*}

where $x^*$ is the maximum of $\sumo jm w_j e^{-f_0(x-\mu_j)}$ and satisfies $\ve{x^*}=O(D)$ where $D=\max\ve{\mu_j}$.
\label{l:maindiscretize-perturb}
\end{lem}

\subsubsection{Putting it all together} 

Finally, we prove Lemma~\ref{lem:a1-correct-perturb}.

\begin{proof}[Proof of Lemma \ref{lem:a1-correct-perturb}] 

The proof is analogous to the one of Lemma~\ref{lem:a1-correct} in combination with the Lemmas from the previous subsections, so we just point out the differences.

We bound $\chi^2(\tilde{P}||Q^0)$ as follows: by the proof of Lemma~\ref{lem:a1-correct}, we have $\chi^2(P||Q^0) = O\pa{\fc{\ell}{w_{\min}} \maxeig^{\fc d2}} $. 
By the definition of $\chi^2$, this means 
\begin{align*} \int \frac{q^0(x)^2}{p(x)} dx \leq O\pa{\fc{\ell}{w_{\min}} \maxeig^{\fc d2}} \end{align*}
This in turn implies that 
\begin{align*} \chi^2(\tilde{P}||Q^0) \leq \int \frac{(q^0(x))^2}{\tilde{p}(x)} dx \leq O\pa{\fc{\ell}{w_{\min}} \maxeig^{\fc d2} e^{\Delta}} \end{align*}

Then, analogously as in Lemma~\ref{lem:a1-correct}, we get 
\begin{align}
\ve{p^T(\cdot|A) - q^T(\cdot|A)}_1&=
O\Bigg(
\eta^2 D^6 (\maxeig+\delsmall)^7 \pa{D^2 \fc{(\maxeig+\delsmall)^2}{\mineig}+d}T\eta \\
&\quad\quad+ \eta^2 D^5 \pa{ \frac{\maxeig}{\mineig}+1} + \eta D^2(\maxeig+\delsmall)^2dT
\Bigg).
\end{align}
Choosing $\eta$ as in the statement of the lemma, $\ve{p-q^T}_1\le \ep$ follows.  
The rest of the lemma is identical to Lemma \ref{lem:a1-correct}. 

\end{proof}

\section{More on Markov process decomposition theorems}
\label{a:decomp}
We will prove a general density decomposition theorem for Markov processes, Theorem~\ref{t:gen-decomp}. Then we show how to specialize Theorem~\ref{t:gen-decomp} to the case of simulated tempering, 
Theorem~\ref{thm:gap-prod-st}, We also give a version of the theorem for a continuous index set, Theorem~\ref{t:decomp-cts}.

\subsection{General density decomposition theorem}

\ift{\nomenclature[2Tij]{$T_{i,j}$}{Transition between different components, in the general density decomposition theorem (Chapter~\ref{ch:mm})}}
\ift{\nomenclature[2Scil]{$S_\cil$}{Subset of $I\times I$ chosen to contain pairs $(i,j)$ such that $P_i,P_j$ are close. (Chapter~\ref{ch:mm})}}
\ift{\nomenclature[2Slra]{$S_\lra$}{Subset of $I\times I$ chosen to contain pairs $(i,j)$ such that there is a lot of probability flow between $P_i$ and $P_j$.  (Chapter~\ref{ch:mm})}}
\ift{\nomenclature[2Escriptcil]{$\sE_\cil$}{$\sE_{\cil}(g,g) = \sum_{i\in I} \sE_i(g,g)$ (Chapter~\ref{ch:mm})}}
\ift{\nomenclature[2Escriptlra]{$\sE_\lra$}{$\sE_{\lra}(g,g)= -\sum_{i,j\in I} w_i \an{g,(\sT_{i,j}-\Id)g}_{P_i}$.  (Chapter~\ref{ch:mm})}}
\ift{\nomenclature[2Qij]{$Q_{i,j}$}{Stationary distribution of $P_i$, times transition kernel: $P_i(dx)T_{i,j}(x,dy)$ (Chapter~\ref{ch:mm})}}
\ift{\nomenclature[2Qtildeij]{$\wt Q_{i,j}$}{Normalized version of $Q_{i,j}$ (Chapter~\ref{ch:mm})}}
\ift{\nomenclature[2Ptildeij]{$\wt P_{i,j}$}{$\wt P_{i,j}(dx) = \int_{y\in \Om_j} \wt Q_{i,j}(dx,dy)$ (Chapter~\ref{ch:mm})}}
\begin{thm}[General density decomposition theorem]\label{t:gen-decomp}
Consider a Markov process $M=(\Om,\sL)$ with stationary distribution $p$. Let $M_i=(\Om_i,\sL_i)$ be Markov processes for $i\in I$ ($|I|$ finite), with $M_i$ supported on $\Om_i\subeq \Om$ (possibly overlapping) and with stationary distribution $P_i$. Let the Dirichlet forms be $\sE(g,h) = -\an{g,\sL h}_P$ and $\sE_i(g,h) = -\an{g,\sL_i h}_{P_i}$. (This only depends on $g|_{\Om_i}$ and $h|_{\Om_i}$.)

Suppose the following hold.
\begin{enumerate}
\item
There is a decomposition
\begin{align}
\an{f,\sL g}_P &= {\sum_{i\in I} w_i\an{f,\sL_i g}_{P_i}} + {\sum_{i,j\in I} w_i \an{f, (\sT_{i,j}-\Id)g}_{P_i}}\\
P &= \sum_{i\in I} w_{i}P_{i}.\label{eq:decomp-general}
\end{align}
where $\sT_{i,j}$ acts on a function $\Om_j\to \R$ (or $\Om\to \R$, by restriction) to give a function $\Om_i\to \R$, by $\sT_{i,j}g(x)=\int_{\Om_j} g(y) T_{i,j}(x,dy)$, and $T_{i,j}$ satisfies the following:\footnote{
We are breaking up the generator into a part that describes flow \emph{within} the components, and flow \emph{between} the components. $P_{i,j}$ describes the probability flow between $\Om_i$ and $\Om_j$. Note that fixing the $M_i=(\Om_i,\sL_i)$, the decomposition into the $\sT_{i,j}$ may be non-unique, because the $\Om_i$'s can overlap.}
\begin{itemize}
\item
For every $x\in \Om_i$, $T_{i,j}(x,\cdot)$ is a measure on $\Om_j$.
\item
$w_i P_i(dx) T_{i,j}(x,dy) = w_jP_j(dx)T_{j,i}(y,dx)$.
\end{itemize}•
\item
(Mixing for each $M_{i}$) $M_{i}$ satisfies the Poincar\'e inequality
\begin{align}
\Var_{P_{i}}(g) &\le C \sE_{i}(g,g).
\end{align}
\item
(Mixing for projected chain) Let $I\times I=S_{\cil}\sqcup S_{\lra}$ be a partition of $I\times I$ such that $(i,j)\in S_\bullet \iff (j,i)\in S_{\bullet}$, for $\bullet\in \{\cil, \lra\}$.\footnote{The idea will be to choose $S_{\cil}$ to contain pairs $(i,j)$ such that the distributions $P_i$, $P_j$ are close, and to choose $S_{\lra}$ to contain pairs such that there is a lot of probability flow between $P_i$ and $P_j$.} Suppose that $\ol T:I\times I\to \R_{\ge 0}$ satisfies\footnote{$\ol T$ represents probability flow, and will be chosen to have large value on $(i,j)$ such that $P_i$, $P_j$ are close (from $S_{\cil}$), and on $(i,j)$ such that $P_i$, $P_j$ have a large probability flow between them, as determined by $\wt P_{i,j}$.}
\begin{align}
\label{e:proj-chain1}
\forall i\in I,&& \sum_{j:(i,j)\in S_{\cil}} \ol T(i,j) \chi^2(P_j||P_i) &\le K_1\\
\label{e:proj-chain2}
\forall i\in I,&& \sum_{j:(i,j)\in S_{\lra}} \ol T(i,j) \chi^2(\wt P_{i,j}||P_i) &\le K_2\\
\label{e:proj-chain3}
\forall (i,j)\in S_{\lra},&& \ol T(i,j) &\le K_3 Q_{i,j}(\Om_i,\Om_j)\\
\label{e:proj-chain4}
\forall i,j\in I,&&w_i\ol T(i,j) &= w_j \ol T(j,i)
\end{align}
where $Q_{i,j}(dx,dy) =P_i(dx)T_{i,j}(x,dy)$ and $\wt P_{i,j}(dx) =\fc{P_i(dx)T_{i,j}(x,\Om_j)}{Q_{i,j}(\Om_i,\Om_j)}$. 

Define the projected chain $\ol M$ as the Markov chain on $I$ generated by $\ol{\sL}$, where $\ol{\sL}$ acts on $\ol g\in L^2(I)$ by
\begin{align}
\ol{\sL} \ol g(i) &= 
\sum_{j\in I} (\ol g(j)-\ol g(i)) \ol T(i,j).
\label{eq:chi-trans-p}
\end{align}
Let $\ol P$ be the stationary distribution of $\ol M$ and $\ol{\cE}(\ol g, \ol g) = -\an{\ol g, \ol{\sL}\ol g}$; $\ol M$ satisfies the Poincar\'e inequality
\begin{align}
\Var_{\ol P} (\ol g) & \le \ol C \ol{\sE}(\ol g,\ol g).
\end{align}
\end{enumerate}
Then $M$ satisfies the Poincar\'e inequality
\begin{align}
\Var_P(g) &\le 
 {\max\bc{C\pa{1+\pa{\rc 2 K_1 + 3K_2}\ol C} , 3K_3\ol C}}\sE(g,g).
\end{align}
\end{thm}
To use this theorem, we would choose $\ol T(i,j) \precsim \rc{\chi^2(P_j||P_i)}$ for $(i,j)\in S_{\cil}$ and $\ol T(i,j)\precsim \rc{\chi^2(\wt P_{i,j}||P_i)}$ for $(i,j)\in S_{\lra}$, i.e., choose the projected chain to have large probability flow between distributions that are close, or have a lot of flow between them; $\wt P_{i,j}$ measures how much $P_i$ is ``flowing" into $P_j$.
\begin{rem}
As in Remark~\ref{r:decomp-simple}, we can replace~\eqref{e:proj-chain1} by
\begin{align}
\forall i\in I,&& \sum_{j:(i,j)\in S_{\cil}} \ol T(i,j) \de_{i,j} &\le K_1&\text{where }\de_{i,j}&=\int_{\Om}\min\bc{\dd{P_j}{P_i}, 1}\,P_i(dx)
\end{align}
and obtain
\begin{align}
\Var_P(g) &\le {\max\bc{C\pa{1+\pa{2 K_1 + 3K_2}\ol C} , 3K_3\ol C}}\sE(g,g).
\end{align}•
\end{rem}

\begin{proof}
First, we make some preliminary observations and definitions.
\begin{enumerate}
\item
A stationary distribution $\ol P$ of $\ol M$ is given by $\ol p(i):=\ol P(\{i\}) = w_i$, because $w_i \ol T(i,j) = w_j \ol T(j,i)$ by~\eqref{e:proj-chain4}. 
\item
For $i,j\in I$, define
\begin{align}
Q_{i,j}(dx,dy) :& = P_i(dx)T_{i,j}(x,dy)\\
\wt Q_{i,j}(dx,dy) :& = \fc{Q_{i,j}(dx,dy)}{Q_{i,j}(\Om_i,\Om_j)}\\
\wt P_{i,j}(dx) :& = \int_{y\in \Om_j} \wt Q_{i,j}(dx,dy)
=\fc{P_i(dx)T_{i,j}(x,\Om_j)}{Q_{i,j}(\Om_i,\Om_j)}
\end{align}
and note
\begin{align}
\fc{w_iP_i(dx)T_{i,j}(x,dy)}{w_iQ_{i,j}(\Om_i,\Om_j)} &=
\fc{w_jP_j(dy)T_{j,i}(y,dx)}{w_jQ_{j,i}(\Om_j,\Om_i)}\\
\implies 
\wt Q_{i,j}(dx,dy) &= \wt Q_{j,i}(dy,dx).
\label{e:p-switch}
\end{align}•
\item
Let
\begin{align}
\sE_{\cil}(g,g) &= \sum_{i\in I}w_i\cE_i(g,g) =-\sum_{i\in I} w_i \an{g,\sL_ig}_{P_i}\\
\sE_{\lra}(g,g) &= -\sum_{i,j\in I} w_i \an{g, (\sT_{i,j}-\Id) g}_{P_i}.
\end{align}
\item
We can write $\sE_{\lra}$ in terms of squares as follows.
\begin{align}
\sE_{\lra}(g,g) &=-\sum_{i,j\in I}\int_{\Om_i} g(x) w_i \int_{\Om_j} (g(y)-g(x)) T_{i,j}(x,dy) \,P_i(dx)\\
&=\sum_{i,j\in I}\Bigg[
\rc 2 \int_{x\in \Om_i}\int_{y\in \Om_j}
g(x)^2w_i P_i(dx) T_{i,j}(x,dy)\\
&\quad - \int_{\Om_i}\int_{\Om_j} g(x)g(y) w_i P_i(dx)T_{i,j}(x,dy)
+ \rc 2 \int_{\Om_i}\int_{\Om_j}
g(y)^2w_i P_i(dx) T_{i,j}(x,dy)
\Bigg]\\
&=\rc 2 \sum_{i,j\in I} \int_{\Om_i}\int_{\Om_j}(g(x)-g(y))^2 w_i P_i(dx) T_{i,j}(x,dy)\\
&=\rc 2 \sum_{i,j\in I} \int_{\Om_i}\int_{\Om_j}(g(x)-g(y))^2 w_i Q_{i,j}(dx,dy).
\label{e:e-lra}
\end{align}•
\end{enumerate}•
Given $g\in L^2(\Om)$, define $\ol g\in L^2(I)$ by $\ol g(i) = \EE_{P_{i}} g$. We decompose the variance into the variance within and between the parts, and then use the Poincar\'e inequality on each part.
\begin{align}\label{e:var-decomp0}
\Var_P(g) &=\sum_{i\in I} w_i \int_{\Om_i} (g(x)-\EE_P[g(x)])^2 \,P_i(dx)\\
&=\sum_{i\in I} w_i
\ba{
\pa{\int (g(x)-\EE_{P_{i}}[g(x)])^2 \, P_{i}(dx)}
+(\EE_{P_{i}} g - \EE_P g)^2}\\
&\le C \sum_{i\in I} w_i \sE_{i}(g,g) + \Var_{\ol P}(\ol g)\\
&\le C \sE_{\cil}(g,g) + \ol C \ol \sE (\ol g, \ol g).
\label{eq:to-cont2}
\end{align}
Now we break up $\ol{\sE}$ as follows, 
\begin{align}\label{e:bE-AB}
\ol \sE(\ol g, \ol g) &=\rc2\ub{ \sum_{(i,j)\in S_{\cil}} (\ol g(i)-\ol g(j))^2
w_i\ol T(i,j)}{A}  + 
\rc 2\ub{\sum_{(i,j)\in S_{\lra}}
 (\ol g(i)-\ol g(j))^2
w_i \ol T(i,j)}{B}.
\end{align}
First, as in Theorem~\ref{thm:gap-prod}, we bound $A$ by Lemma~\ref{lem:change-dist},
\begin{align}
A
&\le\sum_{(i,j)\in S_{\cil}} \Var_{P_i}(g) \chi^2(P_j||P_i) w_i  \ol T(i,j)\\
&\le K_1\sum_{(i,j)\in S_{\cil}} w_i\Var_{P_i}(g) \\
&\le K_1C\sE_{\cil}(g,g).
\end{align}
For the second term, 
\begin{align} 
B
&=
\sum_{(i,j)\in S_{\lra}}\ba{
	\int_{x\in \Om_i}\int_{y\in \Om_j} (g(x)-g(y))P_i(dx)P_j(dy)}w_i\ol T(i,j)
	\\
&=
\sum_{(i,j)\in S_{\lra}}\Bigg[
	\int_{x\in \Om_i}\int_{y\in \Om_j} (g(x)-g(y))(P_i(dx)P_j(dy) - \wt Q_{i,j}(dx,dy))
	\\
	&\quad +\int_{x\in \Om_i}\int_{y\in \Om_j} (g(x)-g(y)) \wt Q_{i,j}(dx,dy) 
\Bigg]^2
w_i \ol T(i,j)
\end{align}
Note that
\begin{align}
\int_{\Om_i}\int_{\Om_j} g(x) (P_i(dx)P_j(dy) - \wt Q_{i,j}(dx,dy))
&= \int_{\Om_i} g(x) \pa{P_i(dx) - \int_{\Om_j} \wt Q_{i,j}(dx,dy)}\\
&=\int_{\Om_i}g(x) \pa{P_i(dx) - \wt P_{i,j}(dx)}
\end{align}
and similarly, because $\wt Q_{i,j}(dx,dy)=\wt Q_{i,j}(dy,dx)$ by~\eqref{e:p-switch}, 
\begin{align}
\int_{\Om_i}\int_{\Om_j} g(y) (P_i(dx)P_j(dy) - \wt Q_{i,j}(dx,dy))
&= \int_{\Om_j}g(y) \pa{P_j(dy) - \wt P_{j,i}(dy)}.
\end{align}
Hence
\begin{align}
B&=
\sum_{(i,j)\in S_{\lra}}\Bigg[
	\int_{x\in \Om_i} g(x) \pa{P_i(dx) - \wt P_{i,j}(dx)}
	-
	\int_{x\in \Om_j} g(y) \pa{P_j(dy) - \wt P_{j,i}(dy)}
	\\&\quad
	+
	\int_{x\in \Om_i}\int_{y\in \Om_j} (g(x)-g(y)) \wt Q_{i,j}(dx,dy) 
\Bigg]^2
w_i\ol T(i,j)
\\
&\le\label{e:B-cs1}
3\sum_{(i,j)\in S_{\lra}}\Bigg[
	\pa{\int_{x\in \Om_i} g(x) \pa{P_i(dx) - \wt P_{i,j}(dx)}}^2
	+
	\pa{\int_{x\in \Om_j} g(y) \pa{P_j(dy) - \wt P_{j,i}(dy)}}^2
	\\&\quad
	+
	\pa{\int_{x\in \Om_i}\int_{y\in \Om_j} (g(x)-g(y)) \wt Q_{i,j}(dx,dy)}^2 
\Bigg]
w_i \ol T(i,j) \\
&\quad \text{by Cauchy-Schwarz}\\ 
&\le 3 
\sum_{(i,j)\in S_{\lra}}\Bigg[
	\Var_{P_i} (g) \chi^2(\wt P_{i,j}||P_i) w_i \ol T(i,j)
	+
	\Var_{P_i} (g) \chi^2(\wt P_{j,i}||P_j) w_j  \ol T(j,i)
	\\&\quad
	+
	\int_{x\in \Om_i}\int_{y\in \Om_j} (g(x)-g(y))^2 w_i \wt Q_{i,j}(dx,dy)  \ol T(i,j)
\Bigg]
\\&\quad\text{by Lemma~\ref{lem:change-dist} and~\eqref{e:proj-chain4} }(w_i\ol T(i,j) = w_j\ol T(j,i))
\\
&\le 3\Bigg[
K_2\sum_{i\in I} w_i \Var_{P_i}(g)
+K_2\sum_{j\in I} w_j \Var_{P_j}(g)\\
&\quad \text{by~\eqref{e:proj-chain2} and~\eqref{e:proj-chain3}}\\
&\quad 
+K_3\sum_{(i,j)\in S_{\lra}}  \int_{x\in \Om_i}\int_{y\in \Om_j}(g(x)-g(y))^2 w_i Q_{i,j} (dx,dy)\Bigg]\\
&\le 6 K_2C\sE_{\cil}(g,g) + 6 K_3\sE_{\lra}(g,g)
\end{align}
where the last line follows from the Poincar\'e inequality on each $M_i$, and~\eqref{e:e-lra}. 
Then
\begin{align}
\Var_P(g) \le \eqref{eq:to-cont2}
&\le C\sE_{\cil}(g,g) + \ol C \pa{\rc 2(A+B)}\\
&\le C\sE_{\cil}(g,g) + \ol C \rc 2(K_1C \sE_{\lra}(g,g) + 6K_2C\sE_{\cil}(g,g) + 6K_3\sE_{\lra}(g,g))\\
&\le {\max\bc{C\pa{1+\pa{\rc 2 K_1 + 3K_2}\ol C} , 3K_3\ol C}}\sE(g,g).
\end{align}•
\end{proof}

\subsection{Theorem for simulated tempering}

\begin{proof}[Proof of Theorem~\ref{thm:gap-prod-st} from~\ref{t:gen-decomp}]
We first relate $M$ to $M'$ on $[L]\times \Om$ defined as follows. $M'$ has transition probability from level $i$ to level $i'=i\pm 1$ given by $\sumo jm r_iw_{i,j}\min\bc{\fc{r_{i'}w_{i',j}p_{(i',j)}(x)}{r_{i}w_{i,j}p_{(i,j)}(x)}, 1}$ rather than $r_i \min \bc{\fc{r_{i'}p_{i'}(x)}{r_ip_i(x)},1}$. We show that $\sE'(g,g)\le \sE(g,g)$ below; this basically follows from the fact that the probability flow between any two distinct points in $M'$ is at most the probability flow in $M$. More precisely (letting $p\sL$ denote the functional defined by $(p\sL f)(x) = p(x) (\sL f)(x)$),
\begin{align}\label{e:L'}
&p\sL' = \sumo iL r_i p_i \sL_i + \fc{\la}2 \sumr{1\le i,i'\le L}{i'=i\pm 1} \sumo jm r_iw_{i,j} \min\bc{\fc{r_{i'}w_{i',j}p_{(i',j)}(x)}{r_{i}w_{i,j}p_{(i,j)}(x)},1}(g_{i'}-g_i)(x)\\
&\sE'(g,g) = -\an{g,\sL'g}_P \\
&= \sumo iL r_i \sE_i(g_i,g_i)\\
&\quad  + \fc{\la}2 \sumr{1\le i,i'\le L}{i'=i\pm 1} \int_{\Om} \sumo jm r_iw_{i,j}p_{(i,j)}(x) \min\bc{\fc{r_{i'}w_{i',j}p_{(i',j)}(x)}{r_{i}w_{i,j}p_{(i,j)}(x)},1}(g_{i}(x)^2-g_i(x)g_{i'}(x))\dx\\
&= \sumo iL r_i \sE_i(g_i,g_i) + \fc{\la}4 \sumr{1\le i,i'\le L}{i'=i\pm 1} \int_{\Om} \sumo jm r_iw_{i,j}p_{(i,j)}(x) \min\bc{\fc{r_{i'}w_{i',j}p_{(i',j)}(x)}{r_{i}w_{i,j}p_{(i,j)}(x)},1}(g_i(x)-g_{i'}(x))^2\dx\\
&=\sumo iL r_i \sE_i(g_i,g_i) + \fc{\la}4 \sumr{1\le i,i'\le L}{i'=i\pm 1} \int_{\Om} \sumo jm \min\bc{{r_{i'}w_{i',j}p_{(i',j)}(x)},{r_{i}w_{i,j}p_{(i,j)}(x)}}(g_i(x)-g_{i'}(x))^2\\
&\le \sumo iL r_i \sE_i(g_i,g_i) + \fc{\la}4 \sumr{1\le i,i'\le L}{i'=i\pm 1} \int_{\Om}\min\bc{{r_{i'}p_{i'}(x)},{r_{i}p_{i}(x)}}(g_i(x)-g_{i'}(x))^2=\sE(g,g).
\end{align}
Thus it suffices to prove a Poincar\'e inequality for $M'$.
We will apply Theorem~\ref{t:gen-decomp} with 
\begin{align}
T_{(i,j),(i',j')}((i,x), (i',dy)) &= 
\begin{cases}
\fc\la2\min\bc{\fc{r_{i'}w_{i',j}p_{(i',j)}(x)}{r_{i}w_{i,j}p_{(i,j)}(x)},1}\de_x(dy),& j=j',i'=i\pm 1\\
0,& \text{else}.
\end{cases}
\label{e:st-decomp-1}
\end{align}
First we calculate 
$\wt P_{(i,j),(i',j)}$. 
We have
\begin{align}\label{e:qijij}
Q_{(i,j),(i',j)}([i]\times \Om,[i']\times \Om) &= \fc{\la}2\int_\Om \min\{r_iw_{i,j}p_{(i,j)}(x), r_{i'}w_{i',j}p_{(i',j)}(x)\}\dx = \fc{\la}2\de_{(i,j),(i',j)}.
\end{align}
so
\begin{align}
\wt p_{(i,j),(i',j)}(x) &= \fc{p_{(i,j)}(x)T_{(i,j),(i',j)}(x,\Om_j)}{Q_{(i,j),(i',j)}([i]\times \Om,[i']\times \Om)}\\
&=\fc{p_{(i,j)}(x)\fc{\la}2 \min\bc{\fc{r_{i'}w_{i',j}p_{(i',j)}(x)}{r_{i}w_{i,j}p_{(i,j)}(x)},1}}{\fc{\la}2\de_{(i,j),(i',j)}} = \rc{\de_{(i,j),(i',j)}}\min\bc{\fc{r_{i'}w_{i',j}p_{(i',j)}(x)}{r_{i}w_{i,j}},p_{(i,j)}(x)}.
\label{e:pijij}
\end{align}•
We check the 3 assumptions in Theorem~\ref{t:gen-decomp}.
\begin{enumerate}
\item
From Assumption 1,~\eqref{e:L'} and~\eqref{e:st-decomp-1},
\begin{align}
p\sL' &= \sumo iL \sumo jm r_i w_{i,j} p_{(i,j)}\sL_{i,j} + \sumr{1\le i,i'\le L}{i'=i\pm 1}\sumo jm r_iw_{i,j}p_{(i,j)}(\sT_{(i,j),(i',j)}-\Id).
\end{align}
\item
This follows immediately from Assumption 2.
\item
Let $S_{\cil}$ consist of all pairs $((1,j),(1,j'))$ and $S_{\lra}$ consist of all pairs $((i,j),(i'=i\pm 1,j))$. (The other pairs $((i,j),(i',j'))$ satisfy $\ol T((i,j),(i',j'))=0$, so they do not matter.)
We check equations~\eqref{e:proj-chain1}--\eqref{e:proj-chain4}.
\begin{enumerate}
\item[\eqref{e:proj-chain1}] By~\eqref{e:st-proj},
\begin{align}
\sumo{j,j'}m \ol T((1,j),(1,j')) \chi^2(P_{(1,j')}||P_{(1,j)}) & = 
\sumo{j,j'}m \fc{w_{1,j'}}{\chi^2_{\max}(P_{(1,j)}||P_{(1,j')})} \chi^2(P_{(1,j')}||P_{(1,j)}) \le 1,
\end{align}
so \eqref{e:proj-chain1} is satisfied with $K_1=1$.
\item[\eqref{e:proj-chain2}] 
We apply Lemma~\ref{l:overlap-chi} 
with $P=P_{(i,j)}$ and $Q=\fc{r_{i'}w_{i',j}p_{(i',j)}}{r_{i}w_{i,j}}$. Noting that $\wt P_{(i,j),(i',j)} = \rc{\de_{i,j,i',j}}Q$ by~\eqref{e:pijij}, we obtain 
\begin{align}\chi^2\pa{\wt P_{(i,j),(i',j)}||P_{(i,j)}}\le \rc{\de_{(i,j),(i',j)}}
\label{e:de-chi}
\end{align}
By~\eqref{e:st-proj} and~\eqref{e:de-chi},
\begin{align}
\sum_{i'=i\pm 1} \ol T((i,j),(i',j)) \chi^2(\wt P_{(i,j),(i',j)}||P_{(i,j)}) 
&\le \sum_{i'=i\pm 1} {K\de_{(i,j),(i',j)}} \chi^2(\wt P_{(i,j),(i',j)}||P_{(i,j)}) \le 2K.
\end{align}
\item[\eqref{e:proj-chain3}] By~\eqref{e:st-proj} and~\eqref{e:qijij},
\begin{align}
\ol T((i,j),(i',j')) &= K\de_{(i,j),(i',j)}
\le \fc{2K}{\la} \cdot \fc{\la}2\de_{(i,j),(i',j)} = \fc{2K}{\la} Q_{i,j,i',j}([i]\times \Om,[j]\times \Om)
\end{align}
so \eqref{e:proj-chain3} is satisfied with $K_3=\fc{2K}{\la}$.
\item[\eqref{e:proj-chain4}] We have that
\begin{align}
&&r_1w_{1,j}\ol T((1,j),(1,j')) & = \fc{r_1w_{1,j}w_{1,j'}}{\chi^2_{\max}(P_{(1,j)}||P_{(1,j')})}
=r_1w_{1,j'}\ol T((1,j'),(1,j))\\
\text{for }i'=i\pm 1,&&
r_iw_{i,j}\ol T((i,j),(i',j)) & =
Kr_iw_{i,j} \de_{(i,j),(i',j)} \\
&&&= K r_{i'}w_{i',j} \de_{(i',j),(i,j)}
=r_{i'}w_{i',j}\ol T((i',j),(i,j)).
\end{align}
\end{enumerate}
\end{enumerate}
Hence the conclusion of Theorem~\ref{t:gen-decomp} holds with $K_1=1$, $K_2=2K$, and $K_3=\fc{2 K}{\la}$.
\end{proof}

\subsection{Continuous version of decomposition theorem}

We consider the case where $I$ is continuous, and the Markov process is the Langevin process. We will take $I=\Om^{(1)}\subeq \R^{d_1}$ and each $\Om_i$ will be a fixed $\Om^{(2)}\subeq \R^{d_2}$, so the space is $\Om^{(1)}\times \Om^{(2)}\subeq \R^{d_1}\times \R^{d_2}=\R^{d_1+d_2}$. 

The proof of the following is very similar to the proof of Theorem~\ref{t:gen-decomp}, and to Lemma 5 in~\cite{mou2019polynomial}. The main difference is that they bound using an $L^\iy$ norm over $\Om^{(1)}\times \Om^{(2)}$, and we bound using a $L^\iy$ norm over just $\Om^{(1)}$; thus this bound is stronger, and can prevent having to consider restrictions. We also note that there is an analogue of the theorem for log-Sobolev inequalities~\cite{lelievre2009general,grunewald2009two}.
\begin{thm}[Poincar\'e inequality from marginal and conditional distribution]\label{t:decomp-cts}
Consider a probability measure $\pi$ with $C^1$ density on $\Om=\Om^{(1)}\times \Om^{(2)}$, where $\Om^{(1)}\subeq \R^{d_1}$ and $\Om^{(2)}\subeq \R^{d_2}$ are closed sets. For $X=(X_1,X_2)\sim P$ with probability density function $p$ (i.e., $P(dx) = p(x)\,dx$ and $P(dx_2|x_1) = p(x_2|x_1)\,dx_2$), suppose that 
\begin{itemize}
\item
The marginal distribution of $X_1$ satisfies a Poincar\'e inequality with constant $C_1$.
\item
For any $x_1\in \Om^{(1)}$, the conditional distribution $X_2|X_1=x_1$ satisfies a Poincar\'e inequality with constant $C_2$.
\end{itemize}
Then $\pi$ satisfies a Poincar\'e inequality with constant
\begin{align}
\wt C&= 
\max\bc{
C_2\pa{1+2C_1 \ve{\int_{\Om^{(2)}} \fc{\ve{\nb_{x_1}p(x_2|x_1)}^2}{p(x_2|x_1)}\dx_2}_{L^\iy(\Om^{(1)})}}, 2C_1
}
\end{align}
\end{thm}
(Note that an alternate way to write $\int_{\Om^{(2)}} \fc{\ve{\nb_{x_1}p(x_2|x_1)}^2}{p(x_2|x_1)}\dx_2$ is $\int_{\Om^{(2)}} \nb_{x_1}p(x_2|x_1) \nb_{x_1}(\ln p(x_2|x_1))\dx_2$.)
Adapting this theorem it would be possible, with additional work, to show that Langevin on the space $[\be_0,1]\times \Om$ (that is, include the temperature as a coordinate in the Langevin algorithm), where the first coordinate is the temperature, will mix.

In the proof we will draw analogies between the discrete and continuous case.
\begin{proof}
To make the analogy, let 
\begin{align}
\sE_{\lra}(g,g) = \sE^{(1)}(g,g) :&= \int_{\Om}\ve{\nb_{x_1} g(x)}^2\,P(dx)\\
\sE_{\cil}(g,g) = \sE^{(2)}(g,g) :&= \int_{\Om}\ve{\nb_{x_2} g(x)}^2\,P(dx)
\end{align}•
and note $\sE=\sE_{\lra}+\sE_\cil = \sE^{(1)}+\sE^{(2)}$.

Let $\ol P$ be the $x_1$-marginal of $P$, i.e., $\ol P(A) = P(A\times \Om^{(2)})$.  Given $g\in L^2(\Om^{(1)}\times \Om^{(2)})$, define $\ol g\in L^2(\Om^{(1)})$ by $\ol g(x) = \E_{x_2\sim P(\cdot |x_1)}[g(x)]$. 
Analogously to~\eqref{e:var-decomp0}, 
\begin{align}
\Var_P(g) &= \int_{\Om^{(1)}}\pa{\int_{\Om^{(2)}} (g(x)-\EE_{P}g(x))^2\,P(dx_2|x_1)}\ol P(dx_1)\\
&=\int_{\Om^{(1)}}\pa{\int_{\Om^{(2)}} (g(x)-\EE_{x_2\sim P(\cdot|x_1)}g(x))^2\,P(dx_2|x_1)
+ \pa{\EE_{x_2\sim P(\cdot |x_1)} [g(x)] - \EE_P[g(x)]}^2}\ol P(dx_1)\\
&\le \int_{\Om^{(1)}} C\sE_{P(\cdot|x_1)}(g,g) \ol P(dx_1) + \Var_{\ol P}(\ol g)\\
&\le C_2\sE^{(2)}(g,g) + C_1 \ol \sE (\ol g, \ol g).
\label{e:var-decomp-cts}
\end{align}
The second term is analogous to the $B$ term in~\eqref{e:bE-AB}. (There is no $A$ term.) Note $\ol \sE(\ol g, \ol g) = \int_{\Om^{(1)}}\ve{\nb_{x_1}\ol g(x_1)}^2\,\ol P(dx_1)$, and we can expand $\nb_{x_1}\ol g(x_1)$ using integration by parts:
\begin{align}
\nb_{x_1}\ol g(x_1) = \nb_{x_1}\pa{\int_{\Om^{(2)}} g(x)\,P(dx_2|x_1)}
&=\int_{\Om^{(2)}} \nb_{x_1}g(x) P(dx_2|x_1) + \int_{\Om^{(2)}} g(x) \nb_{x_1}p(x_2|x_1)\dx_2
\end{align}
Hence by Cauchy-Schwarz, (compare with~\eqref{e:B-cs1})
\begin{align}
\label{e:ol-decomp-cts}
\ol{\sE}(\ol g,\ol g)
&\le 2\ba{
\int_{\Om} \ve{\nb_{x_1}g(x)}^2 P(dx_2|x_1)\ol P(dx_1) +\int_{\Om^{(1)}}\ve{\int_{\Om^{(2)}} g(x) \nb_{x_1}p(x_2|x_1)\dx_2}^2\ol P(dx_1)
}
\end{align}
The first term is $\sE^{(1)}(g,g)$. The second term is bounded by Lemma~\ref{lem:change-dist-cts}, the continuous analogue of Lemma~\ref{lem:change-dist}, with $g\mapsfrom g(x_1,\cdot)$ and $p_{x_1}(x_2)=p(x_2|x_1)$.
\begin{align}
&\quad \int_{\Om^{(1)}}\ve{\int_{\Om^{(2)}} g(x) \nb_{x_1}p(x_2|x_1)\dx_2}^2\ol P(dx_1)\\
&\le\int_{\Om^{(1)}}\Var_{P(\cdot|x_1)}[g(x)]\pa{ \int_{\Om^{(2)}} \fc{\ve{\nb_{x_1}p(x_2|x_1)}^2}{p(x_2|x_1)}\dx_2}\ol P(dx_1)\\
&\le \int_{\Om^{(1)}} C_2\sE_{P(\cdot|x_1)}(g,g)\ol P(dx_1)\cdot \ve{ \int_{\Om^{(2)}} \fc{\ve{\nb_{x_1}p(x_2|x_1)}^2}{p(x_2|x_1)}\dx_2}_{L^\iy(\Om^{(1)})}\\
&\quad \text{by }L^1\text{-}L^\iy\text{ inequality}\\
&=C_2\sE^{(2)}(g,g) \ve{ \int_{\Om^{(2)}} \fc{\ve{\nb_{x_1}p(x_2|x_1)}^2}{p(x_2|x_1)}\dx_2}_{L^\iy(\Om^{(1)})}
\end{align}
Hence, recalling~\eqref{e:var-decomp-cts} and~\eqref{e:ol-decomp-cts},
\begin{align}
\ol \sE(\ol g,\ol g) &\le 2\sE^{(1)}(g,g) + 2\ba{C_2\sE^{(2)}(g,g) \ve{ \int_{\Om^{(2)}} \fc{\ve{\nb_{x_1}p(x_2|x_1)}^2}{p(x_2|x_1)}\dx_2}_{L^\iy(\Om^{(1)})}}\\
\Var_P(g) &\le C_2\sE^{(2)}(g,g) + C_1\sE(\ol g,\ol g)\\
&\le \max\bc{
C_2\pa{1+2C_1 \ve{\int_{\Om^{(2)}} \fc{\ve{\nb_{x_1}p(x_2|x_1)}^2}{p(x_2|x_1)}\dx_2}_{L^\iy(\Om^{(1)})}}, 2C_1
}\sE(g,g).
\end{align}•
\end{proof}
\section{Examples}
\label{sec:examples}
It might be surprising that sampling a mixture of gaussians require a complicated Markov Chain such as simulated tempering. However, many simple strategies seem to fail. 

\paragraph{Langevin with few restarts} One natural strategy to try is simply to run Langevin a polynomial number of times from randomly chosen locations. While the time to ``escape'' a mode and enter a different one could be exponential, we may hope that each of the different runs ``explores'' the individual modes, and we somehow stitch the runs together. The difficulty with this is that when the means of the gaussians are not well-separated, it's difficult to quantify how far each of the individual runs will reach and thus how to combine the various runs.  

\paragraph{Recovering the means of the gaussians} Another natural strategy would be to try to recover the means of the gaussians in the mixture by performing gradient descent on the log-pdf with a polynomial number of random restarts. The hope would be that maybe the local minima of the log-pdf correspond to the means of the gaussians, and with enough restarts, we should be able to find them. 

Unfortunately, this strategy without substantial modifications also seems to not work: for instance, in dimension $d$, consider a mixture of $d+1$ gaussians, $d$ of them with means on the corners of a $d$-dimensional simplex with a side-length substantially smaller than the diameter $D$ we are considering, and one in the center of the simplex. In order to discover the mean of the gaussian in the center, we would have to have a starting point extremely close to the center of the simplex, which in high dimensions seems difficult.

Additionally, this doesn't address at all the issue of robustness to perturbations. Though there are algorithms to optimize ``approximately'' convex functions, they can typically handle only very small perturbations. \cite{belloni2015escaping, risteski2016algorithms}
   

\paragraph{Gaussians with different covariance} Our result requires all the gaussians to have the same variance. This is necessary, as even if the variance of the gaussians only differ by a factor of 2, there are examples where a simulated tempering chain takes exponential time to converge \cite{woodard2009sufficient}. Intuitively, this is illustrated in Figure~\ref{figure:variance}. The figure on the left shows the distribution in low temperature \--- in this case the two modes are separate, and both have a significant mass. The figure on the right shows the distribution in high temperature. Note that although in this case the two modes are connected, the volume of the mode with smaller variance is much smaller (exponentially small in $d$). Therefore in high dimensions, even though the modes can be connected at high temperature, the probability mass associated with a small variance mode is too small to allow fast mixing.

In the next section, we show that even if we do not restrict to the particular simulated tempering chain, no efficient algorithm can efficiently and robustly sample from a mixture of two Gaussians with different covariances.

\begin{figure}[h!]
\centering
\includegraphics[height=2in]{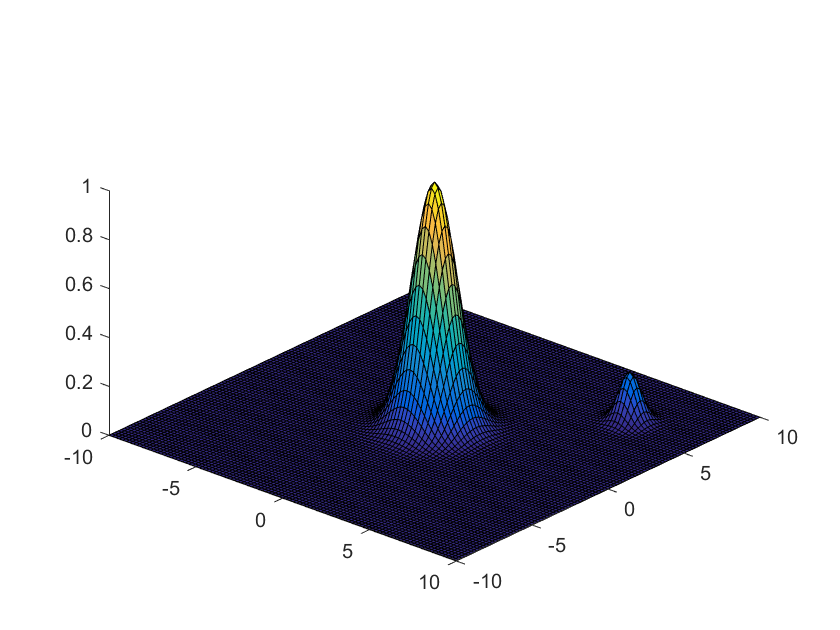}
\includegraphics[height=2in]{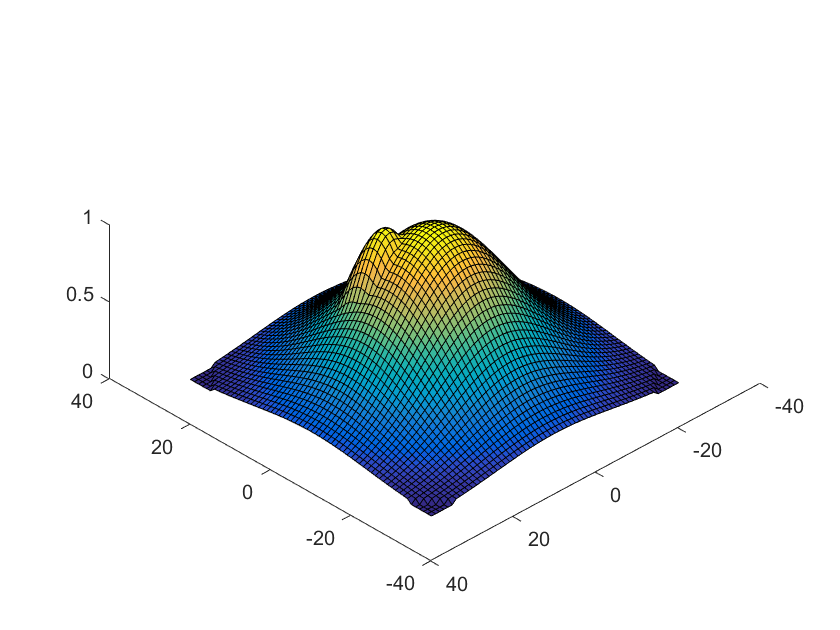}
\caption{Mixture of two gaussians with different covariance at different temperature}
\label{figure:variance}
\end{figure}

\section{Lower bound when Gaussians have different variance}
\label{sec:lb}
In this section, we give a lower bound showing that in high dimensions, if the Gaussians can have different covariance matrices, results similar to our Theorem~\ref{thm:main} cannot hold. In particular, we construct a log density function $\tilde{f}$ that is close to the log density of mixture of two Gaussians (with different variances), and show that any algorithm must query the function at exponentially many locations in order to sample from the distribution. More precisely, we prove the following theorem:

\begin{thm}\label{thm:lowerbound}
There exists a function $\tilde{f}$ such that $\tilde{f}$ is close to a negative log density function $f$ for a mixture of two Gaussians: $\ve{f-\tilde{f}}_\infty \le \log 2$, $\forall x$ $\|\nabla f(x)-\nabla \tilde{f}(x)\|\le O(d)$, $\|\nabla^2 f(x) - \nabla^2 \tilde{f}(x)\| \le O(d)$. Let $\tilde{p}$ be the distribution whose density function is proportional to $\exp(-\tilde{f})$. There exists constant $c > 0, C>0$, such that when $d \ge C$, any algorithm with at most $2^{cd}$ queries to $\tilde{f}$ and $\nabla \tilde{f}$ cannot generate a distribution that is within TV-distance $0.3$ to $\tilde{p}$.
\end{thm}

In order to prove this theorem, we will first specify the mixture of two Gaussians. Consider a uniform mixture of two Gaussian distributions $N(0, 2I)$ and $N(u, I) (u\in \R^d)$ in  $\R^d$.

\begin{df}\label{def:lowerboundf}
Let $f_1 = \|x\|^2/4 + \frac{d}{2}\log(2\sqrt{2}\pi)$ and $f_2 = \|x-u\|^2/2 + \frac{d}{2}\log(2\pi)$. The mixture $f$ used in the lower bound is
$$
f = - \log\pa{\frac{1}{2} (e^{-f_1} + e^{-f_1})}.
$$
\end{df}


In order to prove the lower bound, we will show that there is a function $\tilde{f}$ close to $f$, such that $\tilde{f}$ behaves exactly like a single Gaussian $N(0,2I)$ on almost all points. Intuitively, any algorithm with only queries to $\tilde{f}$ will not be able to distinguish it with a single Gaussian, and therefore will not be able to find the second component $N(u, I)$. More precisely, we have

\begin{lem}\label{lem:outdominate}
When $\|u\| \ge 4d\log 2$, for any point $x$ outside of the ball with center $2u$ and radius $1.5\|u\|$, we have $e^{-f_1(x)} \ge e^{-f_2(x)}$. 
\end{lem}

\begin{proof}
The Lemma follows from simple calculation. In order for $e^{-f_1(x)} \ge e^{-f_2(x)}$, since $e^x$ is monotone we know
$$
-\frac{\|x-u\|^2}{2} \le -\frac{\|x\|^2}{4} - \frac{d}{4}\log 2.
$$
This is a quadratic inequality in terms of $x$, reordering the terms we get
$$
\|x-2u\|^2 \ge d\log 2 + 2\|u\|^2.
$$
Since $d\log 2 \le 0.25\|u\|^2$, we know whenever $\|x-2u\|^2 \ge 1.5\|u\|$ this is always satisfied, and hence $e^{-f_1(x)} \ge e^{-f_2(x)}$.
\end{proof}

The lemma shows that outside of this ball, the contribution from the first Gaussian is dominating. Intuitively, we try to make $\tilde{f} = f_1$ outside of this ball, and $\tilde{f} = f$ inside the ball. To make the function continuous, we shift between the two functions gradually. More precisely, we define $\tilde{f}$ as follows:

\begin{df}\label{def:tildef}
The function 
\begin{equation}
\tilde{f}(x) = g(x) f_1(x) + (1-g(x)) f(x).
\end{equation}
Here the function $g(x)$  (see Definition~\ref{def:funcg}) satisfies
$$
g(x)= \left\{\begin{array}{cc}1 & \|x-2u\|\ge 1.6\|u\| \\ 0 & \|x-2u\| \le 1.5\|u\| \\  \in[0,1] &\mbox{otherwise}\end{array}\right.
$$
Also $g(x)$ is twice differentiable with all first and second order derivatives bounded.
\end{df}

With a carefully constructed $g(x)$, it is possible to prove that $\tilde{f}$ is point-wise close to $f$ in function value, gradient and Hessian, as stated in the Lemma below. Since these are just routine calculations, we leave the construction of $g(x)$ and verification of this lemma at the end of this section.

\begin{lem}\label{lem:funcclose}
For the functions $f$ and $\tilde{f}$ defined in Definitions~\ref{def:lowerboundf} and \ref{def:tildef}, if $\|u\| \ge 4d\log 2$, there exists a large enough constant $C$ such that
\begin{align*}
\ve{f - \tilde{f}}_\infty &\le \log 2 \\
\forall x \quad \|\nabla f(x) - \nabla \tilde{f}(x)\| & \le C\|u\|\\
\forall x \quad \|\nabla^2 f(x) - \nabla^2 \tilde{f}(x)\| & \le C\|u\|^2.
\end{align*}
\end{lem}

Now we are ready to prove the main theorem:

\begin{proof}[Proof of Theorem~\ref{thm:lowerbound}]
We will show that if we pick $\|u\|$ to be a uniform random vector with norm $8d\log 2$, there exists constant $c > 0$ such that for any algorithm, with probability at least $1 - \exp(-cd)$, in the first $\exp{cd}$ iterations of the algorithm there will be no vector $x \ne 0$ such that $\cos \theta(x,u) \ge 3/5$. 

First, by standard concentration inequalities, we know for any fixed vector $x\ne 0$ and a uniformly random $u$,
$$
\Pr[\cos \theta(x,u) \ge 3/5] \le \exp(-c'd),
$$
for some constant $c' > 0$ ($c' = 0.01$ suffices). 

Now, for any algorithm, consider running the algorithm with oracle to $f_1$ and $\tilde{f}$ respectively (if the algorithm is randomized, we also couple the random choices of the algorithm in these two runs). Suppose when the oracle is $f_1$ the queries are $x_1,x_2,...,x_t$ and when the oracle is $\tilde{f}$ the queries are $\tilde{x}_1,...,\tilde{x}_t$. 

Let $c = c'/2$, when $t \le \exp(cd)$, by union bound we know with probability at least $1-\exp(-cd)$, we have $\cos \theta(x_i,u) < 3/5$ for all $i \le t$. On the other hand, every point $y$ in the ball with center $2\|u\|$ and radius $1.6\|u\|$ has $\cos \theta(y,u) \ge 3/5$. We know $\|x_i - 2u\| > 1.6\|u\|$, hence $f_1(x_i) = \tilde{f}(x_i)$ for all $i\le t$ (the derivatives are also the same). Therefore, the algorithm is going to get the same response no matter whether it has access to $f_1$ or $\tilde{f}$. This implies $\tilde{x}_i = x_i$ for all $i\le t$.

Now, to see why this implies the output distribution of the last point is far from $\tilde{p}$, note that when $d$ is large enough $\tilde{p}$ has mass at least $0.4$ in ball $\|x_i - 2u\| \le 1.6\|u\|$ (because essentially all the mass in the second Gaussian is inside this ball), while the algorithm has less than $0.1$ probability of having any point in this region. Therefore the TV distance is at least $0.3$ and this finishes the proof.
\end{proof}

\subsection{Construction of $g$ and closeness of two functions}

Now we finish the details of the proof by construction a function $g$.

\begin{df} \label{def:funcg}
Let $h(x)$ be the following function:
$$
h(x) =\left\{\begin{array}{cc}1 & x\ge 1 \\ 0 & x\le 0 \\ x^2(1-x)^2 + (1-(1-x)^2)^2 & x\in[0,1]\end{array}\right.
$$
We then define $g(x)$ to be $g(x):= h\pa{10\pa{\frac{\|x-2u\|}{\|u\|} - 1.5}}$. 
\end{df}

For this function we can prove:

\begin{lem}\label{lem:boundg}
The function $g$ defined above satisfies
$$
g(x)= \left\{\begin{array}{cc}1 & \|x-2u\|\ge 1.6\|u\| \\ 0 & \|x-2u\| \le 1.5\|u\| \\  \in[0,1] &\mbox{otherwise}\end{array}\right.
$$
Also $g(x)$ is twice differentiable. There exists large enough constant $C_g>0$ such that for all $x$
$$
\|\nabla g(x)\| \le C_g\|u\| \quad \|\nabla^2 g(x)\|\le C_g(\|u\|^2+1).
$$
\end{lem}

\begin{proof}
First we prove properties of $h(x)$. Let $h_0(x) = x^2(1-x)^2 + (1-(1-x)^2)^2$, it is easy to check that $h_0(0) = h'_0(0) = h''_0(0) = 0$, $h_0(1) = 1$ and $h'_0(1) = h''_0(1) = 0$. Therefore the entire function $h(x)$ is twice differentiable. 

Also, we know $h'_0(x) = 2x(4x^2 - 9x +5)$, which is always positive when $x \in [0,1]$. Therefore $h(x)$ is monotone in $[0,1]$. The second derivative $h''_0(x) = 24x^2 - 36x + 10$. Just using the naive bound (sum of absolute values of individual terms) we can get for any $x\in [0,1]$, $|h'_0(x)|\le 36$ and $|h''(x)| \le 60$. (We can of course compute better bounds but it is not important for this proof.)

Now consider the function $g$. We know when $\|x-2u\| \in [1.5,1.6]\|u\|$, 
$$
\nabla g(x) = h'\pa{10\pa{\frac{\|x-2u\|}{\|u\|} - 1.5}}\cdot 10(x-2u).
$$

Therefore $\|\nabla g(x)\| \le 36\times 10\times \|x-2u\| \le C_g\|u\|$ (when $C_g$ is a large enough constant).

For the second order derivative, we know
$$
\nabla^2 g(x) = 100 h''\pa{10\pa{\frac{\|x-2u\|}{\|u\|} - 1.5}}(x-2u)(x-2u)^\top + 10h'\pa{10\pa{\frac{\|x-2u\|}{\|u\|} - 1.5}} I.
$$

Again by bounds on $h'$ and $h''$ we know there exists large enough constants so that $\|\nabla^2 g(x)\| \le C_g(\|u\|^2 + 1)$. 

\end{proof}

Finally we can prove Lemma~\ref{lem:funcclose}.

\begin{proof}[Proof of Lemma~\ref{lem:funcclose}]
We first show that the function values are close. When $\|x-2u\| \le 1.5 \|u\|$, by definition $\tilde{f}(x) = f(x)$. When $\|x-2u\| \ge 1.5\|u\|$, by property of $g$ we know $\tilde{f}(x)$ is between $f(x)$ and $f_1(x)$. Now by Lemma~\ref{lem:outdominate}, in this range $e^{-f_1(x)} \ge e^{-f_2(x)}$, so $f_1(x)-\log 2 \le f(x) \le f_1(x)$. As a result we know $|\tilde{f}(x) - f(x)| \le \log 2$.

Next we consider the gradient. Again when $\|x-2u\| \le 1.5\|u\|$ the two functions (and all their derivatives) are the same. When $\|x-2u\| \in [1.5,1.6]\|u\|$, we have
$$
\nabla \tilde{f}(x) = g(x) \nabla f_1(x) + (1-g(x))\nabla f(x) + (f_1(x)-f(x))\nabla g(x).
$$
By Lemma~\ref{lem:boundg} we have upperbounds for $g(x)$ and $\|\nabla g(x)\|$, also both $\|\nabla f_1(x)\|,\|\nabla f(x)\|$ can be easily bounded by $O(1)\|u\|$, therefore $\|\nabla \tilde{f}(x) - \nabla f(x)\| \le C\|u\|$ for large enough constant $C$.

When $\|x-2u\| \ge 1.6\|u\|$, we know $\nabla \tilde{f}(x) = \nabla f_1(x)$. Calculation shows
$$
\nabla f_1(x) - \nabla f(x) = \frac{e^{-f_2(x)}}{e^{-f_1(x)}+e^{-f_2(x)}} (\nabla f_1(x) - \nabla f_2(x)).
$$

When $\|x\| \le 50\|u\|$, we have $\|\nabla f_1(x)-\nabla f_2(x)\| \le 2\|x\|+2\|u\| \le O(1)\|u\|$. When $\|x\| \ge 50\|u\|$, it is easy to check that $\frac{e^{-f_2(x)}}{e^{-f_1(x)}+e^{-f_2(x)}} \le \exp{-\|x\|^2/5}$ and $\|\nabla f_1(x) - \nabla f_2(x)\| \le 2\|x\|$, therefore in this case the difference in gradient bounded by $\exp(-t^2/5)2t$ which is always small. 

Finally we can check the Hessian. Once again when $\|x-2u\| \le 1.5\|u\|$ the two functions are the same. When $\|x-2u\| \in [1.5,1.6]\|u\|$, we have
\begin{align*}
\nabla^2 \tilde{f}(x)  = &g(x)\nabla^2 f_1(x) + (1-g(x)) \nabla^2 f(x)\\ & + (\nabla f_1(x) - \nabla f(x))(\nabla g(x))^\top + (\nabla g(x))(\nabla f_1(x) - \nabla f(x))^\top\\ & + (f_1-f) \nabla^2 g(x).
\end{align*}
In this case we get bounds for $g(x),\nabla g(x), \nabla^2 g(x)$ from Lemma~\ref{lem:boundg}, $\|\nabla f_1(x)\|,\|\nabla f(x)\|$ can still be bounded by $O(1)\|u\|$, $\|\nabla^2 f(x)\|,\|\nabla^2 f_1(x)\|$ can be bounded by $O(\|u\|^2)$ and $O(1)$ respectively. Therefore we know $\|\nabla^2 \tilde{f}(x) - \nabla^2 f(x)\| \le C\|u\|^2$ for large enough constant $C$.

When $\|x-2u\| \ge 1.6\|u\|$, we have $\tilde f(x) = f_1(x)$, and
\begin{align*}
\nabla^2 f_1(x) - \nabla^2 f(x) = &\frac{e^{-f_2(x)}}{e^{-f_1(x)}+e^{-f_2(x)}} (\nabla^2 f_1(x) - \nabla^2 f_2(x))\\ & + \frac{e^{-f_1(x)-f_2(x)}(\nabla f_1(x) - \nabla f_2(x))(\nabla f_1(x) - \nabla f_2(x))^\top}{(e^{-f_1(x)}+e^{-f_2(x)})^2}.
\end{align*}

Here the first term is always bounded by a constant (because $e^{-f_2(x)}$ is smaller, and $\nabla^2 f_1(x) - \nabla^2 f_2(x) = I/2$). For the second term, 
by arguments similar as before, we know when $\|x\| \le 50\|u\|$ this is bounded by $O(1)\|u\|^2$. When $\|x\| \ge 50\|u\|$ we can check $\frac{e^{-f_1(x)-f_2(x)}}{(e^{-f_1(x)}+e^{-f_2(x)})^2} \le \exp(-\|x\|^2/5)$ and $\|(\nabla f_1(x) - \nabla f_2(x))(\nabla f_1(x) - \nabla f_2(x))^\top\| \le 4\|x\|^2$. Therefore the second term is bounded by $\exp(-t^2/5)\cdot 4t^2$ which is no larger than a constant. Combining all the cases we know there exists a large enough constant $C$ such that $\|\nabla^2 f(x) - \nabla^2 f(x)\|\le C\|u\|^2$ for all $x$.

\end{proof}

\iftoggle{thesis}{}{\section{Calculations on probability distributions}}
\label{sec:calc}
\subsection{Chi-squared and KL inequalities}
\begin{lem}\label{lem:change-dist}
Let $P,Q$ be probability measures
on $\Om$ such that $Q\ll P$, $\chi^2(Q||P)<\iy$, and $g:\Om\to \R$ satisfies $g\in L^2(P)$. Then 
\begin{align}
\pa{
\int_\Om g(x)\,P(dx) - \int_\Om g(x)\,Q(dx)
}^2 \le \Var_P(g) \chi^2(Q||P).
\end{align}
\end{lem}
\begin{proof}
Noting that $\int_{\Om} P(dx)-Q(dx)=0$ and using Cauchy-Schwarz,
\begin{align}
\pa{
\int_\Om g(x)\,dP(x) - \int_\Om g(x)\,Q(dx)
}^2
& = \pa{\int_\Om (g(x)-\EE_P[g(x)]) (P(dx)-Q(dx))}^2\\
&\le \pa{\int_\Om (g(x)-\EE_P[g(x)])^2 P(dx)} \pa{\int_\Om \pa{1-\dd QP}^2P(dx)}\\
&=\Var_P (g) \chi^2(Q||P).
\end{align}
\end{proof}

The continuous analogue of Lemma~\ref{lem:change-dist} is the following.
\begin{lem}\label{lem:change-dist-cts}
Let $\Om = \Om^{(1)}\times \Om^{(2)}$ with $\Om^{(1)}\subeq \R^{d_1}$. 
Suppose $P_{x_1}$ is a probability measure on $\Om^{(2)}$ for each $x_1\in \Om^{(1)}$ with density function $p_{x_1}$ (with respect to some reference measure $dx$), 
$g:\Om^{(2)}$ satisfies $g\in L^2(P_{x_1})$, and $\int_{\Om^{(2)}} \fc{\ve{\nb_{x_1}p_{x_1}(x_2)}^2}{p_{x_1}(x_2)}\dx_2<\iy$. 
Then
\begin{align}
\ve{\int_{\Om^{(2)}}
g(x_2)\nb_{x_1}(p_{x_1}(x_2))\dx_2}^2
&\le 
\Var_{P_{x_1}}(g) \pa{ \int_{\Om^{(2)}} \fc{\ve{\nb_{x_1}p_{x_1}(x_2)}^2}{p_{x_1}(x_2)}\dx_2}
\end{align}
\end{lem}
\begin{proof}
Because each $P_{x_1}$ is a probability measure, 
\begin{align}
\int_{\Om^{(2)}}\nb_{x_1} p_{x_1}(x_2) \dx_2
&= \nb_{x_1}\int_{\Om^{(2)}}p_{x_1}(x_2)\dx_2=\nb_{x_1}(1)=0.
\end{align}
Hence
\begin{align}
& \ve{\int_{\Om^{(2)}} 
g(x_2) \nb_{x_1}p_{x_1}(x_2)\dx_2}^2\\
&\le 
\ve{\int_{\Om^{(2)}} \ba{g(x_2)-\EE_{P_{x_1}}[g(x_2)]} \nb_{x_1}p_{x_1}(x_2)\dx_2}^2\\
&\le \pa{\int_{\Om^{(2)}} \ba{g(x_2)-\EE_{P_{x_1}}[g(x_2)]}^2
p_{x_1}(x_2)\dx}
\pa{\int_{\Om^{(2)}} \fc{\ve{\nb_{x_1}p_{x_1}(x_2)}^2}{p_{x_1}(x_2)}\dx_2}\\
&=\Var_{P_{x_1}}(g)\pa{ \int_{\Om^{(2)}} \fc{\ve{\nb_{x_1}p_{x_1}(x_2)}^2}{p_{x_1}(x_2)}\dx_2}
\end{align}•
\end{proof}
%

\begin{lem}\label{l:overlap-chi}
Let $P$ be a probability measure and $Q$ be a nonnegative measure on $\Om$.
Define the measure $R$ by $R=\min\bc{\dd{Q}{P}, 1}P$. (If $p,q$ are the density functions of $P,Q$, then the density function of $R$ is simply $r(x)=\min\{p(x),q(x)\}$.) Let $\de$ be the overlap $\de=R(\Om)$, and $\wt R=\fc{R}{\de}=\fc{R}{R(\Om)}$ the normalized overlap measure.
Then
\begin{align}
\chi^2\pa{R||P} &\le \rc{\de}.
\end{align}
\end{lem}
\begin{proof}
We make a change of variable to $u = \dd QP$. Let $F(u) = P\pa{\set{x}{\dd QP(x)\le u}}$. Then
\begin{align}
\chi^2\pa{R||P} 
&= \int_{\Om} \pa{\fc{\min\bc{1,\dd QP}}{\de}-1}^2\,P(dx)\\
&=\int_{\Om}\pa{\fc{\min\{1,u\}}{\de}-1}^2 \,dF(u) &\text{(Stieltjes integral)}\\
&\le\rc{\de^2} \int_{\Om}(\min\{1,u\})^2 \,dF(u)\\
&\le \rc{\de^2} \int_{\Om}\min\{1,u\} \,dF(u).
\label{e:clm-to-bd}
\end{align}
Now note that $\int_{\Om}\min\{1,u\} \,dF(u) = \int_{\Om}\min\bc{1,\dd QP}\,P(dx)= \de$. Hence~\eqref{e:clm-to-bd} is at most $\rc{\de^2}\de=\rc\de$.
\end{proof}

\begin{lem}\label{lem:chi-sq-mixture} 
If $P, P_i$ are probability measures on $\Om$ such that $P=\sumo in w_iP_i$ (where $w_i>0$ sum to 1), and $Q\ll P,P_i$, then
\begin{align}
\chi^2(Q||P) &\le \sumo in w_i \chi^2(Q||P_i).
\end{align}
\end{lem}
This inequality follows from convexity of $f$-divergences; for completeness we include a proof.
\begin{proof}
By Cauchy-Schwarz,
\begin{align}
\chi^2(Q||P)&=
\pa{\int_\Om\sumo in  \pa{\dd QP}^2 \,P(dx)}-1\\
&=\int_\Om \pa{ \sumo in w_i \dd{P_i}{P} \dd{Q}{P_i}}^2\,P(dx)-1\\
&\le \int_\Om \pa{\sumo in w_i \dd{P_i}{P}}\pa{\sumo in w_i \pa{\dd Q{P_i}}^2\dd{P_i}P}\,P(dx)-1\\
& \le\sumo in w_i\pa{ \int_{\Om}\pa{\dd Q{P_i}}^2 - 1}\,P_i(dx) = \sumo in w_i \chi^2(Q||P_i)
\end{align}
\end{proof}

\begin{lem}\label{lem:chi-liy}
Suppose $P, \wt P$ are probability distributions on $\Om$ such that $\dd{\wt P}{P}\le K$. Then for any probability measure $Q\ll P$,
\begin{align}
\chi^2(Q||P) &\le K \chi^2(Q||\wt P) + K-1.
\end{align}
\end{lem}
\begin{proof}
\begin{align}
\chi^2(Q||P) &= \int_\Om \pa{\dd QP}^2P(dx)-1 
=\int_\Om \pa{\dd Q{\wt P}}^2 \pa{\dd{\wt P}{P}}^2 P(dx)-1\\
&\le K\pa{\int_\Om\pa{\dd{Q}{\wt P}}^2 \wt P(dx) }-1
= K(\chi^2(Q||\wt P)+1)-1.
\end{align}
\end{proof}

\begin{lem} \label{l:kl-mixture}\label{l:decomposingKL}
Let $W$ and $W'$ be probability measures over $I$, with densities $w(i)$, $w'(i)$ with respect to a reference measure $di$, such that $\mbox{KL}(W||W')<\iy$. 
For each $i\in I$, suppose $P_i, Q_i$ are probability measures over $\Om$. Then
$$ \mbox{KL}\left(\int_{I} w(i) P_i \,di || \int_{I} w'(i) Q_i\,di\right) \leq \mbox{KL}(W || W') + \int_{I} w(i) \mbox{KL}(P_i || Q_i) \,di.$$ 
\end{lem} 

\begin{proof}  
Overloading notation, we will use $KL (a || b) $ for two measures $a,b$ even if they are not necessarily probability distributions, with the obvious definition. Using the convexity of KL divergence,
\begin{align*} 
\mbox{KL}\left(\int_{I} w(i) P_i \,di|| \int_{I} w'(i) Q_i\,di\right) &= \mbox{KL}\left(\int_{I} w(i) P_i\,di || \int_{I} w(i) Q_i \frac{w'(i)}{w(i)}\,di\right) \\ 
&\leq \int_{I} w(i)  \mbox{KL}\left( P_i || Q_i \frac{w'(i)}{w(i)}\right)\,di \\ 
&=\int_{I} w(i) \log\left(\frac{w(i)}{w'(i)}\right)\,di + \int_i w(i)\mbox{KL}(P_i || Q_i) \,di\\ 
&=\mbox{KL}(W || W') + \int_{I} w(i) \mbox{KL}(P_i || Q_i) \,di\end{align*} 
\end{proof}

\subsection{Chi-squared divergence calculations for log-concave distributions}

We calculate the chi-squared divergence between log-concave distributions at different temperatures, and at different locations. In the gaussian case there is a closed formula (Lemma~\ref{lem:chi-squared-N}). The general case is more involved (Lemmas~\ref{lem:chisq-translate} and~\ref{lem:chisq-temp}), and the bound is in terms of the strong convexity and smoothness constants.

\begin{lem}\label{lem:chi-squared-N}
For a matrix $\Si$, let $|\Si|$ denote its determinant. 
The $\chi^2$ divergence between $N(\mu_1,\Si_1)$ and $N(\mu_2,\Si_2)$ is
\begin{align}
&\chi^2(N(\mu_2,\Si_2)||N(\mu_1,\Si_1))\\
&= 
\fc{|\Si_1|^{\rc 2}}{|\Si_2|}
\ab{( 2\Si_2^{-1}-\Si_1^{-1})}^{-\rc 2}\\
& \cdot \exp\pa{
\rc 2(2\Si_2^{-1}\mu_2-\Si_1^{-1}\mu_1)^T (2\Si_2^{-1}-\Si_1^{-1})^{-1}
(2\Si_2^{-1}\mu_2-\Si_1^{-1}\mu_1)
+\rc2\mu_1^T\Si_1^{-1} \mu_1 -\mu_2^T\Si_2^{-1}\mu_2
}-1
\end{align}
In particular, in the cases of equal mean or equal variance,
\begin{align}
\chi^2(N(\mu,\Si_2)||N(\mu,\Si_1))
&=\fc{|\Si_1|^{\rc 2}}{|\Si_2|}
\ab{(2\Si_2^{-1}-\Si_1^{-1})}^{-\rc2}-1\\
\chi^2(N(\mu_2,\Si)||N(\mu_1,\Si))
&=\exp[(\mu_2-\mu_1)^T \Si^{-1} (\mu_2-\mu_1)].
\end{align}
\end{lem}
\begin{proof}
\begin{align}
&
\chi^2(N(\mu,\Si_2)||N(\mu,\Si_1))+1\\
&= \rc{(2\pi)^{\fc d2}} 
\fc{|\Si_1|^{\rc 2}}{|\Si_2|}
\int_{\R^d}  \exp\ba{
-\rc2\pa{2(x-\mu_2)^T\Si_2^{-1}(x-\mu_2) - (x-\mu_1)^T\Si_1^{-1} (x-\mu_1)}
}\dx\\
&=\rc{(2\pi)^{\fc d2}} 
\fc{|\Si_1|^{\rc 2}}{|\Si_2|}\int_{\R^d} \exp\Bigg[
-\rc2 \Bigg(x^T(2\Si_2^{-1} - \Si_1^{-1})x + 2x^T\Si_1^{-1}\mu_1
- 4x^T\Si_2^{-1}x \\
&\quad
- \mu_1^T \Si_1^{-1} \mu_1 + 2\mu_2^T\Si_2^{-1}\mu_2
\Bigg)\Bigg]\dx\\
&=
\rc{(2\pi)^{\fc d2}} 
\fc{|\Si_1|^{\rc 2}}{|\Si_2|}
\int_{\R^d} \exp\ba{-\rc 2({x'}^T(2\Si_2^{-1}-\Si_1^{-1}) x' + c)}\\
x':&= x - (2\Si_2^{-1}-\Si_1^{-1})^{-1}
(\mu_1^T\Si_1^{-1} - 2\mu_2^T \Si_2^{-1})\\
c:&=\rc 2(2\Si_2^{-1}\mu_2-\Si_1^{-1}\mu_1)^T (2\Si_2^{-1}-\Si_1^{-1})^{-1}
(2\Si_2^{-1}\mu_2-\Si_1^{-1}\mu_1)
+\rc2\mu_1^T\Si_1^{-1} \mu_1 -\mu_2^T\Si_2^{-1}\mu_2
\end{align}
Integrating gives the result. For the equal variance case,
\begin{align}
c&=\rc2(2\mu_2-\mu_1)\Si^{-1}(2\mu_2-\mu_1) + \rc 2 \mu_1\Si^{-1}\mu_1 - \mu_2\Si^{-1}\mu_2 = (\mu_2-\mu_1)^T \Si^{-1}(\mu_2-\mu_1)^T.
\end{align}
\end{proof}


The following theorem is essential in generalizing from gaussian to log-concave densities.

\begin{thm}[{Harg\'e, \cite{harge2004convex}}]\label{thm:harge}
Suppose the $d$-dimensional gaussian $N(0,\Si)$ has density $\ga$. Let $p=h\cdot \ga$ be a probability density, where $h$ is log-concave. Let $g:\R^d\to \R$ be convex. Then
\begin{align}
\int_{\R^d} g(x-\EE_p x)p(x)\dx &\le \int_{\R^d} g(x)\ga(x)\dx.
\end{align}
\end{thm}

\begin{lem}[$\chi^2$-tail bound]\label{lem:chi-tail-bound}
Let $\ga=N\pa{0,\rc{\ka}}$. Then
\begin{align}
\forall y\ge \sfc{d}{\mineig},\quad
\Pj_{x\sim \ga}\pa{\ve{x} \ge y} & \le e^{-\fc{\mineig}2 \pa{y- \sfc{d}{\mineig}}^2}.
\end{align}•\end{lem}
\begin{proof}
By the $\chi_d^2$ tail bound in \cite{laurent2000adaptive}, for all $t\ge 0$,
\begin{align}
\Pj_{x\sim \ga}\pa{\ve{x}^2 \ge \rc{\mineig} (\sqrt d + \sqrt{2t})^2}&\le 
\Pj_{x\sim \ga}\pa{\ve{x}^2 \ge \rc{\mineig} (d + 2(\sqrt{dt}+t))}\le e^{-t}\\
\implies
\forall y\ge \sfc{d}{\mineig},\quad
\Pj_{x\sim \ga}\pa{\ve{x} \ge y} & \le e^{-\pf{\sqrt{\mineig} y - \sqrt d}{\sqrt 2}^2} = e^{-\fc{\mineig}2 \pa{y- \sfc{d}{\mineig}}^2}
\end{align}

\end{proof}

\begin{lem}\label{lem:mode-mean}
Let $f:\R^d\to \R$ be a $\mineig$-strongly convex and $\maxeig$-smooth function and let $P$ be a probability measure with density function $p(x)\propto e^{-f(x)}$. Let $x^*=\amin_x f(x)$ and $\ol x = \EE_P f(x)$. Then 
\begin{align}
\ve{x^*-\ol x} &\le \sfc{d}{\mineig}\pa{\sqrt{\ln \pf{\maxeig}{\mineig}}+5}.
\end{align}
\end{lem}
\begin{proof}
We establish both concentration around the mode $x^*$ and the mean $\ol x$. This will imply that the mode and the mean are close. Without loss of generality, assume $x^*=0$ and $f(0)=0$.

For the mode, note that by Lemma \ref{lem:chi-tail-bound}, for all $r\ge \fc{d}{m}$, 
\begin{align}\label{eq:mm1}
\int_{\ve{x}\ge r} e^{-f(x)}\dx &\le \int_{\ve{x}\ge r} e^{-\rc 2 \mineig x^2} \le \pf{2\pi}{\mineig}^{\fc d2} 
{e^{-{\fc{\mineig}2 \pa{r-\sfc{d}{\mineig}}^2}}}\\
\label{eq:mm2}
\int_{\ve{x}<r} e^{-f(x)}\dx &\ge \int_{\ve{x}<r} e^{-\rc 2 \maxeig x^2} \ge \pf{2\pi}{\maxeig}^{\fc d2} 
\pa{1-e^{-\fc{\maxeig}2\pa{r-\sfc{d}{\mineig}}^2}}.
\end{align}
Let $r = \sfc{d}{\mineig}\pa{\sqrt{\ln \pf{\maxeig}{\mineig}}+3}$. Then 
\begin{align}
\int_{\ve{x}\ge r} e^{-f(x)}\dx &\le 
\pf{2\pi}{\mineig}^{\fc d2} e^{-\fc{d}2\pa{\ln \pf{\maxeig}{\mineig}+2}}\le
\pf{2\pi}{\maxeig}^{\fc d2} e^{-d}\\
\int_{\ve{x}<r} e^{-f(x)}\dx &\ge \pf{2\pi}{\maxeig}^{\fc d2}\pa{1-e^{-\fc{\maxeig}2\pa{r-\sfc{d}{\mineig}}^2}}\\
&\ge \pf{2\pi}{\maxeig}^{\fc d2} \pa{1-e^{-\fc{\maxeig d}{2\mineig}\pa{2+\ln \pf{\maxeig}{\mineig\mineig}}}}\ge \pf{2\pi}{\maxeig}^{\fc d2}(1-e^{-d})
\end{align}
Thus
\begin{align}
\Pj_{x\sim P}(\ve{x}\ge r) & = \fc{\int_{\ve{x}\ge r} e^{-f(x)}\dx}{\int_{\ve{x}\ge r} e^{-f(x)}\dx+\int_{\ve{x}<r} e^{-f(x)}\dx } \le e^{-d}\le \rc 2.\label{eq:mode-conc}
\end{align}

Now we show concentration around the mean. By adding a constant to $f$, we may assume that $p(x) = e^{-f(x)}$. 
Note that because $f$ is $\mineig$-smooth, $p$ is the product of $\ga(x)$ 
with a log-concave function, where $\ga(x)$ is the density of $N(0,\rc{\mineig} I_d)$. 
 note that by Harg\'e's Theorem~\ref{thm:harge},
\begin{align}
\int_{\R^d} \ve{x-\ol x}^2 p(x) \,dx &\le \int_{\R^d} \ve{x}^2 \ga(x) \dx = \fc{d}{\mineig}.
\end{align}
By Markov's inequality, 
\begin{align}
\Pj_{x\sim P}\pa{\ve{x-\ol x}\ge \sfc{2d}{\mineig}} &= \Pj_{x\sim P}\pa{\ve{x-\ol x}^2 \ge \fc{2d}{\mineig}} \le \rc 2.\label{eq:mean-conc}
\end{align}
Let $B_r(x)$ denote the ball of radius $r$ around $x$. By~\eqref{eq:mode-conc} and~\eqref{eq:mean-conc}, $B_{\sfc{d}{\mineig}\pa{\sqrt{\ln \pf{\maxeig}{\mineig}}+3}}(x^*)$ and $B_{ \sfc{2d}{\mineig}}(\ol x)$ intersect. Thus $\ve{\ol x - x^*}\le \sfc{d}{\mineig}\pa{\sqrt{\ln \pf{\maxeig}{\mineig}}+5}$. 
\end{proof}

\begin{lem}[Concentration around mode for log-concave distributions]
\label{lem:conc-mode}
Suppose $f:\R^d\to \R$ is $\mineig$-strongly convex and $\maxeig$-smooth. Let $P$ be the probability measure with density function $p(x)\propto e^{-f(x)}$. 
Let $x^*=\amin_x f(x)$. Then
\begin{align}
\Pj_{x\sim P}\pa{
	\ve{x-x^*}^2 \ge \rc{\mineig} 
	\pa{\sqrt{d} + \sqrt{2t + d\ln\pf{\maxeig}{\mineig}}
}^2}\le e^{-t}.
\end{align}
\end{lem}
\begin{proof}
By~\eqref{eq:mm1} and~\eqref{eq:mm2},
\begin{align}
\Pj_{x\sim P}(\ve{x}\ge r)
&\le \pf{\maxeig}{\mineig}^{\fc d2} e^{-\fc{\mineig}2\pa{r-\sfc{d}{\mineig}}^2}.
\end{align}
Substituting in $r=\rc{\sqrt{\mineig}} 
	\pa{\sqrt{d} + \sqrt{2t + d\ln\pf{\maxeig}{\mineig}}}$ gives the lemma.
\end{proof}

\begin{lem}[$\chi^2$-divergence between translates]\label{lem:chisq-translate}
Let $f:\R^d\to \R$ be a $\mineig$-strongly convex and $\maxeig$-smooth function and let $p(x)\propto e^{-f(x)}$ be a probability distribution. Let $\ve{\mu}=D$. Then
\begin{align}
\chi^2(p(x)||p(x-\mu)) &\le 
e^{\rc 2 \mineig D^2 + \maxeig D
\sfc{d}{\mineig}\pa{\sqrt{\ln \pf{\maxeig}{\mineig}}+5}
}\pa{
e^{\maxeig D\sfc{d}{\mineig}}
+
\maxeig D \sfc{4\pi}{\mineig} e^{ \fc{2\maxeig D \sqrt d}{\sqrt{\mineig}} + \fc{\maxeig^2 D^2}{2\mineig}}}-1.
\end{align}
\end{lem}
\begin{proof}
Without loss of generality, suppose $f$ attains minimum at 0, or equivalently, $\nb f(0)=0$. 
We bound
\begin{align}
\chi^2(p(x)||p(x-\mu)) +1& = \int_{\R^d} \fc{e^{-2f(x)}}{e^{-f(x-\mu)}} \dx
=\int_{\R^d} e^{-f(x)} e^{f(x-\mu) - f(x)}\dx\\
&\le \int_{\R^d} e^{-f(x)} e^{\maxeig {D} \ve{x} + \rc 2 \mineig {D}^2}\dx\label{eq:chi-lc1}
\end{align}
Note that because $f$ is $\mineig$-strongly convex, $p$ is the product of $\ga(x)$ 
with a log-concave function, where $\ga(x)$ is the density of $N(0,\rc{\mineig} I_d)$. 
Let $\ol x = \EE_{x\sim p} x$ be the average value of $x$ under $p$. Apply Harg\'e's Theorem~\ref{thm:harge} on $g(x) = e^{\maxeig{D}\ve{x+\ol x}}$, $p(x) = e^{-f(x)}$ to get that
\begin{align}
 \int_{\R^d} e^{-f(x)} e^{\maxeig {D} \ve{x}}\dx &\le \int_{\R^d}
 \ga(x) e^{\maxeig{D}\ve{x+\ol x}}\dx\\
 &= e^{\maxeig {D}\ve{\ol x}}
 \pa{
 e^{\maxeig D\sfc{d}{\mineig}}
 + \int_{\sfc{d}{\mineig}}^\iy \Pj_{x\sim \ga} (\ve{x}\ge y) \maxeig {D} e^{\maxeig {D}y}\,dy}\label{eq:chi-lc2}
\end{align}
where we used the identity $\int_{\R} f(x) p(x)\dx  = f(y_0)+ \int_{y_0}^\iy \Pj_{x\sim p}(x\ge y) f'(y)\,dy$ when $f(x)$ is an increasing function.
By Lemma~\ref{lem:chi-tail-bound}, 
\begin{align}
\forall y\ge \sfc{d}{m},\quad
\Pj_{x\sim \ga}\pa{\ve{x} \ge y} & \le 
 e^{-\fc{\mineig}2 \pa{y- \sfc{d}{\mineig}}^2}\\
\implies 
\int_{\sfc{d}{\mineig}}^\iy \Pj_{x\sim \ga} (\ve{x}\ge y) \maxeig{D} e^{\maxeig {D}y}\,dy 
&\le 
\maxeig D \int_{\sfc{d}{\mineig}}^\iy e^{-\fc{\mineig}2 \pa{y - \fc{d}{\mineig} }^2 + \maxeig D y} \,dy\\
&=
\maxeig D \int_{\sfc{d}{\mineig}}^\iy e^{-\fc{\mineig}2\ba{ \pa{y - \fc{d}{\mineig} }^2 - \fc{2\maxeig D \sqrt d}{\mineig^{\fc 32}} - \fc{\maxeig^2 D^2}{\mineig^2}}} \,dy\\
&=
\maxeig D \int_{\sfc{d}{\mineig}}^\iy e^{-\fc{\mineig}2 \pa{y - \fc{d}{\mineig} }^2 + \fc{2\maxeig D \sqrt d}{\sqrt{\mineig}} + \fc{\maxeig^2 D^2}{2\mineig}} \,dy\\
&\le \maxeig D \sfc{4\pi}{\mineig} e^{ \fc{2\maxeig D \sqrt d}{\sqrt{\mineig}} + \fc{\maxeig^2 D^2}{2\mineig}}.\label{eq:chi-lc3}
\end{align}
Putting together~\eqref{eq:chi-lc1},~\eqref{eq:chi-lc2}, and~\eqref{eq:chi-lc3}, and using Lemma~\ref{lem:mode-mean},
\begin{align}
\chi^2(p(x)||p(x-\mu)) &\le 
e^{\rc 2 \mineig D^2 + \maxeig D
\sfc{d}{\mineig}\pa{\sqrt{\ln \pf{\maxeig}{\mineig}}+5}
}\pa{ 
e^{\maxeig D\sfc{d}{\mineig}}
+
\maxeig D \sfc{4\pi}{\mineig} e^{ \fc{2\maxeig D \sqrt d}{\sqrt{\mineig}} + \fc{\maxeig^2 D^2}{2\mineig}}}.
\end{align}
\end{proof}

\begin{lem}[$\chi^2$-divergence between temperatures]\label{lem:chisq-temp}
Let $f:\R^d\to \R$ be a $\mineig$-strongly convex and $\maxeig$-smooth function and let $P,P_\be$ be probability measures with density functions $p(x)\propto e^{-f(x)}$, $p_\be(x)\propto e^{-\be f(x)}$. Suppose $\be_1,\be_2>0$ and $|\be_2-\be_1|<\fc{\mineig}{\maxeig}$.
Then
\begin{align}
\chi^2(P_{\be_2}||P_{\be_1}) &\le 
 e^{\rc 2\ab{1-\fc{\be_1}{\be_2}} \fc{\maxeig d}{\mineig  -\maxeig\ab{1-\fc{\be_1}{\be_2}}}
\pa{\sqrt{\ln \pf{\maxeig}{\mineig}}+5}^2}
\pa{
\pa{1-\fc{\maxeig}{\mineig}\ab{1-\fc{\be_1}{\be_2}}}\pa{1+\ab{1-\fc{\be_{i-1}}{\be_i}}}
}^{-\fc d2}-1.
\end{align}
\end{lem}
\begin{proof}
Without loss of generality, suppose $f$ attains minimum at 0 (or equivalently, $\nb f(0)=0$), and $f(0)=0$. 
We bound
\begin{align}
\chi^2(P_{\be_2}||P_{\be_1})+1 
&= \fc{\int_{\R^d} e^{-\be_1f(x)}\dx \int_{\R^d} e^{(\be_1-2\be_2)f(x)}\dx}{\pa{\int_{\R^d} e^{-\be_2f(x)}\dx}^2}.
\label{eq:chisq-temp1}
\end{align}
Let $\ol x = \EE_{x\sim P_{\be_2}} x$ be the average value of $x$ under $p_{\be_2}$. Note that because $f$ is $m$-strongly convex, $e^{-\be_1f(x)}$ is the product of $\ga(x)$ with a log-concave function, where $\ga(x)$ is the density of $N\pa{0,\rc{\be_2\mineig}I_d}$. Applying Harg\'e's Theorem~\ref{thm:harge} on $g_1(x) = e^{(\be_2-\be_1)f(x+\ol x)}$ and $g_2(x) = e^{(\be_1-\be_2)f(x+\ol x)}$ to get
\begin{align}
\eqref{eq:chisq-temp1} &\le 
\int_{\R^d} e^{(\be_2-\be_1) f(x+\ol x)}\ga (x) \dx
\int_{\R^d} e^{(\be_1-\be_2) f(x+\ol x)}\ga (x) \dx.
\label{eq:chisq-temp2}
\end{align}
Because $f$ is $m$-strongly convex and $M$-smooth, and $f(0)=0$ is the minimum of $f$,
\begin{align}
\eqref{eq:chisq-temp2}&\le 
\fc{\int_{\R^d} e^{-|\be_2-\be_1| \fc{\maxeig}2\ve{x+\ol x}^2} e^{-\fc{\be_2\mineig}2\ve{x}^2}\dx
\int_{\R^d} e^{-|\be_2-\be_1| \fc{\mineig}2\ve{x+\ol x}^2} e^{-\fc{\be_2\mineig}2\ve{x}^2}\dx}
{\pa{\int_{\R^d} e^{-\fc{\be_2m}2 \ve{x}^2}\dx}^2}.
\label{eq:chisq-temp3}
\end{align}
Using the identity
\begin{align}
a\ve{x+\ol x}^2 + b\ve{x}^2 &= (a+b)\ve{x}^2 + 2a\an{x,\ol x} + \fc{a^2}{a+b}\ve{\ol x}^2 + \fc{ab}{a+b}\ve{\ol x}^2 \\
&= (a+b)\ve{x+\fc{a}{a+b}\ol x}^2 + \fc{ab}{a+b}\ve{\ol x}^2,
\end{align}
we get using Lemma~\ref{lem:mode-mean} ($\cdots$ denote quantities not involving $x$, that we will not need)
\begin{align}
\eqref{eq:chisq-temp3}&=
\rc{\pa{\int_{\R^d} e^{-\fc{\be_2 \mineig}2 \ve{x}^2}\dx}^2}
\Bigg[
e^{\fc{\maxeig \mineig |\be_2-\be_1|\be_2}{2(\mineig \be_2-\maxeig|\be_2-\be_1|)}\ve{\ol x}^2}
\int_{\R^d} e^{\pa{\fc{\maxeig}2|\be_2-\be_1|-\fc{\mineig}2\be_2}\ve{x+\cdots}^2}\dx
\\
&\quad \cdot 
e^{\fc{-\mineig|\be_1-\be_2|\be_2}{2\mineig(\be_2-|\be_2-\be_1|)}\ve{\ol x}^2}
\int_{\R^d} e^{\pa{-\fc{\mineig}2\be_2 - |\be_2-\be_1| \fc{\mineig}2}\ve{x+\cdots}^2}\dx
\Bigg]
\\
&\le e^{\fc{|\be_2-\be_1|}2 \fc{\maxeig\mineig\be_2}{\mineig \be_2 - \maxeig |\be_2-\be_1|}\ve{\ol x}^2}
\pf{2\pi}{\mineig \be_2-\maxeig|\be_2-\be_1|}^{\fc d2} \pf{2\pi}{\mineig(\be_2+|\be_2-\be_1|)}^{\fc d2}\pf{2\pi}{\mineig\be_2}^{-d}\\
&\le e^{
	\rc 2\ab{
		1-\fc{\be_1}{\be_2}
	} \fc{\maxeig d}{\mineig  -\maxeig\ab{1-\fc{\be_1}{\be_2}}} 
	\pa{\sqrt{\ln \pf{\maxeig}{\mineig}}+5}^2
}
\pa{
	\pa{1-\fc{\maxeig}{\mineig}\ab{1-\fc{\be_1}{\be_2}}}
	\pa{1+\ab{1-\fc{\be_{i-1}}{\be_i}}}
}^{-\fc d2}.
\end{align}
\end{proof}

\begin{lem}[$\chi^2$ divergence between gaussian and log-concave distribution]\label{lem:chi-warm}
Suppose that probability measure $P$ has probability density function $p(x)\propto e^{-f(x-\mu)}$, where $f$ is $\mineig$-strongly convex, $\maxeig$-smooth, and attains minimum at 0. Let $D=\ve{\mu}$. Then
\begin{align}
\chi^2\pa{N\pa{0,\rc K I_d}||P}&\le \pf{\maxeig}{\mineig}^{\fc d2} e^{\maxeig D^2}.
\end{align}
\end{lem}
\begin{proof}
We calculate 
\begin{align}
p(x) &= \fc{e^{- f_0(x-\mu)}}{\int_{\R^d} e^{- f_0(u-\mu)}\,du}
\ge \fc{e^{-\fc{\maxeig}{2}\ve{x-\mu}^2}}{\int_{\R^d} e^{-\fc{\mineig}2 \ve{u-\mu}^2}\,du} = \pf{\mineig}{2\pi}^{\fc d2} e^{- \fc{\maxeig}2\ve{x-\mu}^2}.
\end{align}
Then
\begin{align}
\chi^2\pa{N\pa{0,\rc{K}I_d}||P}
&=\int_{\R^d} \fc{\pf{\maxeig}{2\pi}^d e^{-\maxeig \ve{x}^2}}{p(x)}\,dx - 1\\
&\le\pf{\maxeig}{2\pi}^{d} \pf{2\pi}{\mineig}^{\fc d2} \int_{\R^d} e^{- \maxeig(\ve{x}^2 - \rc 2\ve{x-\mu}^2)}\dx\\
&=\pf{\maxeig}{\mineig}^{\fc d2} \pf{\maxeig}{2\pi}^{\fc d2} \int_{\R^d} e^{-\fc{\maxeig}2\ve{x+\mu}^2 + \maxeig \ve{\mu}^2}\dx\\
&\le \pf{\maxeig}{\mineig}^{\fc d2} e^{\maxeig D^2}.
\end{align}
\end{proof}

\subsection{A probability ratio calculation}

\begin{lem}\label{lem:delta} 
Suppose that $f(x) = -\ln \ba{\sumo in  w_i e^{-\fc{\ve{x-\mu_i}^2}2}}$, $p(x)\propto e^{-f(x)}$, and for $\al\ge 0$ let $p_\al(x) \propto e^{-\al f(x)}$, $Z_\al =\int_{\R^d} e^{-\al f(x)}\dx$. 
Suppose that $\ve{\mu_i}\le D$ for all $i$. 

If $\al<\be$, then
\begin{align}
\ba{
\int_{A} \min\{p_{\al}(x),p_\be(x)\} \dx 
}/p_\be(A)
&\ge
\min_x\fc{p_\al(x)}{p_\be(x)} \ge \fc{Z_\be}{Z_\al}\\
\fc{Z_\be}{Z_\al}&
\in \ba{
\rc2 e^{-2(\be-\al)\pa{D+\rc{\sqrt\al}\pa{\sqrt d + 2\sqrt{\ln \pf{2}{w_{\min}}}}}^2}, 1}.
\end{align}
Choosing $\be-\al=O\prc{D^2+\fc{d}{\al}+\rc{\al}\ln\prc{w_{\min}}}$, this quantity is $\Om(1)$.
\end{lem}
This is a special case of the following more general lemma.

\begin{lem}\label{lem:delta-gen}
Suppose that 
$f(x)=-\ln \ba{\sumo in w_i e^{-f_i(x)}}$, where $f_i(x)=f_0(x-\mu_i)$ and 
$f_0$ is $\mineig$-strongly convex and $\maxeig$-smooth. 
Let $P$, $P_\al$ (for $\al> 0$) be probability measures with densities
$p(x)\propto e^{-f(x)}$ and $p_\al(x) \propto e^{-\al f(x)}$. Let $Z_\al =\int_{\R^d} e^{-\al f(x)}\dx$. 
Suppose that $\ve{\mu_i}\le D$ for all $i$. 

Let $C=D + {\rc{\sqrt{\al \mineig}} \pa{\sqrt{d} + \sqrt{d\ln \pf{\maxeig}{\mineig} + 2\ln \pf{2}{w_{\min}}}}}$. 
If $\al<\be$, then
\begin{align}
\ba{
\int_{A} \min\{p_{\al}(x),p_\be(x)\} \dx 
}/p_\be(A)
&\ge
\min_x\fc{p_\al(x)}{p_\be(x)} \ge \fc{Z_\be}{Z_\al}\\
\fc{Z_\be}{Z_\al}&
\in \ba{
\rc2 e^{
-\rc2 (\be-\al) \maxeig C^2 
}, 1}.
\end{align}
If $\be-\al=O\pf{1}{
\maxeig\pa{D^2 + \fc{d}{\al \mineig}\pa{1+\ln\pf{\maxeig}{\mineig}} + \rc{\al\mineig} \ln \prc{w_{\min}}} 
}$, then this quantity is $\Om(1)$.
\end{lem}
\begin{proof}
Let $\wt P_\al$ be the probability measure with density function $\wt p_\al(x) \propto 
\sumo in w_i e^{-\al f_0(x-\mu_i)} 
$. 

By Lemma~\ref{lem:close-to-sum} and Lemma~\ref{lem:conc-mode}, since $\al f$ is $\al \mineig$-strongly, convex,
\begin{align}
\Pj_{x\sim P} (\ve{x}\ge C) 
&\le \rc{w_{\min}} \Pj_{x\sim \wt P_{\al}}(\ve{x}\ge C)\\
&\le \rc{w_{\min}}
\sumo in w_i \Pj_{x\sim \wt P_{\al}} (\ve{x}\ge C)\\
&\le  \rc{w_{\min}}\sumo in w_i \Pj_{x\sim 
P}(\ve{x}^2 \ge (C-D)^2)\\
&=\rc{w_{\min{}}} \Pj_{x\sim P} \ba{
	\ve{x}^2 \ge \rc{\al\mineig} \pa{
	\sqrt{d} + \sqrt{d\ln \pf{\maxeig}{\mineig} + 2\ln \pf{2}{w_{\min}}}	
}^2}\\
&\le  \rc{w_{\min{}}}\fc{w_{\min{}}}2 =\rc2.
\end{align}

Thus, using $f(x)\ge 0$, 
\begin{align}
\ba{
\int_{A} \min\bc{p_{\al}(x),p_\be(x)} \dx 
}/p_\be(A)
&\ge \int_A \min\bc{\fc{p_{\al}(x)}{p_\be(x)},1} p_\be(x)\dx\Big/p_\be(A)\\
&\ge \int_A \min\bc{\fc{Z_\be}{Z_\al} e^{(\be-\al)f(x)}, 1} p_\be(x)\dx\Big /p_\be(A)\\
&\ge \fc{Z_\be}{Z_\al}
\\
& = \fc{\int e^{-\be f(x)}\dx}{\int e^{-\al f(x)}\dx}\\
&=\int_{\R^d} e^{(-\be + \al)f(x)}p_\al(x)\dx\\
&\ge \int_{\ve{x}\le C} e^{(-\be + \al)f(x)}p_\al(x)\dx\\
&\ge \rc2 e^{-(\be-\al)\max_{\ve{x}\le C}(f(x))}\\
&\ge \rc 2 e^{-\rc2(\be-\al)MC^2}.
\end{align}
\end{proof}

\subsection{Other facts}
\begin{lem}\label{lem:poisson-tail}
Let $(N_T)_{T\ge 0}$ be a Poisson process with rate $\la$. Then there is a constant $C$ such that 
\begin{align}
\Pj(N_T \ge n) & \le \pf{Cn}{T\la}^{-n}.
\end{align}
\end{lem}
\begin{proof}
Assume $n>T\la$. We have by Stirling's formula
\begin{align}
\Pj(N_T \ge n) & = e^{-\la T}\sum_{m=n}^{\iy}\fc{(\la T)^m}{m!}\\
&\le e^{-\la T}\rc{n!}\rc{1-\fc{\la T}{n}}(\la T)^n\\
&=e^{-\la T} O\pa{n^{-\rc 2}\pf{e\la T}{n}^n}\\
&\le \pf{C n}{\la T}^n
\end{align}
for some $C$, since $e^{-\la T}\ge e^{-n}$.
\end{proof}

\end{document}